\documentclass{article}

\usepackage[preprint, eandd]{neurips_2026} 

\usepackage[utf8]{inputenc}
\usepackage[T1]{fontenc}
\usepackage{url}
\usepackage{multirow}
\usepackage{booktabs}
\usepackage{tabularx}
\usepackage{textcomp}
\usepackage{capt-of}
\usepackage{amsmath,amssymb,amsfonts}
\usepackage{graphicx}
\usepackage{xcolor}
\usepackage{booktabs}
\usepackage{nicefrac}
\usepackage{microtype}
\usepackage{makecell}
\usepackage{listings}
\usepackage{hyperref}
\hypersetup{hidelinks}
\usepackage{array}
\usepackage{float}
\usepackage{pgfplots}
\pgfplotsset{compat=1.18}
\usepackage{longtable}
\usepackage{pdflscape}
\usepackage{array}
\usepackage{tabularx}
\usepackage{array}
\usepackage{enumitem}
\providecommand{\RaggedRight}{\raggedright}
\definecolor{pbPass}{RGB}{44,160,44}
\definecolor{pbBuildFail}{RGB}{214,39,40}
\definecolor{pbRunFail}{RGB}{255,127,14}
\definecolor{pbVerifyFail}{RGB}{148,103,189}
\definecolor{pbExtractionFail}{RGB}{140,86,75}
\definecolor{pbNA}{RGB}{200,200,200}
\definecolor{pbTeal}{RGB}{23,190,207}
\definecolor{pbTealDark}{RGB}{15,120,140}
\definecolor{pbTealLight}{RGB}{158,218,229}
\definecolor{pbRose}{RGB}{227,119,194}
\definecolor{pbOrange}{RGB}{255,127,14}
\definecolor{pbYellow}{RGB}{255,215,0}
\definecolor{pbGray}{RGB}{150,150,150}
\definecolor{pbRodinia}{RGB}{31,119,180}
\definecolor{pbXSBench}{RGB}{255,127,14}
\definecolor{pbRSBench}{RGB}{44,160,44}
\definecolor{pbMixbench}{RGB}{214,39,40}
\definecolor{pbHeCBench}{RGB}{148,103,189}

\usepackage{caption}
\pgfplotsset{
  parbench compact/.style={
    width=\columnwidth, height=5.5cm,
    enlarge x limits=0.08,
    tick pos=left,
    legend style={font=\footnotesize, draw=gray!50},
    tick label style={font=\footnotesize},
    label style={font=\footnotesize},
    title style={font=\small\bfseries, align=center},
    ymajorgrids=true,
    grid style={draw=gray!20},
    clip=false,
  },
  parbench heatmap/.style={
    width=\columnwidth, height=\columnwidth,
    enlargelimits=false,
    tick label style={font=\tiny, rotate=45, anchor=east},
    title style={font=\small\bfseries},
    colorbar style={font=\tiny},
    axis on top,
  },
}

\definecolor{green2}{RGB}{0,100,0}

\graphicspath{{figures/}}

\colorlet{punct}{red!60!black}
\definecolor{background}{RGB}{242,242,242}
\definecolor{delim}{RGB}{20,105,176} 
\lstdefinelanguage{JSON}{
  basicstyle=\ttfamily\scriptsize,
  backgroundcolor=\color{background},
  showstringspaces=false,
  breaklines=true,
  literate=
    *{0}{{{\color{black}0}}}{1}
     {1}{{{\color{black}1}}}{1}
     {2}{{{\color{black}2}}}{1}
     {3}{{{\color{black}3}}}{1}
     {4}{{{\color{black}4}}}{1}
     {5}{{{\color{black}5}}}{1}
     {6}{{{\color{black}6}}}{1}
     {7}{{{\color{black}7}}}{1}
     {8}{{{\color{black}8}}}{1}
     {9}{{{\color{black}9}}}{1}
     {:}{{{\color{punct}{:}}}}{1}
     {,}{{{\color{punct}{,}}}}{1}
      {\{}{{{\color{delim}{\{}}}}{1}
      {\}}{{{\color{delim}{\}}}}}{1}
      {[}{{{\color{delim}{[}}}}{1}
      {]}{{{\color{delim}{]}}}}{1},
}

\newcommand{\qwen}{Qwen~3.5 397B-A17B}
\newcommand{\qwenshort}{Qwen~3.5}
\newcommand{\gptnew}{GPT-5.4}
\newcommand{\gptprovider}{Azure~OpenAI}
\newcommand{\codex}{GPT-5.3-codex}

\newcommand{\parbench}{\textsc{ParBench}}

\newcommand{\buildfail}{\textsc{Build\_Fail}}
\newcommand{\runfail}{\textsc{Run\_Fail}}
\newcommand{\verifyfail}{\textsc{Verify\_Fail}}
\newcommand{\extractionfail}{\textsc{Extract\_Fail}}
\newcommand{\pass}{\textsc{Pass}}

\newcommand{\knownfail}{\textsc{Known\_Fail}}

\newcommand{\tbd}[1][]{\textbf{\textcolor{red}{TBD\ifx\\#1\\\else~(#1)\fi}}}

\title{ParBench: A Benchmark for Reliable Evaluation of LLM Parallel Code Translation}

\author{%
    Samyak Jhaveri \\
    University of California, Irvine, USA \\
    \texttt{samyaknj@uci.edu} \\
    \And
    Erel Kaplan \\
    Technion, Haifa, Israel \\
    \texttt{erel.kaplan@campus.technion.ac.il} \\
    \And
    Tom Yotam \\
    Code Metal, USA \\
    \texttt{tom@codemetal.ai} \\
    \And
    Le Chen \\
    Argonne National Laboratory, Lemont, USA \\
    \texttt{lechen@anl.gov} \\
    \And
    Tomer Bitan \\
    Technion, Haifa, Israel \\
    \texttt{tomerbitan@campus.technion.ac.il} \\
    \And
    Niranjan Hasabnis \\
    Code Metal, USA \\ 
    \texttt{niranjan@codemetal.ai} \\
    \And
    Gal Oren \\
    Stanford University, Technion, USA \\
    \texttt{galoren@stanford.edu} \\
}

\begin{document}

\maketitle

\begin{abstract}
Modern compute-intensive software is written against a rapidly changing ecosystem of accelerators, programming APIs, compiler stacks, and portability layers. As hardware and software platforms evolve, kernels in AI, scientific computing, simulation, graphics, and data-intensive workloads often need to migrate across CUDA, OpenMP, OpenCL, OpenMP target offload, and related parallel APIs. Large language models and autonomous coding agents are increasingly proposed as tools for this migration, but the field still lacks a reliable way to measure whether they can preserve the low-level parallel semantics that make such translations behaviorally valid under declared oracles, including thread indexing, synchronization, memory management, host--device coordination, and API-specific execution structure.

We present \parbench{}, a kernel-centric benchmark framework designed to isolate and measure LLM-based parallel API translation under executable, reproducible conditions. Unlike general code benchmarks that are largely sequential, parallel-code benchmarks that emphasize generation from natural-language prompts, or repository-level studies that confound translation with build-system reconstruction, \parbench{} fixes the surrounding build, run, and verification infrastructure through declarative benchmark specifications and asks models to translate only the computational kernels. The benchmark draws on multiple open-source HPC suites to cover representative cross-API translation directions among CUDA, OpenMP, OpenCL, and OpenMP target offload. To probe whether success reflects robust translation rather than surface-form memorization, \parbench{} also includes AST-driven intended behavior-preserving, baseline-validated source augmentation. We further evaluate \parbench{} on state-of-the-art open and proprietary LLMs, showing that it exposes persistent barriers to reliable parallel code translation, including direction asymmetry, multi-file coordination, incomplete API adaptation, and uneven robustness to source-level perturbations. Code is available at \url{https://github.com/Scientific-Computing-Lab/ParBench}.
\end{abstract}

\begin{figure*}[t]
\centering
\includegraphics[width=\textwidth]{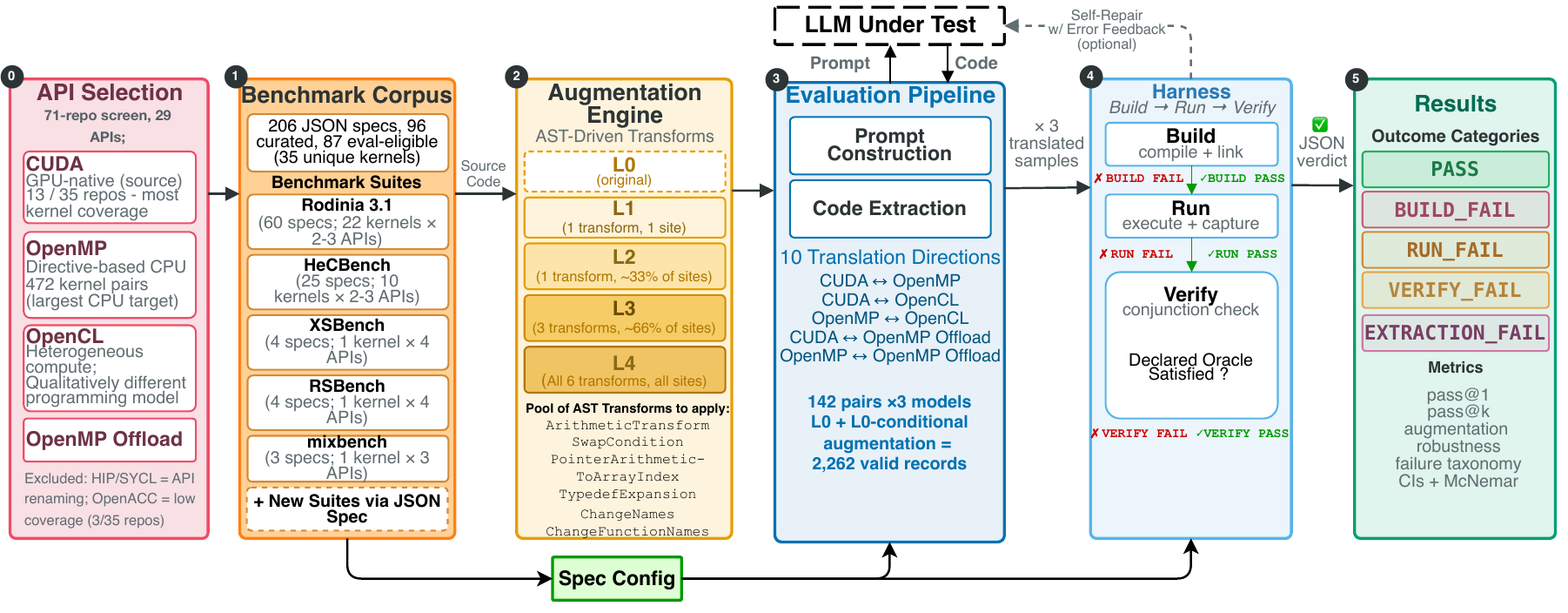}
\caption{
\textbf{\textsc{\parbench{}} isolates parallel API translation from repository-level confounds.}
The benchmark pipeline starts by selecting representative parallel APIs and curating executable kernel specifications from multi-API HPC suites. Each specification fixes the build, run, and verification context, while the model is asked to translate only the computational kernel code. AST-driven source augmentations create surface-varied source variants for robustness testing, and translated outputs are evaluated through an end-to-end build--run--verify harness with conjunctive correctness checks. The final outcome taxonomy separates harness-passing translations (oracle \pass{}: compiled, ran, and satisfied the spec's declared correctness oracle) from extraction, build, runtime, and verification failures, enabling both aggregate scoring and diagnostic analysis of where parallel translation breaks.
}
\label{fig:architecture}
\end{figure*}

\section{Introduction}
\label{sec:introduction}

Modern compute-intensive software is written against a rapidly changing ecosystem of accelerators, programming APIs, compiler stacks, and portability layers. Kernels in AI systems, scientific computing, simulation, graphics, and data-intensive workloads often need to move across CUDA, OpenMP, OpenCL, OpenMP target offload, and related parallel APIs as hardware platforms evolve or deployment requirements change. Large language models (LLMs) and autonomous coding agents are increasingly being explored for this migration task, but parallel API translation is fundamentally different from ordinary code translation. Correctness depends not only on preserving program intent, but also on maintaining thread-index arithmetic, synchronization semantics, memory-management patterns, host--device coordination, and API-specific execution structure--properties that go far beyond surface-level token replacement.

Can current models actually perform this translation reliably? Existing evaluations leave this question unresolved because they do not isolate parallel API translation as the measured capability. Widely adopted code-generation benchmarks such as HumanEval and SWE-bench~\cite{HumanEval2021, SWEbench2024} are predominantly sequential and therefore do not exercise parallel programming semantics. Parallel code-generation benchmarks such as ParEval~\cite{ParEval2024} primarily evaluate synthesis from natural-language descriptions, rather than translation of existing implementations across APIs. Repository-level translation work exposes important integration challenges, but it also entangles kernel translation with build-system reconstruction, file organization, dependency management, and execution setup~\cite{ParEvalRepo2025,Wang2026RepoTransBench}\footnote{An extended comparison with code-generation benchmarks, repository-level translation studies, HPC-specific LLM evaluation, and robustness benchmarks in Appendix~\ref{sec:appendix-d}.}. This leaves two core questions open: can models translate computational kernels across a broad set of parallel API directions when the surrounding infrastructure is fixed, and do successful translations reflect robust parallel-code translation rather than memorization of widely circulated benchmark code? 

We present \textsc{\parbench{}}, a benchmark framework designed to answer these questions by isolating parallel kernel translation from repository reconstruction. When evaluating parallel code translation, build-system configuration, file scope, run arguments, and verification criteria vary across kernels and APIs; failures in any of these can confound assessment of translation quality itself. \textsc{\parbench{}} fixes the surrounding build, run, and verification infrastructure through declarative benchmark specifications, asks models to translate only computational kernel files, and evaluates each generated kernel with an end-to-end build-run-verify harness using conjunctive correctness checks. The corpus is survey-grounded: we screened 71 open-source HPC repositories spanning 29 parallel programming APIs and retained 35 that met inclusion criteria for kernel-level co-occurrence analysis (Appendix Figure~\ref{fig:repo-vs-kernel}). From these, we selected five benchmark suites that provide equivalent multi-API kernels and automated correctness checks (Section~\ref{sec:suite-selection}). To test whether successful translations reflect robust translation rather than surface-form memorization, \textsc{\parbench{}} includes an AST-driven augmentation engine that generates source variants intended to preserve behavior and validated against the baseline harness on the retained subset.

This paper makes three contributions:

\begin{itemize}
\item \textbf{Parallel code translation benchmark.} We introduce \textsc{\parbench{}}, a benchmark for LLM-based parallel code translation that provides executable kernel specifications spanning five HPC benchmark suites, six standard translation directions among CUDA, OpenMP, and OpenCL, and four OpenMP-target case-study directions. By fixing the surrounding build infrastructure as context, \textsc{\parbench{}} enables direct build-run-verify evaluation of kernel translation without the build-system-generation confound identified by repository-level work.

\item \textbf{AST-driven augmentation engine.} We develop an augmentation engine with six AST-driven source-level transformations applied at four intensity levels (L1--L4). The engine produces structurally varied source variants 
to test whether models preserve parallel semantics beyond surface similarity to widely circulated benchmark code.

\item \textbf{Empirical characterization.} Across three models and 2{,}262 valid translation records, we report a multi-suite, multi-direction evaluation of LLM-based parallel code translation under stochastic sampling with pass@$k$ analysis. Pass@1 ranges from 23.9\% (\qwenshort{}) to 62.7\% (\gptnew{} and \codex{}) on 142 unique no-augmentation (L0) tasks, with the code-specialized \codex{} statistically indistinguishable from the general-purpose \gptnew{}. Key findings include build-stage API adaptation as the main bottleneck, strong direction and file-boundary effects, generally stable performance under augmentation with model-dependent variation at higher levels, and limited gains from repeated sampling (with $n{=}3$ samples, most failures remain hard failures across all attempts).
\end{itemize}

\section{\parbench{} Framework}
\label{sec:framework}

To better evaluate LLM-based translation of parallel programming APIs, \parbench{} defines a controlled, kernel-centric setting that separates parallel kernel translation from repository-level challenges such as build-system reconstruction. The framework realizes this setting through four components: (1)~declarative specifications, (2)~an AST-driven augmentation engine, (3)~an evaluation pipeline that orchestrates LLM translation and records diagnostic failure categories, and (4)~a build-run verify harness. Together, these components enable direct measurement of parallel API translation capability under end-to-end executability constraints while supporting controlled robustness analysis. Figure~\ref{fig:architecture} shows the overall architecture.

\subsection{Declarative Specifications and Kernel-Centric Translation}
\label{sec:spec-schema}

Each benchmark kernel–API variant in \parbench{} is encoded as a JSON specification (spec) that serves as a declarative contract defining what a passing execution means for that kernel--API variant under the declared oracle. The spec encodes all information required to build, run, and verify a benchmark without manual intervention, enabling fully automated evaluation. The schema is formally defined in JSON Schema and enforced by an offline validator. A spec declares the kernel's identity, specifies the target API, and includes provenance details to ensure reproducibility by including the pinned repository and commit. The spec partitions source files into \texttt{prompt\_payload} (source files shown to the LLM), \texttt{support\_files} (build infrastructure and shared headers; header content is included in the prompt as read-only context for interface compatibility), and \texttt{verification\_only} (reference implementations never shown to the LLM). A \texttt{translation\_targets} field identifies the subset of payload files the LLM must rewrite, enabling the kernel-centric translation described in Section~\ref{sec:eval-pipeline}.

\textbf{Build, run, and verification.} The \texttt{build} block specifies compiler commands, toolchain variables, and expected executable path. The \texttt{run} block defines input configurations with arguments and timeouts. The \texttt{verification} block declares an ordered list of verification strategies--exit-code checks, stdout pattern matching, quantitative comparison, and cryptographic hash verification--evaluated in conjunction; all must pass for an overall \pass{} verdict (Section~\ref{sec:harness}).

\textbf{Baseline results.} The \texttt{baseline\_results} block records the output from running the original implementation on a reference platform, providing reference values from which verification strategy parameters are calibrated. Section~\ref{sec:benchmark-curation} describes the construction of 87 verified evaluation specifications, with an example BFS specification shown in Listing~\ref{lst:bfs_spec_appendix}.


\subsection{Augmentation Engine: Probing Robustness via Code Perturbations}
\label{sec:augmentation-engine}

Widely used suites such as Rodinia~\cite{Rodinia2009} are likely present in LLM training data, so success on unmodified source may reflect memorization. The augmentation engine generates surface-varied source variants by applying six AST-level transforms backed by the \texttt{libclang} API: condition operand swapping (e.g., \texttt{a < b} $\to$ \texttt{b > a}), arithmetic form conversion (e.g., \texttt{x += 1} $\leftrightarrow$ \texttt{x = x + 1}), scope-aware variable renaming, typedef expansion, pointer-to-array notation conversion (e.g., \texttt{*(arr + i)} $\to$ \texttt{arr[i]}), and static function renaming. The transform set is designed to preserve parallel structure (thread indexing, synchronization, memory management) while changing surface-level code features most likely to be memorized verbatim. Transforms that would alter parallel semantics (e.g., loop reordering) are deliberately omitted.

Five augmentation levels (L0--L4) control transform density. L0 is unmodified source. L1 applies one randomly selected transform to one candidate site. L2--L4 apply increasing fractions of applicable transforms ($f{=}0.33$, $0.66$, $1.0$), each to the same fraction of candidate sites. Baseline validation supports the intended behavior-preserving design on the retained subset: all 87 non-\knownfail{} specs pass at L1--L2, and 80 of 87 pass at all levels through L4. The 7 high-intensity baseline failures occur exclusively in \texttt{omp\_target} variants, where condition swapping is associated with numerically different device-code behavior under the offloading toolchain. Variable renaming is disabled for files containing OpenMP pragmas. These validation failures are treated as augmentation/toolchain-specific exclusions rather than evidence about model robustness for the affected \texttt{omp\_target} baselines. Level definitions and per-transform details appear in Appendix Table~\ref{tab:augmentation-levels} and Appendix~\ref{sec:appendix-e1}.

\subsection{Evaluation Pipeline: Kernel-Centric Translation}
\label{sec:eval-pipeline}

The key design choice is kernel-centric translation: each spec partitions files into three roles. Fixed build infrastructure (Makefiles, build commands, verification logic) is never part of the translation task. Read-only context--source headers from \texttt{support\_files} and target-side infrastructure (non-kernel payload files, shared headers)--is provided to the LLM for interface compatibility but is not part of the translation output. Translated source files (\texttt{translation\_targets}) are what the LLM must rewrite; for full-program translations (e.g., CUDA$\leftrightarrow$OpenMP), these include host-side API calls, whereas for kernel-only translations (e.g., OpenCL \texttt{.cl} files), only the device kernel is translated. This separation isolates parallel API translation from repository reconstruction--the confound that drives ParEval-Repo's 0\% pass rate on applications above 133~SLoC~\cite{ParEvalRepo2025}.

The pipeline constructs a structured prompt containing the source code, target API and output filenames, the build command, and read-only infrastructure context, with prompt anonymization (generic file labels, stripped kernel names and source comments) to reduce benchmark identity leakage. For cross-API translations, argument and verification asymmetry are handled automatically: full-program translations use the source spec's run arguments and match against a union of source and target output patterns, while kernel-only translations use the target spec directly since the host runtime is unchanged. Each LLM response is parsed via a four-tier extraction strategy (Appendix~\ref{sec:appendix-a}) that progressively relaxes from explicit filename annotations to fuzzy proximity matching and single-file elimination. If any expected file cannot be recovered, the task is classified as \extractionfail{}. Source--target pairs are further classified during analysis by translation cardinality--\texttt{single\_file}, \texttt{multi\_to\_single}, \texttt{single\_to\_multi}, or \texttt{multi\_to\_multi}--enabling stratified difficulty analysis. The full prompt template is in Appendix~\ref{sec:appendix-g}.

\subsection{Harness: Build, Run, Verify}
\label{sec:harness}

The harness is a three-stage pipeline where failure at any stage short-circuits subsequent stages. The \textbf{build stage} resolves the spec's working directory, substitutes platform-specific paths, and invokes compilation. The \textbf{run stage} executes the compiled binary with spec-defined arguments and timeout. The \textbf{verify stage} applies verification strategies in conjunction--all must pass for a \pass{} verdict. Five strategy types are supported: \texttt{exit\_code}, \texttt{stdout\_pattern}, \texttt{stdout\_exclude\_pattern}, \texttt{numeric\_comparison}, and \texttt{file\_hash}. Accordingly, \pass{} denotes successful build, execution, and satisfaction of the declared verification oracle, not a proof of full semantic equivalence.

Conjunctive verification is essential. For example, a CUDA translation of Gaussian elimination compiles, runs to completion with exit code zero, and produces partial stdout--but omits the expected timing summary, indicating a systematic translation defect that compile-only (\cite{CodeRosetta2024}) or exit-code-only verification would miss. Each evaluation trial receives one of four failure labels--\buildfail{}, \runfail{}, \verifyfail{}, or \extractionfail{} (when no compilable code can be parsed from the LLM response)--or \pass{}, enabling systematic failure-mode analysis (Section~\ref{sec:failure-taxonomy}).

%

\section{Benchmark Curation}
\label{sec:benchmark-curation}
The benchmark corpus was assembled through a four-stage systematic process: surveying open-source HPC benchmark repositories, quantifying kernel-level translation opportunities across parallel APIs, filtering candidate kernels through automated build-run-verify checks, and verifying each selected kernel through the complete pipeline on the evaluation platform. This curation populates the Source Code component of the \parbench{} architecture (Figure~\ref{fig:architecture}) with kernel files from five benchmark suites spanning four parallel APIs--CUDA, OpenMP, OpenCL, and OpenMP target offload--with the first three serving as standard translation directions and OMP-target as a case-study extension (Section~\ref{sec:experimental-setup}).

\subsection{Suite and Kernel Selection}
\label{sec:suite-selection}

We screened 71 open-source HPC repositories spanning benchmark suites, mini-applications, proxy applications, libraries, and microbenchmarks across 29 parallel programming APIs, retaining 35 that met inclusion criteria for detailed kernel-level analysis (Appendix~\ref{sec:appendix-b1}).

A central finding is that \textbf{repository-level counting dramatically understates the available material for parallel translation evaluation}: although only 6~repositories contain both CUDA and OpenMP implementations, kernel-level analysis reveals 472~independent CUDA--OpenMP translation pairs--a 79$\times$ multiplier (Appendix Figure~\ref{fig:repo-vs-kernel}). This gap motivates \parbench{}'s individual kernel-centric evaluation design, rather than evaluating repository-level translation.

Five suites were selected based on their open-source availability, multi-API kernel equivalence, build-run-verify automation, self-checking correctness labels, and domain diversity; detailed criteria and exclusion rationale are provided in Appendix~\ref{sec:appendix-c2}--\ref{sec:appendix-c6}. \textbf{Rodinia}~\cite{Rodinia2009} is the largest contributor, providing 60~specs across 22~kernels over CUDA, OpenMP, and OpenCL; 53 pass baseline verification and 7 are \knownfail{}. \textbf{HeCBench}~\cite{HeCBench2023} provides the largest initial multi-API pool: from 522~kernels, a structured funnel yields 10 curated kernels--five with CUDA, OpenMP, and OMP-target variants, and five with CUDA and OMP-target only--totaling 25~specs.
\textbf{XSBench}~\cite{XSBench2014} and \textbf{RSBench}~\cite{RSBench2015} contribute nuclear-physics proxy kernels, each with 4~specs, while \textbf{mixbench}~\cite{mixbench2017} contributes 3~GPU roofline micro-benchmark specs. All XSBench, RSBench, and mixbench specs pass baseline verification. \footnote{Corpus commits: Rodinia~\href{https://github.com/yuhc/gpu-rodinia/tree/9c10d3ea}{\texttt{9c10d3ea}}; HeCBench~\href{https://github.com/ORNL/HeCBench/tree/22785cdd}{\texttt{22785cdd}}; XSBench~\href{https://github.com/ANL-CESAR/XSBench/tree/ba08e522}{\texttt{ba08e522}}; RSBench~\href{https://github.com/ANL-CESAR/RSBench/tree/34b64478}{\texttt{34b64478}}; mixbench~\href{https://github.com/ekondis/mixbench/tree/32edeca9}{\texttt{32edeca9}}.}

\subsection{Evaluation Corpus}
\label{sec:eval-corpus}

The 87 verified-\pass{} specs constitute the evaluation corpus (Table~\ref{tab:suite-summary}). Kernel complexity spans more than an order of magnitude--80 to 3{,}304~SLoC, median 271--and 31 of 35 kernels (89\%) exceed the 133-SLoC scale at which prior repository-level translation reports 0\% pass@1~\cite{ParEvalRepo2025}. Of the 96 curated specs (87~\pass{} plus 9~\knownfail{}), 25\% require multi-file translation; CUDA has the highest multi-file rate (51\%) owing to separate host and kernel sources, versus 12\% for OpenMP and 13\% for OpenCL (Appendix Table~\ref{tab:benchmark-characterization}). These properties place the corpus well beyond single-function benchmarks while retaining fixed build--run--verify infrastructure. The corpus draws predominantly from Rodinia (53 of 87 non-\knownfail{} specs, 61\%), so aggregate pass rates partly reflect its characteristic stencil, graph-traversal, and simulation-loop kernels.

\section{Experimental Setup}
\label{sec:experimental-setup}

\begin{table*}[!tbp]
\centering
\caption{Evaluation corpus. SLoC is measured on the CUDA source variant when available. Multi-File denotes specs requiring more than one translated source file. KF = \knownfail{}.}
\label{tab:suite-summary}
\small
\setlength{\tabcolsep}{3.5pt}
\renewcommand{\arraystretch}{1.08}
\begin{tabularx}{\textwidth}{@{}lcccc>{\raggedright\arraybackslash}Xrcr@{}}
\toprule
Suite & Kernels & Specs & PASS & KF & APIs & SLoC Range & Med. & Multi-File \\
\midrule
Rodinia   & 22 & 60 & 53 & 7 & CUDA, OMP, OCL & 195--3{,}304 & 334 & 21/60 (35\%) \\
HeCBench  & 10 & 25 & 23 & 2 & CUDA, OMP, OMP-tgt & 80--235 & 180 & 0/25 (0\%) \\
XSBench   & 1  & 4  & 4  & 0 & CUDA, OMP, OCL, OMP-tgt & 1{,}390 & -- & 2/4 (50\%) \\
RSBench   & 1  & 4  & 4  & 0 & CUDA, OMP, OCL, OMP-tgt & 1{,}016 & -- & 1/4 (25\%) \\
mixbench  & 1  & 3  & 3  & 0 & CUDA, OMP, OCL & 312 & -- & 0/3 (0\%) \\
\midrule
\textbf{Total} & \textbf{35} & \textbf{96} & \textbf{87} & \textbf{9} & & \textbf{80--3{,}304} & \textbf{271} & \textbf{24/96 (25\%)} \\
\bottomrule
\end{tabularx}
\end{table*}


\textbf{Models.}
We evaluate three models: \qwen{}, a 397B-parameter MoE open-weight model accessed via Together~AI; \gptnew{}, a general-purpose proprietary model accessed via \gptprovider{}; and \codex{}, a code-specialized model optimized via reinforcement learning on software engineering tasks~\cite{GPT53Codex2026}, also accessed via \gptprovider{}.

All three expose provider-specific reasoning modes, configured at provider-recommended settings (\texttt{enable\_thinking} for \qwenshort{}, \texttt{reasoning\_effort=medium} for \gptnew{} and \codex{}). Because these thinking modes differ in mechanism, observed performance gaps may reflect reasoning-mode effects in addition to base capability. Detailed model configurations, hardware specifications, and cost accounting are provided in Appendix~\ref{sec:appendix-a-extended}.

\textbf{Translation Protocol and Scope.}
From the 96-spec corpus (Section~\ref{sec:benchmark-curation}), we exclude 9~specifications classified as \knownfail{} due to pre-existing toolchain incompatibilities (Appendix~\ref{sec:appendix-a-extended}), leaving 87~verified-\pass{} specs that yield 142 unique translation tasks across ten directions: six standard bidirectional directions among CUDA, OpenMP, and OpenCL, plus four OMP-target case-study directions. Translations follow the kernel-centric protocol (Section~\ref{sec:framework}): only kernel source files are translated while host and build infrastructure remain fixed.

We evaluate oracle-defined translation correctness only (Section~\ref{sec:harness}); performance speedup is out of scope.

\textbf{Evaluation Phases.} 
The canonical campaign~(L0) evaluates the 142 non-\knownfail{} pairs from unmodified source code for each model, drawing $n=3$ independent samples per task in single-attempt mode (no iterative self-repair). Sampling uses temperature~0.7 for \qwenshort{}; for \gptnew{} and \codex{}, temperature is provider-controlled. Across three models this produces 1{,}278 L0 records. The augmentation campaign applies an L0-conditional filter, retaining only pairs where at least one L0 sample passes, and evaluates $n=1$
sample at each augmentation level L1--L4. Extended sampling configurations and deterministic seed derivation are described in Appendix~\ref{sec:appendix-a-extended}.

\textbf{Metrics.} 
The primary metric is \textbf{pass@$k$}, using the unbiased estimator from \citet{HumanEval2021} with normal-approximation confidence intervals on the task-level mean. Separately, we report \textbf{per-record pass rates} treating each record as an independent Bernoulli trial, computed with Wilson 95\% confidence intervals. Full metric derivations are in Appendix~\ref{sec:appendix-a-extended}.

\textbf{Oracle strength.} 
\pass{} denotes successful build, execution, and satisfaction of the declared verification oracle--not a proof of full semantic equivalence. Of the 87 eval-eligible specs, 2 use file-hash verification, 5 use numeric comparison, and the remaining 80 use stdout-pattern plus exit-code checks. We therefore interpret pass rates as \emph{declared-oracle correctness} and report oracle limitations explicitly in Section~\ref{sec:discussion} and Appendix~\ref{sec:appendix-k}.

With the methodology established, Section~\ref{sec:results} presents the pass rates, failure taxonomy, direction asymmetry, and augmentation robustness results.

\section{Results}
\label{sec:results}

We first report the primary task-level pass@$k$ results on the balanced L0 task set, then summarize record-level outcomes across the full campaign.

\subsection{pass@$k$ Analysis}
\label{sec:passk-analysis}
Restricting to the 142 L0 tasks, \qwenshort{} achieves pass@1\,=\,23.9\% and pass@3\,=\,35.2\%, while \gptnew{} achieves pass@1\,=\,62.7\% and pass@3\,=\,69.7\%. \codex{} matches \gptnew{}'s pass@1 exactly (62.7\%) with a slightly narrower pass@1-to-pass@3 gap (5.6\,pp versus 7.0\,pp for \gptnew{}; Appendix~\ref{sec:appendix-e4}). Stochastic resampling provides diminishing returns: for \qwenshort{}, 64.8\% of tasks produce no passing sample across three attempts, while only 13.4\% pass all three. \gptnew{} and \codex{} show the inverse profile--53.5\% and 57.0\% pass deterministically--yet 30.3\% and 31.7\% still hard-fail on every attempt. Across all models, failures are predominantly systematic (incorrect API-surface mapping) rather than stochastic; resampling cannot rescue them.

\subsection{Record-Level Diagnostic Outcomes}
\label{sec:overall-pass}

Kernel-centric evaluation isolates executable kernel translation from repository reconstruction. Across all valid records (L0 and L0-conditional augmentation), \qwenshort{} passes 36.7\% (230/626), \gptnew{} passes 75.5\% (621/822), and \codex{} passes 74.2\% (604/814; full breakdown in Appendix Table~\ref{tab:overall-pass}). \codex{} is statistically indistinguishable from \gptnew{} across all valid records (Bonferroni-corrected $p = 1.0$, OR\,=\,0.93 [0.74, 1.16], Cohen's $h = -0.031$); this should not be interpreted as evidence that HPC-specific fine-tuning is unnecessary. Prior work such as ParEval-Repo reports 0\% pass@1 on repository-level applications larger than 133~SLoC~\cite{ParEvalRepo2025}; 89\% of our kernels exceed that threshold, yet \parbench{}'s kernel isolation enables \qwenshort{} to achieve 40.3\% on CUDA-to-OpenMP. This contrast suggests that repository reconstruction is a major confound in full-application benchmarks, though the different corpora, oracles, and models used in each study preclude a direct comparison.

\subsection{Direction Dependence}
\label{sec:direction-analysis}

Per-record pass rates vary strongly across directions (Appendix Table~\ref{tab:direction-rates}), reflecting structural API asymmetries. The four OMP-target rows are exploratory case studies ($m \leq 8$ kernels, $n \leq 24$ records) and should not be compared directly to the six standard directions; their wide confidence intervals preclude strong conclusions about OMP-target translation in general. Translating from CUDA to OpenMP replaces explicit device memory and launch configuration with compiler directives. The reverse requires introducing explicit synchronization and memory management. OpenCL translations introduce further complexity with separate host/kernel source and JIT compilation, leading to 0\% success for OpenCL-to-CUDA under \qwenshort{}. \codex{}'s direction profile closely tracks \gptnew{} across five of six standard directions (maximum gap 7\,pp, with two directions matching exactly). The exception is OMP-to-OpenCL, where \codex{} outperforms \gptnew{} by 9.8\,pp (82.4\% versus 72.5\%). Overall, direction difficulty is predominantly a property of the translation task, not the model. At the kernel level, performance is highly heterogeneous (full breakdown in Appendix~\ref{sec:appendix-e4}). Simple kernels pass at high rates, while those with irregular memory access or multi-file dependencies frequently fail entirely.

\begin{figure}[t]
  \centering
  \input{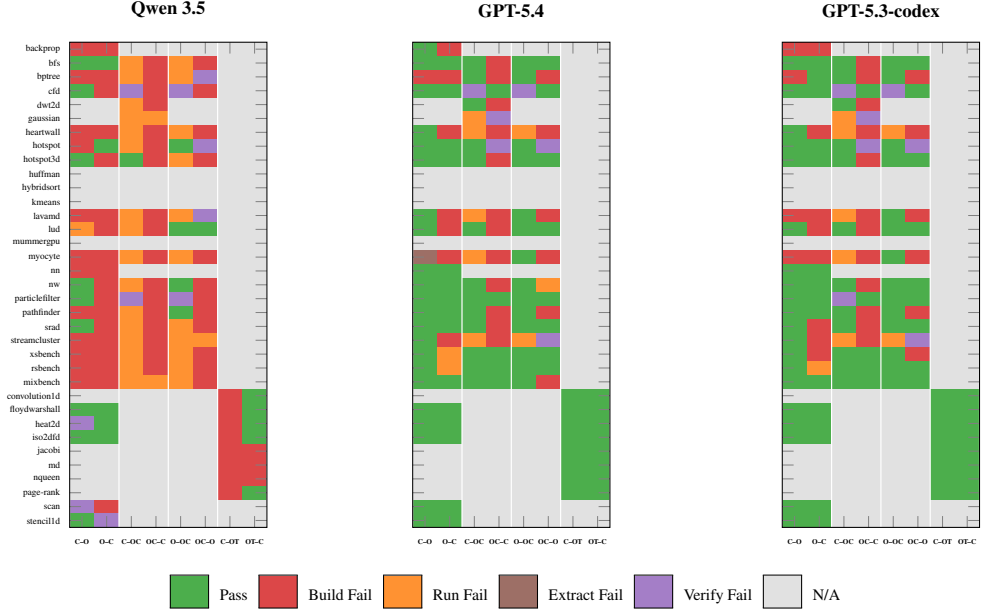}
  \caption{Per-kernel translation status across all directions and three models (L0, first sample per task, 35~kernels $\times$ up to 8~directions). All-grey rows denote kernels with only \knownfail{} or phantom specs.}
  \label{fig:kernel-heatmap-unified}
\end{figure}

\subsection{Failure Taxonomy}
\label{sec:failure-taxonomy}

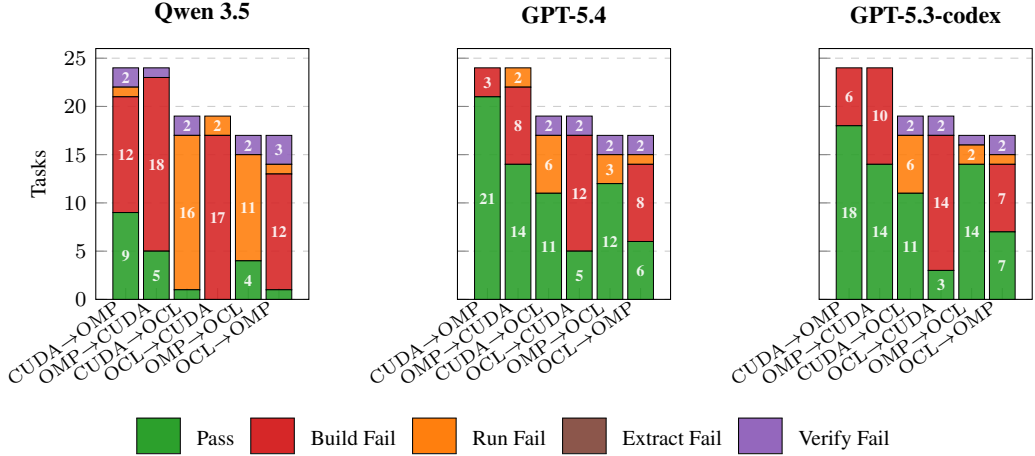
\begin{figure}[t]
\centering

\begingroup
\pgfplotsset{
  ffour failure axis/.style={
    width=\linewidth,
    height=4.9cm,
    ybar stacked,
    bar width=0.34cm,
    enlarge x limits=0.07,
    xtick={0,...,5},
    xticklabels={
        {\ensuremath{\mathrm{CUDA}{\to}\mathrm{OMP}}},
        {\ensuremath{\mathrm{OMP}{\to}\mathrm{CUDA}}},
        {\ensuremath{\mathrm{CUDA}{\to}\mathrm{OCL}}},
        {\ensuremath{\mathrm{OCL}{\to}\mathrm{CUDA}}},
        {\ensuremath{\mathrm{OMP}{\to}\mathrm{OCL}}},
        {\ensuremath{\mathrm{OCL}{\to}\mathrm{OMP}}}},
    xticklabel style={font=\scriptsize, rotate=35, anchor=east},
    yticklabel style={font=\footnotesize},
    xmin=-0.55,
    xmax=5.55,
    ymin=0,
    ymax=26,
    ytick={0,5,10,15,20,25},
    title style={font=\small\bfseries, align=center},
    ymajorgrids=true,
    xmajorgrids=false,
    major grid style={dashed, draw=gray!40},
    tick pos=left,
    clip=false,
  },
  ffour segment/.style={
    ybar stacked,
    draw=black,
    line width=0.3pt,
    fill opacity=0.9,
    nodes near coords={
      \pgfmathfloattoint{\pgfplotspointmeta}%
      \ifnum\pgfmathresult>1\relax\pgfmathresult\fi},
    nodes near coords align={center},
    every node near coord/.style={
      font=\tiny\bfseries,
      text=white,
      inner sep=1pt,
    },
  },
}

\newcommand{\ffourpanel}[7]{%
  \begin{minipage}[t]{0.315\textwidth}
  \centering
  \begin{tikzpicture}
  \begin{axis}[
    ffour failure axis,
    title={#1},
    #2
  ]
    \addplot[ffour segment, fill=pbPass]
      coordinates {#3};
    \addplot[ffour segment, fill=pbBuildFail]
      coordinates {#4};
    \addplot[ffour segment, fill=pbRunFail]
      coordinates {#5};
    \addplot[ffour segment, fill=pbExtractionFail]
      coordinates {#6};
    \addplot[ffour segment, fill=pbVerifyFail]
      coordinates {#7};
  \end{axis}
  \end{tikzpicture}
  \end{minipage}%
}

\ffourpanel{\qwenshort{}}%
  {ylabel={Tasks}, ylabel style={font=\footnotesize}}%
  {(0,9) (1,5) (2,1) (3,0) (4,4) (5,1)}%
  {(0,12) (1,18) (2,0) (3,17) (4,0) (5,12)}%
  {(0,1) (1,0) (2,16) (3,2) (4,11) (5,1)}%
  {(0,0) (1,0) (2,0) (3,0) (4,0) (5,0)}%
  {(0,2) (1,1) (2,2) (3,0) (4,2) (5,3)}%
\hfill
\ffourpanel{\gptnew{}}%
  {yticklabels={,,,,,,}}%
  {(0,21) (1,14) (2,11) (3,5) (4,12) (5,6)}%
  {(0,3) (1,8) (2,0) (3,12) (4,0) (5,8)}%
  {(0,0) (1,2) (2,6) (3,0) (4,3) (5,1)}%
  {(0,0) (1,0) (2,0) (3,0) (4,0) (5,0)}%
  {(0,0) (1,0) (2,2) (3,2) (4,2) (5,2)}%
\hfill
\ffourpanel{\codex{}}%
  {yticklabels={,,,,,,}}%
  {(0,18) (1,14) (2,11) (3,3) (4,14) (5,7)}%
  {(0,6) (1,10) (2,0) (3,14) (4,0) (5,7)}%
  {(0,0) (1,0) (2,6) (3,0) (4,2) (5,1)}%
  {(0,0) (1,0) (2,0) (3,0) (4,0) (5,0)}%
  {(0,0) (1,0) (2,2) (3,2) (4,1) (5,2)}%

\vspace{0.35ex}
\begin{center}
\footnotesize
\setlength{\tabcolsep}{3pt}
\begin{tabular}{@{}llllllllll@{}}
\fcolorbox{black}{pbPass}{\rule{0pt}{7pt}\rule{10pt}{0pt}} & Pass &
\fcolorbox{black}{pbBuildFail}{\rule{0pt}{7pt}\rule{10pt}{0pt}} & Build Fail &
\fcolorbox{black}{pbRunFail}{\rule{0pt}{7pt}\rule{10pt}{0pt}} & Run Fail &
\fcolorbox{black}{pbExtractionFail}{\rule{0pt}{7pt}\rule{10pt}{0pt}} & Extract Fail &
\fcolorbox{black}{pbVerifyFail}{\rule{0pt}{7pt}\rule{10pt}{0pt}} & Verify Fail
\end{tabular}
\end{center}
\endgroup
\caption{Failure taxonomy by translation direction across all three models (L0, first sample per task). The figure includes only the six standard directions, spanning 120~non-\knownfail{} tasks in total (24, 24, 19, 19, 17, and 17 tasks by direction), with OMP-target case-study directions omitted for cross-model comparability. Build-stage failures dominate \qwenshort{} across nearly all directions; \gptnew{} and \codex{} show substantially higher pass counts with residual build failures concentrated in OpenCL$\to$CUDA and OMP$\to$CUDA.}
\label{fig:failure-taxonomy}
\end{figure}


Figure~\ref{fig:failure-taxonomy} places all three models side-by-side on the same scale. Build-stage adaptation is the dominant failure mode: for \qwenshort{}, \buildfail{} accounts for 39.1\% of all records and 61.9\% of failures; these are incomplete API-surface adaptations (e.g., retained CUDA identifiers in OpenMP targets). \gptnew{} and \codex{} reduce build failures to 15.0\% and 17.1\% respectively, with residual failures concentrated in OCL$\to$CUDA and OMP$\to$CUDA directions. \verifyfail{} accounts for 4.6\% of \qwenshort{} records, proving that compile-only or zero-exit-code checks are insufficient for rigorous evaluation. \runfail{} (19.3\% for \qwenshort{}) is frequently driven by OpenCL JIT compilation errors at runtime.
  
\subsection{Augmentation Robustness}
\label{sec:augmentation-robustness}

To assess whether baseline success is driven primarily by surface-form memorization, we evaluated augmented source variants (L1--L4) on L0-conditional subsets (detailed in Appendix~\ref{sec:appendix-e4}). \codex{} shows the same plateau pattern as \gptnew{}: 85.6\%--88.7\% at L1--L4, in contrast to \qwenshort{}'s decline from 74.0\% at L1 to 56.0\% at L4 on the same L0-conditional design (Cochran--Armitage $z = -1.84$, $p = 0.065$; the decline is monotonic in sample means but does not reach significance at $\alpha = 0.05$ given $n = 50$ per level). This stability across both GPT models is compatible with robustness to the tested surface-form perturbations. These results are descriptive rather than confirmatory given the L0-conditional design; survivorship bias prevents ruling out deeper structural memorization entirely.


Per-kernel analysis reveals exceptions: on \texttt{stencil1d}, \qwenshort{} produces a passing parallel loop at L0 but reverts to a non-portable CUDA-style tiling pattern at L1 (\buildfail{}) and L2 (\verifyfail{}), recovers at L3 (\pass{}), then introduces a boundary-condition error at L4 (\verifyfail{})--illustrating how surface-form perturbation can expose fragile pattern-matching even when the model occasionally recovers.

\section{Discussion, Limitations, and Conclusion}
\label{sec:discussion}

\parbench{} shows that, when repository reconstruction is fixed, current LLMs can produce declared-oracle-passing translations for a non-trivial fraction of kernel-level parallel API tasks, but success remains highly structured by API direction, failure stage, and model tier. The primary bottleneck is build-stage API adaptation, especially when translations must introduce explicit device-memory and runtime structure. A task-level omnibus test on 142 L0 tasks ($\chi^2(2) = 44.4$, $p < 10^{-10}$, Cram\'er's $V = 0.32$) detects significant heterogeneity across three models, where each task is classified as pass-any (at least one of three samples passes) or all-fail; pairwise analysis reveals two tiers: \gptnew{} and \codex{} achieve near-identical task-level pass rates (99/142 and 97/142; Fisher's $p = 1.0$), while both surpass \qwenshort{} by $4.2\times$ odds (Cohen's $h = 0.71$). Because the providers use unmatched sampling conditions (Section~\ref{sec:sampling-config}), the Qwen--GPT gap cannot be causally attributed to model capability alone; the within-provider comparison controls for this confound. Full pairwise statistical details appear in Appendix~\ref{sec:appendix-e4}.

The tier gap is one of determinism, not marginal accuracy: GPT-family models either solve a task reliably across all three samples or fail every attempt, whereas Qwen's failures are equally systematic but far more frequent (Section~\ref{sec:passk-analysis}). All three models share the same dominant bottleneck--incomplete API-surface adaptation at build time--but the GPT family resolves the majority of these, leaving residual build failures concentrated in the hardest directions (Section~\ref{sec:failure-taxonomy}). Direction rankings are consistent across models: removing explicit parallel constructs (CUDA-to-OpenMP) is substantially easier than introducing them (OpenCL-to-CUDA), suggesting that the difficulty gradient tracks API-surface complexity rather than model-specific memorization (Section~\ref{sec:direction-analysis}). Augmentation robustness reinforces this interpretation: GPT-family pass rates are stable across L1--L4, while \qwenshort{}'s declining augmentation rates suggest greater reliance on surface-form patterns, but do not reach significance at $\alpha = 0.05$ (Section~\ref{sec:augmentation-robustness}).

\textbf{Limitations.} The benchmark is correctness-focused: it does not measure speedups, occupancy, memory bandwidth, or efficiency. A harness-passing translation may still be unusably slow, a concern also raised by TRACE~\cite{TRACE2026}. The corpus is finite; some directions have small paired samples. The L0-conditional augmentation subset ranges from 3--22 kernels per model per standard direction (3--8 for omp\_target directions), with the smallest cells reflecting \qwenshort{}'s low L0 pass rate, but the balanced cross-model comparison is limited to 12 common kernels in a single direction. Small per-level sample sizes and the L0-conditioned selection of augmentation subsets complicate formal trend interpretation. Ten specs across six kernels were downgraded from strong or medium oracles to weak (stdout pattern plus exit code): six due to cross-API floating-point reduction-order divergence (cfd, hotspot, myocyte) and four due to synthesis asymmetry between reference implementations (bfs, nw, nn). This limits oracle strength but correctly avoids penalizing faithful translations. All results are from a single platform (NVIDIA RTX~4070, Ubuntu~24.04, HPC SDK~24.3); for GPU and offload code, compiler and runtime behavior varies across NVIDIA architecture generations and toolchain versions, so results are reproducible on this pinned platform but cross-platform generality remains future work. The stochastic evaluation (temperature~0.7, three samples) captures variability but does not measure iterative repair or agentic translation strategies. The three models use different reasoning mechanisms and sampling configurations (Section~\ref{sec:sampling-config}). The Qwen--GPT gap is large (task-level Cohen's $h = 0.71$) but cannot be fully attributed to model capability given unmatched sampling conditions. The within-provider \gptnew{}--\codex{} comparison controls for provider-side differences but both share the same Azure OpenAI infrastructure, limiting generalizability beyond this provider. \codex{}'s code specialization targets general software engineering tasks; the null result for parallel translation may not generalize to models specifically fine-tuned on HPC code. Appendix~\ref{sec:appendix-k} provides a structured Evaluation Card and reporting protocol; the artifact (Appendix~\ref{sec:artifact-availability}) documents per-record outputs that support reproduction and extension.




\bibliographystyle{plainnat}
\bibliography{references}

\begin{thebibliography}{46}
\providecommand{\natexlab}[1]{#1}
\providecommand{\url}[1]{\texttt{#1}}
\expandafter\ifx\csname urlstyle\endcsname\relax
  \providecommand{\doi}[1]{doi: #1}\else
  \providecommand{\doi}{doi: \begingroup \urlstyle{rm}\Url}\fi

\bibitem[Bitan et~al.(2025)Bitan, Kadosh, Kaplan, Meiri, Chen, Morales,
  Hasabnis, and Oren]{UniPar2025}
Tomer Bitan, Tal Kadosh, Erel Kaplan, Shira Meiri, Le~Chen, Peter Morales,
  Niranjan Hasabnis, and Gal Oren.
\newblock Unipar: A unified llm-based framework for parallel and accelerated
  code translation in hpc.
\newblock In \emph{2025 IEEE High Performance Extreme Computing Conference
  (HPEC)}, pages 1--9, 2025.
\newblock \doi{10.1109/HPEC67600.2025.11196677}.

\bibitem[Chaturvedi et~al.(2025)Chaturvedi, Nichols, Singh, and
  Bhatele]{HPCCoderV2}
Aman Chaturvedi, Daniel Nichols, Siddharth Singh, and Abhinav Bhatele.
\newblock Hpc-coder-v2: Studying code llms across low-resource parallel
  languages.
\newblock In \emph{ISC High Performance 2025 Research Paper Proceedings (40th
  International Conference)}, pages 1--14, 2025.
\newblock \doi{10.23919/ISC.2025.11017585}.

\bibitem[Che et~al.(2009)Che, Boyer, Meng, Tarjan, Sheaffer, Lee, and
  Skadron]{Rodinia2009}
Shuai Che, Michael Boyer, Jiayuan Meng, David Tarjan, Jeremy~W. Sheaffer,
  Sang-Ha Lee, and Kevin Skadron.
\newblock Rodinia: A benchmark suite for heterogeneous computing.
\newblock In \emph{2009 IEEE International Symposium on Workload
  Characterization (IISWC)}, pages 44--54, 2009.
\newblock \doi{10.1109/IISWC.2009.5306797}.

\bibitem[Chen et~al.(2024)Chen, Bhattacharjee, Ahmed, Hasabnis, Oren, Vo, and
  Jannesari]{OMPGPT2024}
Le~Chen, Arijit Bhattacharjee, Nesreen Ahmed, Niranjan Hasabnis, Gal Oren,
  Vy~Vo, and Ali Jannesari.
\newblock Ompgpt: A generative pre-trained transformer model for openmp.
\newblock In \emph{Euro-Par 2024: Parallel Processing: 30th European Conference
  on Parallel and Distributed Processing, Madrid, Spain, August 26–30, 2024,
  Proceedings, Part I}, page 121–134, Berlin, Heidelberg, 2024.
  Springer-Verlag.
\newblock ISBN 978-3-031-69576-6.
\newblock \doi{10.1007/978-3-031-69577-3_9}.
\newblock URL \url{https://doi.org/10.1007/978-3-031-69577-3_9}.

\bibitem[Chen et~al.(2025)Chen, Ahmed, Capotă, Willke, Hasabnis, and
  Jannesari]{PCEBench2025}
Le~Chen, Nesreen Ahmed, Mihai Capotă, Ted Willke, Niranjan Hasabnis, and Ali
  Jannesari.
\newblock Pcebench: A multi-dimensional benchmark for evaluating large language
  models in parallel code generation.
\newblock In \emph{2025 IEEE International Parallel and Distributed Processing
  Symposium (IPDPS)}, pages 546--557, 2025.
\newblock \doi{10.1109/IPDPS64566.2025.00055}.

\bibitem[Chen et~al.(2021)Chen, Tworek, Jun, Yuan, de~Oliveira~Pinto, Kaplan,
  Edwards, Burda, Joseph, Brockman, Ray, Puri, Krueger, Petrov, Khlaaf, Sastry,
  Mishkin, Chan, Gray, Ryder, Pavlov, Power, Kaiser, Bavarian, Winter, Tillet,
  Such, Cummings, Plappert, Chantzis, Barnes, Herbert-Voss, Guss, Nichol,
  Paino, Tezak, Tang, Babuschkin, Balaji, Jain, Saunders, Hesse, Carr, Leike,
  Achiam, Misra, Morikawa, Radford, Knight, Brundage, Murati, Mayer, Welinder,
  McGrew, Amodei, McCandlish, Sutskever, and Zaremba]{HumanEval2021}
Mark Chen, Jerry Tworek, Heewoo Jun, Qiming Yuan, Henrique~Ponde
  de~Oliveira~Pinto, Jared Kaplan, Harri Edwards, Yuri Burda, Nicholas Joseph,
  Greg Brockman, Alex Ray, Raul Puri, Gretchen Krueger, Michael Petrov, Heidy
  Khlaaf, Girish Sastry, Pamela Mishkin, Brooke Chan, Scott Gray, Nick Ryder,
  Mikhail Pavlov, Alethea Power, Lukasz Kaiser, Mohammad Bavarian, Clemens
  Winter, Philippe Tillet, Felipe~Petroski Such, Dave Cummings, Matthias
  Plappert, Fotios Chantzis, Elizabeth Barnes, Ariel Herbert-Voss,
  William~Hebgen Guss, Alex Nichol, Alex Paino, Nikolas Tezak, Jie Tang, Igor
  Babuschkin, Suchir Balaji, Shantanu Jain, William Saunders, Christopher
  Hesse, Andrew~N. Carr, Jan Leike, Josh Achiam, Vedant Misra, Evan Morikawa,
  Alec Radford, Matthew Knight, Miles Brundage, Mira Murati, Katie Mayer, Peter
  Welinder, Bob McGrew, Dario Amodei, Sam McCandlish, Ilya Sutskever, and
  Wojciech Zaremba.
\newblock Evaluating large language models trained on code.
\newblock 2021.

\bibitem[Davis et~al.(2025)Davis, Nichols, Khillan, and
  Bhatele]{ParEvalRepo2025}
Joshua~H. Davis, Daniel Nichols, Ishan Khillan, and Abhinav Bhatele.
\newblock Pareval-repo: A benchmark suite for evaluating llms with
  repository-level hpc translation tasks.
\newblock In \emph{Proceedings of the 54th International Conference on Parallel
  Processing}, ICPP '25, page 94–103, New York, NY, USA, 2025. Association
  for Computing Machinery.
\newblock ISBN 9798400720741.
\newblock \doi{10.1145/3754598.3754669}.
\newblock URL \url{https://doi.org/10.1145/3754598.3754669}.

\bibitem[Dearing et~al.(2024)Dearing, Tao, Wu, Lan, and Taylor]{LASSI2024}
Matthew~T. Dearing, Yiheng Tao, Xingfu Wu, Zhiling Lan, and Valerie Taylor.
\newblock Lassi: An llm-based automated self-correcting pipeline for
  translating parallel scientific codes.
\newblock In \emph{2024 IEEE International Conference on Cluster Computing
  Workshops (CLUSTER Workshops)}, pages 136--143, 2024.
\newblock \doi{10.1109/CLUSTERWorkshops61563.2024.00029}.

\bibitem[Du et~al.(2024)Du, Tuan, Ji, Liu, and Ng]{Mercury2024}
Mingzhe Du, Luu~Anh Tuan, Bin Ji, Qian Liu, and See-Kiong Ng.
\newblock Mercury: A code efficiency benchmark for code large language models.
\newblock In A.~Globerson, L.~Mackey, D.~Belgrave, A.~Fan, U.~Paquet,
  J.~Tomczak, and C.~Zhang, editors, \emph{Advances in Neural Information
  Processing Systems}, volume~37, pages 16601--16622. Curran Associates, Inc.,
  2024.
\newblock \doi{10.52202/079017-0529}.
\newblock URL
  \url{https://proceedings.neurips.cc/paper_files/paper/2024/file/1df1df43b58845650b8dada00fca9772-Paper-Datasets_and_Benchmarks_Track.pdf}.

\bibitem[Etienne et~al.(2026)Etienne, Garcia~de Gonzalo, and
  Arnold]{CuratedOpenMPDataset2026}
Nichole Etienne, Simon Garcia~de Gonzalo, and Dorian Arnold.
\newblock Openmp-annotated code dataset for large language model fine-tuning on
  parallel programming tasks.
\newblock \emph{Frontiers in High Performance Computing}, 4, 2026.
\newblock ISSN 2813-7337.
\newblock \doi{10.3389/fhpcp.2026.1771927}.
\newblock URL
  \url{https://www.frontiersin.org/journals/high-performance-computing/articles/10.3389/fhpcp.2026.1771927}.

\bibitem[Gebru et~al.(2021)Gebru, Morgenstern, Vecchione, Vaughan, Wallach,
  III, and Crawford]{Datasheets2021}
Timnit Gebru, Jamie Morgenstern, Briana Vecchione, Jennifer~Wortman Vaughan,
  Hanna Wallach, Hal~Daum\'{e} III, and Kate Crawford.
\newblock Datasheets for datasets.
\newblock \emph{Commun. ACM}, 64\penalty0 (12):\penalty0 86–92, November
  2021.
\newblock ISSN 0001-0782.
\newblock \doi{10.1145/3458723}.
\newblock URL \url{https://doi.org/10.1145/3458723}.

\bibitem[Godoy et~al.(2024)Godoy, Valero-Lara, Teranishi, Balaprakash, and
  Vetter]{LargeLLMEvalHPC2024}
William~F. Godoy, Pedro Valero-Lara, Keita Teranishi, Prasanna Balaprakash, and
  Jeffrey~S. Vetter.
\newblock Large language model evaluation for high-performance computing
  software development.
\newblock \emph{Concurrency and Computation: Practice and Experience},
  36\penalty0 (26):\penalty0 e8269, 2024.
\newblock \doi{https://doi.org/10.1002/cpe.8269}.
\newblock URL \url{https://onlinelibrary.wiley.com/doi/abs/10.1002/cpe.8269}.

\bibitem[Gong et~al.(2026)Gong, Sun, Huang, Liang, Zhang, and Hao]{TRACE2026}
Zhihao Gong, Zeyu Sun, Dong Huang, Qingyuan Liang, Jie~M. Zhang, and Dan Hao.
\newblock Trace: Evaluating execution efficiency of llm-based code translation,
  2026.
\newblock URL \url{https://arxiv.org/abs/2508.11468}.

\bibitem[Harel et~al.(2023)Harel, Pinter, and Oren]{LearningToParallelize2023}
Re'em Harel, Yuval Pinter, and Gal Oren.
\newblock Learning to parallelize in a shared-memory environment with
  transformers.
\newblock In \emph{Proceedings of the 28th ACM SIGPLAN Annual Symposium on
  Principles and Practice of Parallel Programming}, PPoPP '23, page 450–452,
  New York, NY, USA, 2023. Association for Computing Machinery.
\newblock ISBN 9798400700156.
\newblock \doi{10.1145/3572848.3582565}.
\newblock URL \url{https://doi.org/10.1145/3572848.3582565}.

\bibitem[Harel et~al.(2025)Harel, Kadosh, Hasabnis, Mattson, Pinter, and
  Oren]{PragFormer2025}
Re’em Harel, Tal Kadosh, Niranjan Hasabnis, Timothy Mattson, Yuval Pinter,
  and Gal Oren.
\newblock Pragformer: Data-driven parallel source code classification with
  transformers.
\newblock \emph{Int. J. Parallel Program.}, 53\penalty0 (1), October 2025.
\newblock ISSN 0885-7458.
\newblock \doi{10.1007/s10766-024-00778-9}.
\newblock URL \url{https://doi.org/10.1007/s10766-024-00778-9}.

\bibitem[Ibrahimzada et~al.(2025)Ibrahimzada, Ke, Pawagi, Abid, Pan, Sinha, and
  Jabbarvand]{AlphaTrans2025}
Ali~Reza Ibrahimzada, Kaiyao Ke, Mrigank Pawagi, Muhammad~Salman Abid, Rangeet
  Pan, Saurabh Sinha, and Reyhaneh Jabbarvand.
\newblock Alphatrans: A neuro-symbolic compositional approach for
  repository-level code translation and validation.
\newblock \emph{Proc. ACM Softw. Eng.}, 2\penalty0 (FSE), June 2025.
\newblock \doi{10.1145/3729379}.
\newblock URL \url{https://doi.org/10.1145/3729379}.

\bibitem[ichiro Hayashi et~al.(2026)ichiro Hayashi, Morita, Mukunoki, Hoshino,
  and Katagiri]{VibeCodeHPC2026}
Shun ichiro Hayashi, Koki Morita, Daichi Mukunoki, Tetsuya Hoshino, and
  Takahiro Katagiri.
\newblock Vibecodehpc: An agent-based iterative prompting auto-tuner for hpc
  code generation using llms, 2026.
\newblock URL \url{https://arxiv.org/abs/2510.00031}.

\bibitem[Jimenez et~al.(2024)Jimenez, Yang, Wettig, Yao, Pei, Press, and
  Narasimhan]{SWEbench2024}
Carlos~E Jimenez, John Yang, Alexander Wettig, Shunyu Yao, Kexin Pei, Ofir
  Press, and Karthik~R Narasimhan.
\newblock {SWE}-bench: Can language models resolve real-world github issues?
\newblock In \emph{The Twelfth International Conference on Learning
  Representations}, 2024.
\newblock URL \url{https://openreview.net/forum?id=VTF8yNQM66}.

\bibitem[Jin and Vetter(2023)]{HeCBench2023}
Zheming Jin and Jeffrey~S. Vetter.
\newblock A benchmark suite for improving performance portability of the sycl
  programming model.
\newblock In \emph{2023 IEEE International Symposium on Performance Analysis of
  Systems and Software (ISPASS)}, pages 325--327, 2023.
\newblock \doi{10.1109/ISPASS57527.2023.00041}.

\bibitem[Jin et~al.(2025)Jin, Pophale, and Teranishi]{ChatPORTSYCL2025}
Zheming Jin, Swaroop Pophale, and Keita Teranishi.
\newblock Enhancing chatport with cuda-to-sycl kernel translation capability.
\newblock In \emph{Proceedings of the SC '25 Workshops of the International
  Conference for High Performance Computing, Networking, Storage and Analysis},
  SC Workshops '25, page 524–533, New York, NY, USA, 2025. Association for
  Computing Machinery.
\newblock ISBN 9798400718717.
\newblock \doi{10.1145/3731599.3767398}.
\newblock URL \url{https://doi.org/10.1145/3731599.3767398}.

\bibitem[Kadosh et~al.(2023{\natexlab{a}})Kadosh, Hasabnis, Mattson, Pinter,
  and Oren]{HPCorpus2023}
Tal Kadosh, Niranjan Hasabnis, Timothy Mattson, Yuval Pinter, and Gal Oren.
\newblock Quantifying openmp: Statistical insights into usage and adoption.
\newblock In \emph{2023 IEEE High Performance Extreme Computing Conference
  (HPEC)}, pages 1--7, 2023{\natexlab{a}}.
\newblock \doi{10.1109/HPEC58863.2023.10363459}.

\bibitem[Kadosh et~al.(2023{\natexlab{b}})Kadosh, Schneider, Hasabnis, Mattson,
  Pinter, and Oren]{OMPify2023}
Tal Kadosh, Nadav Schneider, Niranjan Hasabnis, Timothy Mattson, Yuval Pinter,
  and Gal Oren.
\newblock Advising openmp parallelization via a graph-based approach with
  transformers.
\newblock In \emph{OpenMP: Advanced Task-Based, Device and Compiler
  Programming: 19th International Workshop on OpenMP, IWOMP 2023, Bristol, UK,
  September 13–15, 2023, Proceedings}, page 3–17, Berlin, Heidelberg,
  2023{\natexlab{b}}. Springer-Verlag.
\newblock ISBN 978-3-031-40743-7.
\newblock \doi{10.1007/978-3-031-40744-4_1}.
\newblock URL \url{https://doi.org/10.1007/978-3-031-40744-4_1}.

\bibitem[Kadosh et~al.(2024{\natexlab{a}})Kadosh, Hasabnis, Soundararajan, Vo,
  Capota, Ahmed, Pinter, and Oren]{OMPar2024}
Tal Kadosh, Niranjan Hasabnis, Prema Soundararajan, Vy~A. Vo, Mihai Capota,
  Nesreen Ahmed, Yuval Pinter, and Gal Oren.
\newblock Ompar: Automatic parallelization with ai-driven source-to-source
  compilation, 2024{\natexlab{a}}.
\newblock URL \url{https://arxiv.org/abs/2409.14771}.

\bibitem[Kadosh et~al.(2024{\natexlab{b}})Kadosh, Hasabnis, Vo, Schneider,
  Krien, Capotă, Wasay, Tamir, Willke, Ahmed, Pinter, Mattson, and
  Oren]{MonoCoder2024}
Tal Kadosh, Niranjan Hasabnis, Vy~A. Vo, Nadav Schneider, Neva Krien, Mihai
  Capotă, Abdul Wasay, Guy Tamir, Ted Willke, Nesreen Ahmed, Yuval Pinter,
  Timothy Mattson, and Gal Oren.
\newblock Monocoder: Domain-specific code language model for hpc codes and
  tasks.
\newblock In \emph{2024 IEEE High Performance Extreme Computing Conference
  (HPEC)}, pages 1--7, 2024{\natexlab{b}}.
\newblock \doi{10.1109/HPEC62836.2024.10938441}.

\bibitem[Kaplan et~al.(2026)Kaplan, Bitan, Ghrayeb, Chen, Yotam, Hasabnis, and
  Oren]{ParaCodex2026}
Erel Kaplan, Tomer Bitan, Lian Ghrayeb, Le~Chen, Tom Yotam, Niranjan Hasabnis,
  and Gal Oren.
\newblock Paracodex: A profiling-guided autonomous coding agent for reliable
  parallel code generation and translation, 2026.
\newblock URL \url{https://arxiv.org/abs/2601.04327}.

\bibitem[Ke et~al.(2025)Ke, Zhang, Wang, Ding, Li, Wen, Zhang, Xu, Qin, Guo,
  Wang, Li, Guo, and Chen]{QiMengMuPa2025}
Changxin Ke, Rui Zhang, Shuo Wang, Li~Ding, Guangli Li, Yuanbo Wen, Shuoming
  Zhang, Ruiyuan Xu, Jin Qin, Jiaming Guo, Chenxi Wang, Ling Li, Qi~Guo, and
  Yunji Chen.
\newblock Qimeng-mupa: Mutual-supervised learning for sequential-to-parallel
  code translation.
\newblock \emph{arXiv preprint arxiv:2506.11153}, 2025.
\newblock URL \url{https://arxiv.org/abs/2506.11153}.

\bibitem[Konstantinidis and Cotronis(2017)]{mixbench2017}
Elias Konstantinidis and Yiannis Cotronis.
\newblock A quantitative roofline model for gpu kernel performance estimation
  using micro-benchmarks and hardware metric profiling.
\newblock \emph{Journal of Parallel and Distributed Computing}, 107:\penalty0
  37--56, 2017.
\newblock ISSN 0743-7315.
\newblock \doi{https://doi.org/10.1016/j.jpdc.2017.04.002}.
\newblock URL
  \url{https://www.sciencedirect.com/science/article/pii/S0743731517301247}.

\bibitem[Liang et~al.(2025)Liang, Garg, and Moghaddam]{SWEbenchIllusion2025}
Shanchao Liang, Spandan Garg, and Roshanak~Zilouchian Moghaddam.
\newblock The {SWE-Bench} illusion: When state-of-the-art {LLMs} remember
  instead of reason.
\newblock arXiv preprint arXiv:2506.12286, 2025.
\newblock URL \url{https://arxiv.org/abs/2506.12286}.

\bibitem[Mahmud et~al.(2025)Mahmud, TehraniJamsaz, Phan, Chen, Capot{\u{a}},
  Willke, Ahmed, and Jannesari]{AutoParLLM2025}
Quazi~Ishtiaque Mahmud, Ali TehraniJamsaz, Hung~D Phan, Le~Chen, Mihai
  Capot{\u{a}}, Theodore~L. Willke, Nesreen~K. Ahmed, and Ali Jannesari.
\newblock {A}uto{P}ar{LLM}: {GNN}-guided context generation for zero-shot code
  parallelization using {LLM}s.
\newblock In Luis Chiruzzo, Alan Ritter, and Lu~Wang, editors,
  \emph{Proceedings of the 2025 Conference of the Nations of the Americas
  Chapter of the Association for Computational Linguistics: Human Language
  Technologies (Volume 1: Long Papers)}, pages 11821--11841, Albuquerque, New
  Mexico, April 2025. Association for Computational Linguistics.
\newblock ISBN 979-8-89176-189-6.
\newblock \doi{10.18653/v1/2025.naacl-long.593}.
\newblock URL \url{https://aclanthology.org/2025.naacl-long.593/}.

\bibitem[Mi{\v{s}}i{\'c} and Dodovi{\'c}(2024)]{OpenMPAssessment2024}
Marko Mi{\v{s}}i{\'c} and Matija Dodovi{\'c}.
\newblock An assessment of large language models for {OpenMP}-based code
  parallelization: A user perspective.
\newblock \emph{Journal of Big Data}, 11:\penalty0 161, 2024.
\newblock \doi{10.1186/s40537-024-01019-z}.
\newblock URL \url{https://doi.org/10.1186/s40537-024-01019-z}.

\bibitem[Mitchell et~al.(2019)Mitchell, Wu, Zaldivar, Barnes, Vasserman,
  Hutchinson, Spitzer, Raji, and Gebru]{ModelCards2019}
Margaret Mitchell, Simone Wu, Andrew Zaldivar, Parker Barnes, Lucy Vasserman,
  Ben Hutchinson, Elena Spitzer, Inioluwa~Deborah Raji, and Timnit Gebru.
\newblock Model cards for model reporting.
\newblock In \emph{Proceedings of the Conference on Fairness, Accountability,
  and Transparency}, FAT* '19, page 220–229, New York, NY, USA, 2019.
  Association for Computing Machinery.
\newblock ISBN 9781450361255.
\newblock \doi{10.1145/3287560.3287596}.
\newblock URL \url{https://doi.org/10.1145/3287560.3287596}.

\bibitem[Nichols et~al.(2024{\natexlab{a}})Nichols, Davis, Xie, Rajaram, and
  Bhatele]{ParEval2024}
Daniel Nichols, Joshua~H. Davis, Zhaojun Xie, Arjun Rajaram, and Abhinav
  Bhatele.
\newblock Can large language models write parallel code?
\newblock In \emph{Proceedings of the 33rd International Symposium on
  High-Performance Parallel and Distributed Computing}, HPDC '24, page
  281–294, New York, NY, USA, 2024{\natexlab{a}}. Association for Computing
  Machinery.
\newblock ISBN 9798400704130.
\newblock \doi{10.1145/3625549.3658689}.
\newblock URL \url{https://doi.org/10.1145/3625549.3658689}.

\bibitem[Nichols et~al.(2024{\natexlab{b}})Nichols, Marathe, Menon, Gamblin,
  and Bhatele]{HPCCoder2024}
Daniel Nichols, Aniruddha Marathe, Harshitha Menon, Todd Gamblin, and Abhinav
  Bhatele.
\newblock Hpc-coder: Modeling parallel programs using large language models.
\newblock In \emph{ISC High Performance 2024 Research Paper Proceedings (39th
  International Conference)}, pages 1--12, 2024{\natexlab{b}}.
\newblock \doi{10.23919/ISC.2024.10528929}.

\bibitem[{OpenAI}(2026)]{GPT53Codex2026}
{OpenAI}.
\newblock {GPT}-5.3-codex system card, 2026.
\newblock URL \url{https://openai.com/index/gpt-5-3-codex-system-card/}.
\newblock Accessed 2026-05-01.

\bibitem[Ouyang et~al.(2025)Ouyang, Guo, Arora, Zhang, Hu, Ré, and
  Mirhoseini]{KernelBench2025}
Anne Ouyang, Simon Guo, Simran Arora, Alex~L. Zhang, William Hu, Christopher
  Ré, and Azalia Mirhoseini.
\newblock Kernelbench: Can llms write efficient gpu kernels?
\newblock 2025.
\newblock URL \url{https://arxiv.org/abs/2502.10517}.

\bibitem[Pophale et~al.(2025)Pophale, Jin, and Teranishi]{ChatPORT2025}
Swaroop Pophale, Zheming Jin, and Keita Teranishi.
\newblock Chatport: Fine-tuned llm for easy code {PORT}ing.
\newblock In \emph{OpenMP: Balancing Productivity and Performance Portability:
  21st International Workshop on OpenMP, IWOMP 2025, Charlotte, NC, USA,
  October 1-3, 2025 Proceedings}, page 197–211, Berlin, Heidelberg, 2025.
  Springer-Verlag.
\newblock ISBN 978-3-032-06342-7.
\newblock \doi{10.1007/978-3-032-06343-4_13}.
\newblock URL \url{https://doi.org/10.1007/978-3-032-06343-4_13}.

\bibitem[Rao et~al.(2025)Rao, Zhao, Zhu, Xiao, Wang, and Wang]{CodeMorph2025}
Hongzhou Rao, Yanjie Zhao, Wenjie Zhu, Ling Xiao, Meizhen Wang, and Haoyu Wang.
\newblock Codemorph: Mitigating data leakage in large language model
  assessment.
\newblock In \emph{2025 IEEE/ACM 47th International Conference on Software
  Engineering: Companion Proceedings (ICSE-Companion)}, page 267–278. IEEE
  Press, 2025.
\newblock \doi{10.1109/ICSE-Companion66252.2025.00081}.
\newblock URL \url{https://doi.org/10.1109/ICSE-Companion66252.2025.00081}.

\bibitem[Roziere et~al.(2020)Roziere, Lachaux, Chanussot, and
  Lample]{TransCoder2020}
Baptiste Roziere, Marie-Anne Lachaux, Lowik Chanussot, and Guillaume Lample.
\newblock Unsupervised translation of programming languages.
\newblock In \emph{Proceedings of the 34th International Conference on Neural
  Information Processing Systems}, NIPS '20, Red Hook, NY, USA, 2020. Curran
  Associates Inc.
\newblock ISBN 9781713829546.

\bibitem[Schneider et~al.(2023)Schneider, Kadosh, Hasabnis, Mattson, Pinter,
  and Oren]{MPIRICAL2023}
Nadav Schneider, Tal Kadosh, Niranjan Hasabnis, Timothy Mattson, Yuval Pinter,
  and Gal Oren.
\newblock Mpi-rical: Data-driven mpi distributed parallelism assistance with
  transformers.
\newblock In \emph{Proceedings of the SC '23 Workshops of the International
  Conference on High Performance Computing, Network, Storage, and Analysis},
  SC-W '23, page 2–10, New York, NY, USA, 2023. Association for Computing
  Machinery.
\newblock ISBN 9798400707858.
\newblock \doi{10.1145/3624062.3624063}.
\newblock URL \url{https://doi.org/10.1145/3624062.3624063}.

\bibitem[Schneider et~al.(2024)Schneider, Hasabnis, Vo, Kadosh, Krien, Capota,
  Tamir, Willke, Ahmed, Pinter, Mattson, and Oren]{MPIrigen2024}
Nadav Schneider, Niranjan Hasabnis, Vy~A. Vo, Tal Kadosh, Neva Krien, Mihai
  Capota, Guy Tamir, Theodore~L. Willke, Nesreen Ahmed, Yuval Pinter, Timothy
  Mattson, and Gal Oren.
\newblock Mpirigen: Mpi code generation through domain-specific language
  models.
\newblock In \emph{Proceedings of the 2024 Workshop on AI For Systems}, AI4Sys
  '24, page 1–6, New York, NY, USA, 2024. Association for Computing
  Machinery.
\newblock ISBN 9798400706523.
\newblock \doi{10.1145/3660605.3660944}.
\newblock URL \url{https://doi.org/10.1145/3660605.3660944}.

\bibitem[TehraniJamsaz et~al.(2024)TehraniJamsaz, Bhattacharjee, Chen, Ahmed,
  Yazdanbakhsh, and Jannesari]{CodeRosetta2024}
Ali TehraniJamsaz, Arijit Bhattacharjee, Le~Chen, Nesreen~K. Ahmed, Amir
  Yazdanbakhsh, and Ali Jannesari.
\newblock Coderosetta: Pushing the boundaries of unsupervised code translation
  for parallel programming.
\newblock In A.~Globerson, L.~Mackey, D.~Belgrave, A.~Fan, U.~Paquet,
  J.~Tomczak, and C.~Zhang, editors, \emph{Advances in Neural Information
  Processing Systems}, volume~37, pages 100965--100999. Curran Associates,
  Inc., 2024.
\newblock \doi{10.52202/079017-3202}.
\newblock URL
  \url{https://proceedings.neurips.cc/paper_files/paper/2024/file/b6edb87876bec4ac2260bffa083cb992-Paper-Conference.pdf}.

\bibitem[Tramm et~al.(2014)Tramm, Siegel, Islam, and Schulz]{XSBench2014}
John~R Tramm, Andrew~R Siegel, Tanzima Islam, and Martin Schulz.
\newblock {XSBench} - the development and verification of a performance
  abstraction for {M}onte {C}arlo reactor analysis.
\newblock In \emph{{PHYSOR} 2014 - The Role of Reactor Physics toward a
  Sustainable Future}, Kyoto, 2014.
\newblock URL \url{https://www.mcs.anl.gov/papers/P5064-0114.pdf}.

\bibitem[Tramm et~al.(2015)Tramm, Siegel, Forget, and Josey]{RSBench2015}
John~R. Tramm, Andrew~R. Siegel, Benoit Forget, and Colin Josey.
\newblock Performance analysis of a reduced data movement algorithm for neutron
  cross section data in monte carlo simulations.
\newblock In Stefano Markidis and Erwin Laure, editors, \emph{Solving Software
  Challenges for Exascale}, pages 39--56, Cham, 2015. Springer International
  Publishing.
\newblock ISBN 978-3-319-15976-8.

\bibitem[Valero-Lara et~al.(2026)Valero-Lara, Young, Naughton~III, Engelmann,
  Geist, Vetter, Teranishi, and Godoy]{ChatMPI2026}
Pedro Valero-Lara, Aaron Young, Thomas Naughton~III, Christian Engelmann,
  Al~Geist, Jeffrey~S. Vetter, Keita Teranishi, and William~F. Godoy.
\newblock Chatmpi: Llm-driven mpi code generation for hpc workloads.
\newblock In \emph{Proceedings of the Supercomputing Asia and International
  Conference on High Performance Computing in Asia Pacific Region}, SCA/HPCAsia
  '26, page 19–30, New York, NY, USA, 2026. Association for Computing
  Machinery.
\newblock ISBN 9798400720673.
\newblock \doi{10.1145/3773656.3773659}.
\newblock URL \url{https://doi.org/10.1145/3773656.3773659}.

\bibitem[Wang et~al.(2026)Wang, Wang, Wang, Guo, Chen, Grundy, Liu, Ma, Mao,
  Zhang, and Zheng]{Wang2026RepoTransBench}
Yanli Wang, Yanlin Wang, Suiquan Wang, Daya Guo, Jiachi Chen, John Grundy,
  Xilin Liu, Yuchi Ma, Mingzhi Mao, Hongyu Zhang, and Zibin Zheng.
\newblock Repotransbench: A real-world multilingual benchmark for
  repository-level code translation.
\newblock \emph{IEEE Transactions on Software Engineering}, 52\penalty0
  (2):\penalty0 675--690, 2026.
\newblock \doi{10.1109/TSE.2025.3645056}.

\bibitem[Zhang et~al.(2025)Zhang, Chen, Zhong, Peng, Wang, and
  Shang]{MemorizeOrGeneralize2025}
Lizhe Zhang, Wentao Chen, Li~Zhong, Letian Peng, Zilong Wang, and Jingbo Shang.
\newblock Memorize or generalize? evaluating llm code generation with code
  rewriting, 2025.
\newblock URL \url{https://arxiv.org/abs/2503.02296}.

\end{thebibliography}

\appendix




\newpage

\clearpage
\phantomsection
\section*{Appendix Contents}
\label{sec:appendix-contents}

This appendix provides supporting material for reproducing and interpreting \parbench{}. Appendix~A expands the related-work positioning. Appendices~B--D document the framework, API-selection rationale, and benchmark curation process. 
Appendix~E groups the augmentation protocol and extended evaluation results, and Appendix~F reproduces the translation prompt. Appendices~G--H cover reproducibility details and the benchmark's measurement scope. Appendix~I describes artifact availability. Appendix~J holds the remaining dense reference figures and full per-kernel tables.

\vspace{0.75em}

\begin{center}
\small
\renewcommand{\arraystretch}{1.15}
\setlength{\tabcolsep}{5pt}

\newlength{\appendixsectionwidth}
\settowidth{\appendixsectionwidth}{Additional Tables and Figures\quad}

\begin{tabularx}{0.98\textwidth}{@{}
>{\bfseries}p{0.035\textwidth}
>{\RaggedRight\arraybackslash}p{\appendixsectionwidth}
X
@{}}
\toprule
\textbf{} & \textbf{Section} & \textbf{Contents} \\
\midrule

\hyperref[sec:appendix-d]{A}
& \hyperref[sec:appendix-d]{Related Work}
& Extended comparison with general code benchmarks, repository-level translation benchmarks,
OpenMP/MPI corpora, LLM-based parallel-code systems, and robustness or efficiency benchmarks.
Includes Tables~\ref{tab:related-work-a}--\ref{tab:related-work-d}. \\

\hyperref[sec:appendix-a]{B}
& \hyperref[sec:appendix-a]{Framework and Evaluation Details}
& Specification schema, augmentation levels, prompt anonymization, response extraction,
\knownfail{} exclusion policy, sampling configuration, and metric definitions. \\

\hyperref[sec:appendix-b]{C}
& \hyperref[sec:appendix-b]{API Selection Rationale}
& Full API co-occurrence analysis, kernel-level API-pair coverage, translation-difficulty taxonomy,
and rationale for including CUDA, OpenMP, OpenCL, and OpenMP Target while excluding HIP,
SYCL, and OpenACC from the main evaluation. \\

\hyperref[sec:appendix-c]{D}
& \hyperref[sec:appendix-c]{Benchmark Kernel Survey}
& Repository inventory, quality-tier classification, HeCBench selection funnel, Rodinia inventory,
XSBench/RSBench/mixbench profiles, and exclusion log for benchmarks and kernels. \\

\hyperref[sec:appendix-e]{E}
& \hyperref[sec:appendix-e]{Augmentation Protocol and Extended Evaluation Results}
& Augmentation transform frequencies, qualitative failure case studies, extended pass-rate analyses,
direction-level results, and additional statistical details. \\

\hyperref[sec:appendix-g]{F}
& \hyperref[sec:appendix-g]{Translation Prompt Template}
& Full system and user prompt templates used for kernel-centric translation, including output-format
requirements and anonymization details. \\

\hyperref[sec:appendix-j]{G}
& \hyperref[sec:appendix-j]{Evaluation Cost and Reproducibility}
& Token usage, task counts, wall-clock evaluation time, and practical budgeting notes for reproducing
the three model campaigns. \\

\hyperref[sec:appendix-k]{H}
& \hyperref[sec:appendix-k]{Evaluation Card}
& Structured benchmark card defining what \parbench{} measures, what it does not measure,
valid and invalid claims, assumptions, user guidance, and recommended reporting protocol. \\

\hyperref[sec:artifact-availability]{I}
& \hyperref[sec:artifact-availability]{Artifact Availability}
& Description of the public artifact release, including curated specifications,
the build-run-verify harness, augmentation and evaluation pipelines, per-record result files,
licensing, and archival plan. \\

\hyperref[sec:appendix-f]{J}
& \hyperref[sec:appendix-f]{Additional Tables and Figures}
& Remaining dense reference material: direction-level pass@$k$ figures, cross-suite comparisons, and full per-kernel tables for all three model campaigns. \\

\bottomrule
\end{tabularx}
\end{center}


\newpage
\vspace{1.0em}
\phantomsection
\section*{Figures and Tables Roadmap}
\label{sec:appendix-float-roadmap}

The appendix contains many figures and tables because each group serves a different
evidentiary role. The roadmap below explains how to read the additional floats:
some position \parbench{} against prior work, some justify benchmark design choices,
some document corpus construction, some support the empirical findings, and others
define reproducibility, cost, artifact availability, and reporting scope.

\vspace{0.5em}

\begin{center}
\tiny
\renewcommand{\arraystretch}{1.18}
\setlength{\tabcolsep}{3pt}

\begin{tabularx}{0.98\textwidth}{@{}
>{\RaggedRight\arraybackslash}p{0.20\textwidth}
>{\RaggedRight\arraybackslash}p{0.30\textwidth}
X
@{}} 
\toprule
\textbf{Evidence block} & \textbf{Figures and tables} & \textbf{Why they are included} \\
\midrule

Positioning against prior work
&
Appendix~A:
Tables~\ref{tab:related-work-a}--\ref{tab:related-work-d}
&
These tables explain why \parbench{} is not just another code-generation,
parallelization, or repository-level benchmark. They separate prior work by task,
granularity, verification style, corpus status, and robustness testing, making clear
which evaluation gap \parbench{} is designed to fill. \\

Benchmark mechanics and reproducibility
&
Appendix~B:
Listing~\ref{lst:bfs_spec_appendix};
Table~\ref{tab:augmentation-levels}
&
These items make the benchmark executable and reproducible. The listing shows the
declarative JSON specification structure, while the table defines the L0--L4
augmentation levels used throughout the robustness analysis. \\

API-selection evidence
&
Appendix~C:
Figures~\ref{fig:api-heatmap}--\ref{fig:kernel-network};
Figure~\ref{fig:repo-vs-kernel};
Table~\ref{tab:api-characteristics}
&
These floats justify the choice of CUDA, OpenMP, OpenCL, and OpenMP Target. They
show both repository-level and kernel-level API availability, expose why repository
counts alone are misleading, and summarize the programming-model differences that
make the selected translation directions technically meaningful. \\

Corpus construction and filtering
&
Appendix~D:
Figures~\ref{fig:benchmark-types}--\ref{fig:verification-network};
Table~\ref{tab:benchmark-characterization}
&
These floats document how the final benchmark corpus was selected from a much larger
HPC benchmark landscape. They show the repository inventory, API breadth, candidate
ranking, HeCBench funnel, verification methods, and final corpus characteristics,
supporting the claim that the corpus is systematic rather than cherry-picked. \\

Augmentation and memorization probes
&
Appendix~E:
Figure~\ref{fig:transform-freq};
Table~\ref{tab:aug-balanced};
Table~\ref{tab:augmentation-rates};
Figures~\ref{fig:augmentation}--\ref{fig:aug-heatmap}
&
These floats explain the source-perturbation experiment. They show which AST
transforms were applied, how often they appear, and whether model success persists
under L1--L4 perturbations. Their role is to support the robustness and
surface-form-memorization analysis. \\

Prompt and anonymization protocol
&
Appendix~F:
Listings~\ref{lst:system-msg} and~\ref{lst:user-msg}
&
These listings document the actual translation prompt templates (system and user messages). They are included so that future users can distinguish model capability from prompt-format differences. \\

Overall evaluation setup
&
Appendix~G:
Tables~\ref{tab:model-config} and~\ref{tab:hardware}
&
These tables record the model, provider, reasoning-mode, hardware, compiler, and
software configuration details needed to interpret the empirical results. They also
make explicit where sampling conditions differ across providers. \\

Pass-rate, direction analysis, and sampling behavior
&
Appendix~E:
Table~\ref{tab:overall-pass};
Table~\ref{tab:direction-rates};
Table~\ref{tab:pass-at-k-main};
Table~\ref{tab:pass-at-k};
Appendix~J:
Figures~\ref{fig:pass-at-k-qwen}--\ref{fig:pass-at-k-codex}
&
These floats support the record-level pass-rate summary, direction-level analysis, and pass@$k$ analysis. They distinguish single-sample success from repeated-sampling success and show whether failures are stochastic or systematic across directions and models. \\

Kernel-level difficulty
&
Appendix~E:
Table~\ref{tab:per-kernel-main};
Main body: Figure~\ref{fig:kernel-heatmap-unified}; Appendix~J: Tables~\ref{tab:per-kernel-full}, \ref{tab:per-kernel-full-gpt54}, and~\ref{tab:per-kernel-full-codex}
&
These floats show that translation difficulty is highly uneven across kernels.
They are included to prevent the aggregate pass rates from hiding hard kernels,
easy kernels, multi-file effects, and direction-specific failure patterns. \\

Failure taxonomy and direction asymmetry
&
Appendix~E:
Table~\ref{tab:lavamd-errors};
Table~\ref{tab:stats-summary}
&
These floats diagnose where translation breaks. They separate build, runtime,
verification, and extraction failures, and they support the paper's claims about
API-surface adaptation, direction asymmetry, and multi-file or host-scaffolding
difficulty. \\

Cross-suite generality
&
Appendix~J:
Figures~\ref{fig:cross-suite-qwen}--\ref{fig:cross-suite-codex}
&
These figures show whether results are concentrated in one benchmark suite or
persist across Rodinia, HeCBench, XSBench, RSBench, and mixbench. Their role is to
support claims about suite-level generality and domain variation. \\

Repository-level versus kernel-level opportunity
&
Appendix~C:
Figure~\ref{fig:repo-vs-kernel}
&
This figure explains one of the benchmark-design motivations: repository-level API
co-occurrence substantially underestimates the number of independent kernel-level
translation opportunities. It supports the kernel-centric design choice. \\

Evaluation cost and practical reproducibility
&
Appendix~G:
Table~\ref{tab:eval-cost}
&
This table reports the practical cost of running the evaluation campaign. It helps
future users estimate token usage, task volume, wall-clock time, and budget before
reproducing or extending \parbench{}. \\

Benchmark scope and reporting protocol
&
Appendix~H:
Tables~\ref{tab:eval-card-summary},
\ref{tab:reporting-required}, and~\ref{tab:reporting-recommended}
&
These tables define what \parbench{} measures, what it does not measure, which
claims are valid, and what future papers should report. They are included to make
the benchmark harder to misuse or overclaim. \\

\bottomrule
\end{tabularx}
\end{center}

\section{Related Work}
\label{sec:appendix-d}

The main text positions \parbench{} through the evaluation gap it addresses. 
This appendix expands that comparison in Tables~\ref{tab:related-work-a}--\ref{tab:related-work-d}, which organize prior work by evaluation level and role: general code and repository-level benchmarks, OpenMP/MPI and local parallel-code corpora, LLM-based parallel-code translation and porting systems, and efficiency or robustness benchmarks. 
The comparison separates five questions that are often conflated in prior work: whether the task is generation, parallelization, porting, or translation; what unit of code is evaluated; whether the corpus is a reusable benchmark or a paper-specific local evaluation set; whether correctness is checked by execution; and whether robustness to surface-form variation is measured. 
This distinction is important because recent LLM-for-parallel-programming work has produced many useful local corpora, benchmark subsets, and task-specific evaluation protocols, but comparatively few reusable executable frameworks for controlled cross-API translation of existing parallel kernels.

\begin{table*}[p]
\centering
\caption{
Representative related work, Part I-a. General code benchmarks and repository-level translation benchmarks.
}
\label{tab:related-work-a}
\scriptsize
\setlength{\tabcolsep}{3.2pt}
\renewcommand{\arraystretch}{1.05}
\begin{tabularx}{\textwidth}{@{}
>{\RaggedRight\arraybackslash}p{0.19\textwidth}
>{\RaggedRight\arraybackslash}p{0.24\textwidth}
>{\RaggedRight\arraybackslash}p{0.25\textwidth}
>{\RaggedRight\arraybackslash}X
@{}}
\toprule
Work & Primary task / granularity & Corpus and verification basis & Relation to \parbench{} \\
\midrule

HumanEval~\cite{HumanEval2021}
& General code generation; function level
& Public benchmark with unit tests
& Establishes execution-based evaluation, but does not exercise parallel API semantics. \\

TransCoder~\cite{TransCoder2020}
& General source-to-source translation; function level
& Public multilingual corpus with unit tests
& Provides a code-translation precedent, but not for HPC or parallel APIs. \\

SWE-bench~\cite{SWEbench2024}
& Repository issue repair; repository level
& Public GitHub-issue benchmark with repository tests
& Demonstrates executable repository evaluation, but is mostly non-HPC and non-parallel. \\

RepoTransBench~\cite{Wang2026RepoTransBench}
& Repository-level code translation
& Public repository benchmark with build/test signals
& Shows repository translation complexity, but confounds kernel translation with scaffolding. \\

AlphaTrans~\cite{AlphaTrans2025}
& Repository translation and validation
& Repository-scale translation tasks with validation/tests
& Adjacent repository-level translation work, but not focused on parallel API semantics. \\

\bottomrule
\end{tabularx}
\end{table*}

\begin{table*}[p]
\centering
\caption{
Representative related work, Part I-b. OpenMP/MPI/local parallel-code corpora and generation benchmarks.
}
\label{tab:related-work-b}
\scriptsize
\setlength{\tabcolsep}{3.2pt}
\renewcommand{\arraystretch}{1.05}
\begin{tabularx}{\textwidth}{@{}
>{\RaggedRight\arraybackslash}p{0.19\textwidth}
>{\RaggedRight\arraybackslash}p{0.24\textwidth}
>{\RaggedRight\arraybackslash}p{0.25\textwidth}
>{\RaggedRight\arraybackslash}X
@{}}
\toprule
Work & Primary task / granularity & Corpus and verification basis & Relation to \parbench{} \\
\midrule

Learning to Parallelize~\cite{LearningToParallelize2023}
& OpenMP parallelization prediction; loop/region level
& Curated OpenMP training and evaluation corpus
& Shows local corpus construction for OpenMP assistance, not executable cross-API translation. \\

MPI-RICAL~\cite{MPIRICAL2023}
& MPI assistance; function/region level
& Curated MPI-focused corpus with recommendation metrics
& Relevant as MPI-specific parallel-code assistance using a local dataset. \\

OpenMP Graph Transformer~\cite{OMPify2023}
& OpenMP parallelization advice; loop/region level
& Curated OpenMP corpus with recommendation/classification metrics
& Relevant to pragma advice, but not translation evaluation. \\

OMPGPT~\cite{OMPGPT2024}
& OpenMP-oriented code modeling; function/task level
& OpenMP-oriented model corpus and generation metrics
& Domain-specific modeling work, but not reusable translation infrastructure. \\

MPIrigen~\cite{MPIrigen2024}
& MPI code generation; function/task level
& MPI-specific curated corpus with generation checks
& Generation-oriented MPI work, not API-to-API kernel translation. \\

MonoCoder~\cite{MonoCoder2024}
& HPC code modeling; task/function level
& HPC-oriented model corpus and task metrics
& Relevant HPC specialization work, but method/corpus-specific. \\

PragFormer~\cite{PragFormer2025}
& Parallel-code classification; loop/region level
& Curated classification corpus
& Relevant representation work, but not executable translation. \\

OMPar~\cite{OMPar2024}
& AI-driven OpenMP source-to-source parallelization; program/loop level
& Paper-specific OpenMP evaluation set
& Automatic parallelization rather than translation of existing parallel kernels. \\

AutoParLLM~\cite{AutoParLLM2025}
& Zero-shot code parallelization; loop/function level
& Local parallelization evaluation set
& Measures introducing parallelism, not preserving existing parallel semantics across APIs. \\

OpenMP Assessment~\cite{OpenMPAssessment2024}
& LLMs for OpenMP parallelization; snippet/program level
& Paper-specific assessment set
& Shows the need for OpenMP-specific evaluation, but not cross-API translation. \\

ParEval~\cite{ParEval2024}
& Parallel code generation with limited translation settings; task level
& Public benchmark with correctness checks
& Includes generation and some translation settings, but not \parbench{}'s fixed executable kernel corpus or broader cross-API direction coverage. \\

PCEBench~\cite{PCEBench2025}
& Multi-dimensional parallel code generation; task level
& Public benchmark with correctness/task metrics
& Complements \parbench{} by evaluating generation rather than existing-kernel translation. \\

\bottomrule
\end{tabularx}
\end{table*}

\begin{table*}[p]
\centering
\caption{
Representative related work, Part II. LLM-based parallel-code translation, porting, and HPC-code modeling systems.
}
\label{tab:related-work-c}
\scriptsize
\setlength{\tabcolsep}{3.2pt}
\renewcommand{\arraystretch}{1.05}
\begin{tabularx}{\textwidth}{@{}
>{\RaggedRight\arraybackslash}p{0.19\textwidth}
>{\RaggedRight\arraybackslash}p{0.24\textwidth}
>{\RaggedRight\arraybackslash}p{0.25\textwidth}
>{\RaggedRight\arraybackslash}X
@{}}
\toprule
Work & Primary task / granularity & Corpus and verification basis & Relation to \parbench{} \\
\midrule

LASSI~\cite{LASSI2024}
& Agentic parallel-code translation; kernel level
& Curated HeCBench subset with build-run-verify
& Closest in verification style, but smaller in scope and focused on iterative repair. \\

CodeRosetta~\cite{CodeRosetta2024}
& Parallel code translation and fine-tuning; function/kernel level
& Model-specific C++/CUDA corpus with compile/similarity/local checks
& Relevant translation work, but lacks fixed multi-API executable specifications. \\

HPC-Coder~\cite{HPCCoder2024}
& HPC code modeling; task/function level
& HPC-oriented corpus and model-evaluation metrics
& Domain-specific HPC modeling, not benchmark-infrastructure focused. \\

HPC-Coder-V2~\cite{HPCCoderV2}
& Code LLMs across low-resource parallel languages; task level
& Parallel-language tasks with pass@$k$ and task checks
& Relevant to parallel-code LLMs, but not controlled executable kernel translation. \\

ParEval-Repo~\cite{ParEvalRepo2025}
& Repository-level HPC translation
& Public HPC repository benchmark with build and functional tests
& Motivates \parbench{} by showing that repository integration can dominate failures. \\

UniPar~\cite{UniPar2025}
& Parallel and accelerated code translation; task/program level
& System-specific evaluation corpus with local correctness checks
& Relevant multi-paradigm translation, but not primarily a reusable benchmark substrate. \\

QiMeng-MuPa~\cite{QiMengMuPa2025}
& Sequential-to-parallel translation; task/program level
& Curated training/evaluation corpus with local correctness checks
& Related parallelization/translation work, but different from API-to-API translation of existing kernels. \\

ChatPORT~\cite{ChatPORT2025}
& LLM-assisted code porting; kernel/program level
& Local porting evaluation with build/run checks
& Relevant porting work, but method-specific. \\

ChatPORT CUDA-to-SYCL~\cite{ChatPORTSYCL2025}
& CUDA-to-SYCL translation; kernel level
& Local CUDA-to-SYCL evaluation with build/run checks
& Directly relevant API migration, but narrower in direction and scope. \\

ChatMPI~\cite{ChatMPI2026}
& MPI code generation; function/workload level
& MPI-focused local evaluation with functional checks
& Distributed-parallel generation, not cross-API kernel translation. \\

Large LLM Evaluation for HPC~\cite{LargeLLMEvalHPC2024}
& Evaluating LLMs for HPC software development; task/program level
& HPC-oriented evaluation suite with task-specific checks
& Broad HPC LLM context, but not controlled cross-API translation. \\

VibeCodeHPC~\cite{VibeCodeHPC2026}
& Agentic HPC code generation/tuning; application/kernel level
& Small local HPC benchmark set with build/run/performance signals
& Agentic generation and optimization, not existing-kernel API translation. \\

ParaCodex~\cite{ParaCodex2026}
& Profiling-guided agentic parallel generation/translation; kernel/program level
& Curated suite mixture with build-run-verify and profiling feedback
& Directly relevant method-side work; \parbench{} can evaluate such systems under fixed specs. \\

\bottomrule
\end{tabularx}
\end{table*}

\begin{table*}[p]
\centering
\caption{
Representative related work, Part III. Efficiency, robustness, and benchmark-validity studies relevant to \parbench{}.
}
\label{tab:related-work-d}
\scriptsize
\setlength{\tabcolsep}{3.2pt}
\renewcommand{\arraystretch}{1.05}
\begin{tabularx}{\textwidth}{@{}
>{\RaggedRight\arraybackslash}p{0.19\textwidth}
>{\RaggedRight\arraybackslash}p{0.24\textwidth}
>{\RaggedRight\arraybackslash}p{0.25\textwidth}
>{\RaggedRight\arraybackslash}X
@{}}
\toprule
Work & Primary task / granularity & Corpus and verification basis & Relation to \parbench{} \\
\midrule

TRACE~\cite{TRACE2026}
& Efficiency of LLM-based code translation; function/task level
& Efficiency benchmark with correctness and runtime
& Complementary performance axis; \parbench{} focuses first on correctness. \\

KernelBench~\cite{KernelBench2025}
& Efficient GPU-kernel generation; kernel level
& GPU-kernel benchmark with correctness and speed
& Accelerator-kernel generation, not translation of existing parallel APIs. \\

Mercury~\cite{Mercury2024}
& Efficiency of LLM code synthesis; function/task level
& Efficiency benchmark with runtime/resource metrics
& Evaluates efficiency rather than parallel API semantic preservation. \\

CodeMorph~\cite{CodeMorph2025}
& Robustness and leakage via code transformation; function level
& Existing benchmarks with transformed variants and unit tests
& Provides the source-perturbation principle adapted by \parbench{} to HPC translation. \\

Semantic Code Rewrite~\cite{MemorizeOrGeneralize2025}
& Memorization versus generalization under code rewriting; function level
& Rewritten code benchmarks with unit tests
& Motivates source rewriting as a contamination and robustness probe. \\

SWE-bench Illusion~\cite{SWEbenchIllusion2025}
& Benchmark memorization analysis; repository level
& SWE-bench-style evaluation
& Supports robustness checks for public code benchmarks. \\

\textbf{\parbench{}}
& \textbf{Cross-API translation of existing parallel kernels; kernel level}
& \textbf{Reusable executable specs with conjunctive build-run-verify}
& \textbf{Fixes infrastructure, isolates parallel API translation, and adds L0--L4 AST augmentation.} \\

\bottomrule
\end{tabularx}
\end{table*}

\paragraph{Local parallel-code corpora and paper-specific benchmarks.}
A recurring pattern in recent LLM-assisted parallel-programming work is that evaluation is often built around a corpus or benchmark subset introduced for the specific paper. Earlier OpenMP and MPI work constructed curated datasets for source-to-source parallelization, pragma recommendation, parallel-region classification, MPI API assistance, and domain-specific model training~\cite{LearningToParallelize2023,MPIRICAL2023,OMPify2023,OMPGPT2024,MPIrigen2024,MonoCoder2024,PragFormer2025,CuratedOpenMPDataset2026}. Prior corpus studies also show that OpenMP usage itself has been systematically characterized at scale~\cite{HPCorpus2023}, reinforcing that parallel-programming corpora are valuable but do not by themselves define a controlled LLM translation benchmark. Later LLM-based systems for automatic parallelization, porting, and HPC code modeling continued this pattern, evaluating on paper-specific mixtures of OpenMP, MPI, CUDA, mini-applications, or benchmark-suite kernels~\cite{OMPar2024,AutoParLLM2025,OpenMPAssessment2024,HPCCoder2024,HPCCoderV2,LargeLLMEvalHPC2024,ChatMPI2026,VibeCodeHPC2026}. These studies are directly relevant because they show sustained demand for parallel-code evaluation, but they do not provide a shared measurement substrate for controlled cross-API translation of existing parallel kernels.

\paragraph{Generation, parallelization, porting, and translation measure different capabilities.}
Parallel code generation benchmarks such as ParEval and PCEBench ask whether a model can synthesize a parallel program from a natural-language description, and ParEval also includes limited translation settings~\cite{ParEval2024,PCEBench2025}. These are important capabilities, but they differ from source-to-source API translation under fixed executable infrastructure. In generation, the model may choose its own decomposition, data layout, and implementation strategy; in translation, it must preserve the structure and behavior of an existing implementation while adapting thread indexing, synchronization, memory movement, host--device coordination, and API-specific launch structure. Automatic-parallelization systems such as OMPar, AutoParLLM, and related OpenMP-focused assessments measure yet another capability: identifying and introducing parallelism into serial or partially parallel code~\cite{OMPar2024,AutoParLLM2025,OpenMPAssessment2024}. Porting frameworks such as ChatPORT and CUDA-to-SYCL translation work are closer to \parbench{} in spirit, but are typically narrower in API direction, corpus scope, or evaluation protocol~\cite{ChatPORT2025,ChatPORTSYCL2025}. \parbench{} instead targets cross-API translation of already-parallel kernels, where the reported claim is satisfaction of the declared oracle for an existing implementation rather than the existence of some correct parallel solution.

\paragraph{Repository-level realism versus kernel-level attribution.}
Repository-level benchmarks expose realistic integration barriers. SWE-bench and RepoTransBench demonstrate the importance of repository-level executable evaluation in general software settings~\cite{SWEbench2024,Wang2026RepoTransBench}, while ParEval-Repo shows that full-repository HPC translation can be dominated by build-system and scaffolding failures~\cite{ParEvalRepo2025}. Those barriers matter for deployment, but they can obscure the narrower scientific question of whether the model can translate the computational kernel itself. \parbench{} occupies the intermediate granularity between single-function benchmarks and full repositories. It uses real kernels from established HPC suites, but fixes the surrounding build, run, and verification infrastructure through declarative specifications. This design preserves executable evaluation while making failure attribution sharper: a failed task is more directly tied to API adaptation, multi-file kernel coordination, or semantic preservation, rather than to missing scaffolding or repository reconstruction.

\paragraph{Agentic repair and specialized models are complementary to benchmark design.}
Several systems improve parallel-code generation or translation through fine-tuning, prompting, or iterative agentic repair. CodeRosetta and HPC-Coder-style models demonstrate the value of domain-specific corpora and model adaptation for HPC and parallel code~\cite{CodeRosetta2024,HPCCoder2024,HPCCoderV2}. LASSI evaluates an automated self-correcting pipeline for parallel scientific-code translation on a curated HeCBench subset~\cite{LASSI2024}. UniPar, QiMeng-MuPa, VibeCodeHPC, and ParaCodex similarly explore multi-paradigm generation, sequential-to-parallel transformation, agentic HPC code generation, or profiling-guided repair using local or mixed benchmark suites~\cite{UniPar2025,QiMengMuPa2025,VibeCodeHPC2026,ParaCodex2026}. These systems answer questions about method design: how far can feedback, fine-tuning, or agentic iteration push performance? \parbench{} is complementary: it provides an evaluation substrate on which such methods can be compared under a controlled kernel-centric translation protocol, including first-attempt evaluation, fixed infrastructure, and uniform failure taxonomy.

\paragraph{Verification rigor.}
Verification practices vary substantially across the literature. Some works report prediction accuracy, pragma classification, compilation success, similarity, or pass@$k$ on task-specific checks~\cite{LearningToParallelize2023,OMPify2023,CodeRosetta2024,HPCCoderV2}. Others incorporate build, run, and functional validation, especially in agentic or repository-level settings~\cite{LASSI2024,ParEvalRepo2025,ParaCodex2026}. \parbench{} adopts conjunctive build-run-verify evaluation at the kernel level: a translation must compile, execute, and satisfy all declared oracle checks. This matters because parallel translations can compile and even terminate successfully while still violating output constraints, omitting required host--device transfers, mishandling reductions, or preserving only superficial API structure. Compile-only or exit-code-only evaluation would therefore overestimate translation capability.

\paragraph{Efficiency, performance reasoning, and executable validity.}
A separate line of work evaluates whether generated or translated code is efficient rather than merely passing task oracles. TRACE, KernelBench, Mercury, and related GPU-kernel or efficiency benchmarks study runtime, kernel quality, optimization, or resource behavior~\cite{TRACE2026,KernelBench2025,Mercury2024}. These works are complementary because declared-oracle PASS and performance are separable objectives: a translation can pass the declared oracle but be slow, or run fast while still violating the intended transform. \parbench{} deliberately scopes to executable validity under declared oracles in order to isolate the translation-capability question before adding performance tuning as a second axis.

\paragraph{Robustness and benchmark validity.}
Public code benchmarks face contamination and memorization risks, and recent work shows that surface-form perturbations can reveal brittle pattern matching~\cite{SWEbenchIllusion2025,CodeMorph2025,MemorizeOrGeneralize2025}. CodeMorph applies intended semantics-preserving transformations to standard code benchmarks to diagnose leakage and robustness. \parbench{} applies the same principle to HPC translation, where contamination risk is especially plausible because suites such as Rodinia and HeCBench appear in public repositories and prior papers. Its AST-driven augmentation engine creates L0--L4 source variants using behavior-preserving transformations such as identifier renaming, arithmetic rewriting, pointer/array interchange, typedef expansion, and condition rewriting, with baseline execution used to validate the retained subset. The goal is not only to detect possible memorization, but to measure whether translation success survives controlled perturbations of the source surface form.




Appendix~\ref{sec:appendix-k} provides a structured Evaluation Card declaring \parbench{}'s measurement scope, assumptions, valid and invalid claims, and a recommended reporting protocol for future users. 

\parbench{} is evaluation infrastructure rather than a new translation method. Its contribution is to make the measurement problem sharper: kernel-centric isolation avoids repository-level build confounds, declarative specs make declared-oracle evaluation reproducible, augmentation probes robustness, and the harness exposes where translations fail. The initial results show that current LLMs have real but uneven parallel translation capability, with build adaptation, direction asymmetry, and multi-file coordination as central barriers to reliable use. The benchmark code, all 96 curated JSON specs (87 eval-eligible), and per-record result files for all three models are available in the public repository at \url{https://github.com/Scientific-Computing-Lab/ParBench} to enable independent verification and extension.

\section{Framework and Evaluation Details}
\label{sec:appendix-a}

This appendix expands the framework description in Section~\ref{sec:framework}. It records the spec structure, augmentation level definitions, and implementation details that are useful for reproducing the benchmark but too detailed for the main narrative.

\subsection{Specification Schema}
\label{sec:appendix-a-schema}

Each benchmark kernel–API is encoded by a JSON specification. The schema defines five primary field groups (with additional \texttt{implementation}, \texttt{hardware}, \texttt{metadata}, \texttt{baseline\_results}, and \texttt{performance} fields defined in the schema):

\begin{itemize}
    \item \textbf{Identity and provenance}: globally unique identifier, source suite, kernel name, target API, repository URL, and commit.
    \item \textbf{File partitioning}: prompt payload files, translation targets, support files, and verification-only files.
    \item \textbf{Build}: build system, clean/configure/build commands, environment variables, and expected executable path.
    \item \textbf{Run}: executable, input configuration, command-line arguments, timeouts, and optional environment variables.
    \item \textbf{Verification}: ordered verification strategies applied conjunctively. Five types are implemented: \texttt{exit\_code}, \texttt{stdout\_pattern}, \texttt{stdout\_exclude\_pattern}, \texttt{numeric\_comparison}, and \texttt{file\_hash}.
\end{itemize}

\begin{lstlisting}[
  language=JSON,
  float,
  floatplacement=tbp,
  numbers=none,
  label={lst:bfs_spec_appendix},
  caption={Condensed specification structure for \texttt{rodinia-bfs-cuda}.}]
{
  "identity": {
    "unique_id": "rodinia-bfs-cuda",
    "parallel_api": "cuda",
    "source_suite": "rodinia"
  },
  "files": {
    "prompt_payload": ["bfs.cu", "kernel.cu", "kernel2.cu"],
    "support_files": ["Makefile"],
    "translation_targets": ["bfs.cu", "kernel.cu", "kernel2.cu"]
  },
  "build": {
    "commands": {"build": "make CUDA_DIR=..."},
    "outputs": {"executable": "./bfs.out"}
  },
  "run": {
    "executable": "./bfs.out",
    "timeout_seconds": 300,
    "input_configurations": {
      "correctness": {
        "arguments": ["../../data/bfs/graph1MW_6.txt"]
      }
    }
  },
  "verification": {
    "strategies": [
        {"type": "stdout_pattern",
        "pattern": "Result stored in result\\.txt"},
        {"type": "exit_code", "expected": 0}
        ]
    }
}
\end{lstlisting}

\subsection{Prompt Anonymization}
\label{sec:appendix-a-anonymization}

Before prompt construction, \parbench{} applies six anonymization measures: it removes the kernel name and description, strips C/C++ comments from all files shown to the LLM, replaces source filenames with generic labels (\texttt{Source File~1}, etc.), genericizes target filenames (\texttt{translated\_0.ext}, etc.) and support-file names (\texttt{Header File~$N$}, \texttt{Code File~$N$}, \texttt{Infrastructure File~$N$}), and anonymizes kernel identifiers in build commands. These measures complement source augmentation: anonymization reduces benchmark identity leakage, while augmentation changes the source surface form. The complete prompt template is reproduced in Appendix~\ref{sec:appendix-g}.

\subsection{Response Extraction Strategy}
\label{sec:appendix-a-extraction}

Each LLM response is parsed to recover the translated source files using a multi-tier extraction strategy that progressively relaxes matching criteria. The tiers are applied in order until all expected target files are recovered:

\begin{enumerate}
    \item \textbf{Explicit filename match.} The parser searches for markdown code fences annotated with the expected target filename (e.g., \texttt{```cpp filename=translated\_0.cpp}). This is the highest-confidence tier.
    \item \textbf{Extension-based match.} If explicit filenames are absent, the parser matches code fences by file extension (e.g., \texttt{.cu}, \texttt{.cl}, \texttt{.cpp}) against the expected target file types.
    \item \textbf{Fuzzy match.} Partial filename matches and common naming variants are attempted (e.g., matching \texttt{kernel.cu} to an expected \texttt{translated\_0.cu}).
    \item \textbf{Elimination-based assignment.} When multiple code blocks remain unmatched, they are assigned to the remaining expected files by elimination order.
\end{enumerate}

If any expected target file cannot be recovered or the extracted content is empty after all tiers, the task is classified as \extractionfail{}. Across the 2{,}262 evaluation-eligible records, extraction failures are rare (1 for \qwenshort{}, 3 for \gptnew{}, 0 for \codex{}), indicating that the structured prompt format (Appendix~\ref{sec:appendix-g}) effectively guides models to produce parseable output.

\subsection{Extended Experimental Setup Details}
\label{sec:appendix-a-extended}

This section expands on the experimental setup described in Section~\ref{sec:experimental-setup}.

\paragraph{\knownfail{} Exclusion Policy.} Nine specs are classified as \knownfail{} due to toolchain incompatibilities or pre-existing runtime failures unrelated to LLM translation quality: seven Rodinia specs affected by deprecated CUDA~12 texture APIs, missing system libraries, pre-existing segmentation faults, or silent OpenCL runtime failures, and two HeCBench \texttt{omp\_target} specs with build or verification failures on the evaluation platform.
A record is excluded when \emph{either} the source or the target spec is \knownfail{}. Source-side exclusion is necessary because source baseline validity is unverified on the failing platform; target-side exclusion is necessary because the target build or verification infrastructure is broken. For \qwenshort{}, this policy excludes 82~of 708~total records, leaving 626~evaluation-eligible records (426~L0 + 200~augmentation records). The \gptnew{} and \codex{} evaluation batches pre-exclude \knownfail{} specs, so all 822~\gptnew{} and 814~\codex{} on-disk records are evaluation-eligible.

\paragraph{Sampling Configuration.}
\label{sec:sampling-config}
For \qwenshort{}, we set temperature~0.7 with top-$p$~=~1.0 (full-distribution sampling). For \gptnew{} and \codex{}, Azure API does not permit explicit temperature or top-$p$ parameters on reasoning-model deployments--these are controlled internally by the provider and cannot be overridden by the caller. Cross-model differences in pass rates may therefore partly reflect unmatched sampling conditions rather than model capability alone; this confound is imposed by provider API constraints, not benchmark design, and we report it transparently. A deterministic seed derived from SHA-256 of the concatenation \texttt{source\_spec|target\_spec|sample\_id} (pipe-delimited, truncated to 31~bits) is passed to \qwenshort{} (Together~AI) and \gptnew{} (Azure Chat Completions). The Responses API used by \codex{} does not accept a seed parameter. Seeds are recorded for auditing but do not guarantee bitwise-identical outputs: Together~AI treats seeds as best-effort for MoE routing, and Azure provides no determinism guarantee even with a fixed seed. Each sample receives a single LLM call with no iterative feedback, isolating first-attempt translation capability.

\paragraph{Canonical Evaluation and Augmentation.}
The canonical campaign evaluates all 142 unique pairs at augmentation level L0 (unmodified source code) with three samples each, producing 426 L0 records per model and 1{,}278 total. Each record passes through the full build--run--verify pipeline with conjunction semantics. Most specs verify via exit-code and stdout-pattern checks; five additionally declare numeric-comparison oracles for floating-point outputs, and two verify via exit-code and file-hash oracles instead.
For the augmentation, an L0-conditional filter is applied: a pair is included only if any of its three canonical L0 samples passes. This avoids spending budget on pairs the model cannot translate even from unmodified source. For \qwenshort{}, 50~of the 142~pairs qualify (35\%); for \gptnew{}, 99~pairs qualify (70\%); for \codex{}, 97~pairs qualify (68\%). Each qualifying pair is evaluated once ($k$~=~1) at each of levels L1 through L4.

\paragraph{Metrics Details.}
For pass@$k$, following \citet{HumanEval2021}, we use the unbiased estimator $1 - \binom{n-c}{k}/\binom{n}{k}$, where $c$ is the number of passing samples out of $n$~total. We report pass@1 (expected single-sample success rate, equivalent to $c/n$ averaged across tasks) and pass@3 (fraction of tasks where any sample passes).
Separately, we report per-record pass rates with Wilson 95\%~confidence intervals. Wilson intervals are preferred over the Wald interval for their superior coverage at extreme proportions. Since three samples per task share the same kernel and direction, the independence assumption is anti-conservative and the Wilson CI understates the true uncertainty. For augmentation analysis, we apply the Cochran--Armitage trend test where per-level sample sizes are adequate, with Cohen's~$h$ for adjacent-level effect sizes.


\section{API Selection Rationale}
\label{sec:appendix-b}

This appendix provides the complete quantitative evidence underlying \parbench{}'s selection of CUDA, OpenMP, and OpenCL as target parallel programming APIs. Section~\ref{sec:suite-selection} in the main paper summarizes the corpus selection process; here we present the expanded survey data and kernel-level analysis that informed the final selection.

\subsection{Full API Co-Occurrence Data}
\label{sec:appendix-b1}
The benchmark landscape survey initially covered 71 repositories spanning benchmark suites, mini-applications, proxy applications, libraries, full applications, and microbenchmarks across 29 distinct parallel programming APIs. Of these, 35 met the inclusion criteria for detailed analysis (Section~\ref{sec:appendix-c2}). The main paper (Section~\ref{sec:suite-selection}) reports statistics for these 35 qualifying repositories.

\begin{figure}[htbp]
\centering
\input{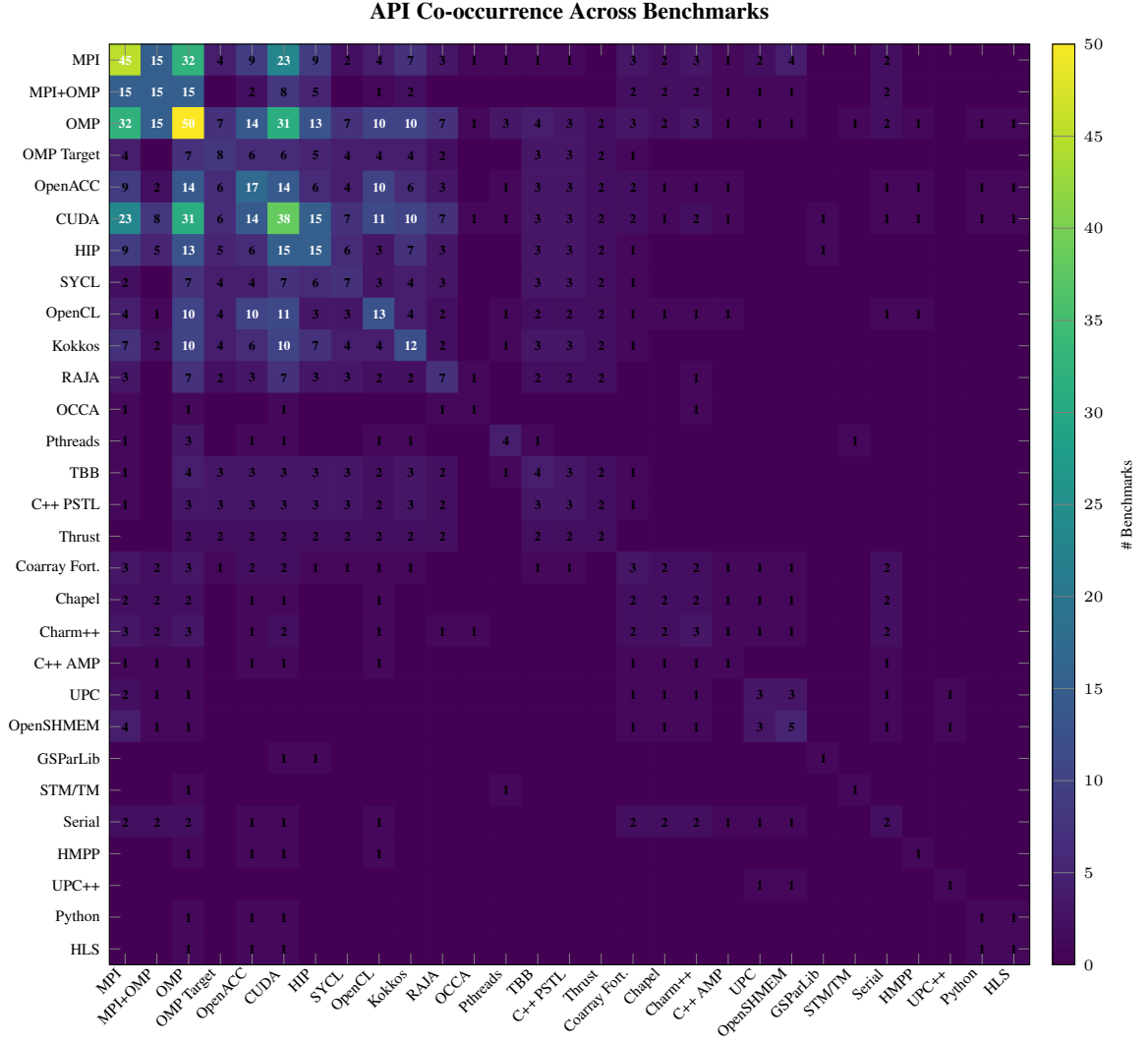}
\caption{API co-occurrence heatmap across all 71 surveyed repositories. Cell values indicate the number of repositories providing implementations in both the row and column APIs.}
\label{fig:api-heatmap}
\end{figure}

Figure~\ref{fig:api-heatmap} shows the full survey. The dominant pattern is a dense co-occurrence cluster among MPI, OpenMP, CUDA, HIP, and OpenACC, with sparser connectivity among portability-layer (Kokkos, RAJA, SYCL) and directive-based (OpenMP Target) APIs. Peripheral APIs such as Chapel, STM/TM, HLS, and Python show near-zero co-occurrence.

We classify the co-occurrence structure into three groupings:
\begin{description}
\item[Core cluster.] MPI, OpenMP, CUDA, HIP, and OpenACC form a densely interconnected cluster in the co-occurrence matrix, with the highest pairwise counts concentrated in suites with broad API coverage (HeCBench, BabelStream, RAJAPerf, CloverLeaf). OpenCL co-occurs with fewer core APIs (present in 13 of the 71 surveyed repositories, versus 38--50 for the top three) but is included in \parbench{}'s target API set because it exercises a qualitatively distinct programming model (Section~\ref{sec:appendix-b3}).
\item[Portability periphery.] SYCL, Kokkos, RAJA, OpenMP Target, and MPI+OpenMP connect to the core cluster with moderate co-occurrence, concentrated in a smaller number of suites (primarily RAJAPerf, BabelStream, and HeCBench). Pthreads, TBB, and C++ PSTL appear in 3--4 repositories each and occupy a similar peripheral role.
\item[Isolated APIs.] OCCA, Thrust, Chapel, C++ AMP, HMPP, Python, HLS, GSParLib, STM/TM, Serial, and UPC++ have maximum pairwise co-occurrence at most~2. Coarray Fortran, Charm++, UPC, and OpenSHMEM reach co-occurrence of 3--4 with MPI or with each other but remain disconnected from the GPU-centric core.
\end{description}

\begin{figure}[htbp]
    \centering
\begin{tikzpicture}
\begin{axis}[
  ybar,
  bar width=8pt,
  height=6cm,
  width=\textwidth,
  xmin=-1, xmax=29,
  ylabel={\#Benchmarks with API},
  ymin=0, ymax=58,
  ytick={0,10,20,30,40,50},
  xtick={0,1,2,3,4,5,6,7,8,9,10,11,12,13,14,15,16,17,18,19,20,21,22,23,24,25,26,27,28},
  xticklabels={%
    OpenMP, MPI, CUDA, OpenACC, HIP, MPI+OMP,
    OpenCL, Kokkos, OMP-Tgt, SYCL, RAJA,
    OpenSHMEM, Pthreads, TBB, Charm++, C++PSTL, UPC,
    CoarrayF, Chapel, Thrust, Serial,
    C++AMP, OCCA, GSParLib, STM/TM, HMPP, UPC++, Python, HLS},
  xticklabel style={font=\small, rotate=45, anchor=east},
  tick label style={font=\small},
  label style={font=\small},
  title style={font=\normalsize\bfseries},
  ymajorgrids=true,
  major grid style={dashed, draw=gray!40},
  title={API Coverage Across Surveyed Repositories},
  clip=false,
]
  \addplot[fill=pbTealDark, draw=pbTealDark!70!black, line width=0.5pt, bar shift=0pt]
    coordinates {
      (0,50)(1,45)(2,38)(3,17)(4,15)(5,15)
      (6,13)(7,12)(8,8)(9,7)(10,7)
      (11,5)(12,4)(13,4)(14,3)(15,3)(16,3)
      (17,3)(18,2)(19,2)(20,2)
      (21,1)(22,1)(23,1)(24,1)(25,1)(26,1)(27,1)(28,1)};
\end{axis}
\end{tikzpicture}
    \caption{API coverage across the benchmark survey. OpenMP (50 benchmarks), MPI (45), and CUDA (38) dominate, with OpenACC (17), HIP (15), and MPI+OpenMP (15) forming a second tier. OpenCL (13) and Kokkos (12) bridge the second and third tiers. OpenMP Target (8), SYCL (7), and RAJA (7) form a third tier.}
    \label{fig:api-coverage}
\end{figure}
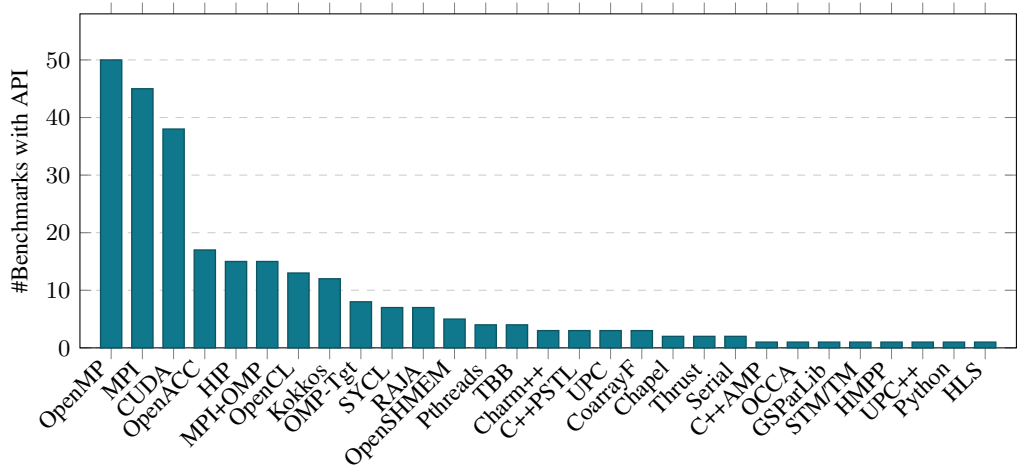

Figure~\ref{fig:api-coverage} presents the per-API coverage in a ranked bar chart, making the quantitative dominance of the top-3 APIs clear.

APIs below 5 benchmarks were excluded from evaluation consideration due to insufficient benchmark material for statistical evaluation.

\subsubsection*{Kernel-Level Co-Occurrence}

The repository-level analysis in Figures~\ref{fig:api-heatmap}--\ref{fig:api-coverage} counts how many \emph{repositories} provide implementations in a given API pair, but does not capture the depth of coverage within each repository. A suite like HeCBench provides 522 individual kernels, each potentially implemented in multiple APIs, while a single-kernel mini-application like miniBUDE provides only~1 kernel across 12+ APIs. For translation evaluation, the kernel count is the relevant unit: it determines how many independent evaluation instances are available per direction.

\begin{figure}[htbp]
    \centering
\begin{tikzpicture}
\begin{axis}[
  width=\columnwidth,
  height=\columnwidth,
  enlargelimits=false,
  axis on top,
  y dir=reverse,
  tick label style={font=\tiny},
  title style={font=\small\bfseries, align=center},
  colorbar,
  colorbar style={
    font=\tiny, width=0.3cm,
    ylabel={\# Kernels},
    ylabel style={font=\tiny},
  },
  xtick={0,1,2,3,4,5,6,7,8,9,10,11,12},
  xticklabels={CUDA,HIP,SYCL,OpenCL,OpenMP,{OpenMP Target},OpenACC,Kokkos,RAJA,TBB,Thrust,stdpar,Seq.},
  xticklabel style={font=\tiny, rotate=45, anchor=east},
  ytick={0,1,2,3,4,5,6,7,8,9,10,11,12},
  yticklabels={CUDA,HIP,SYCL,OpenCL,OpenMP,{OMP Target},OpenACC,Kokkos,RAJA,TBB,Thrust,stdpar,Seq.},
  yticklabel style={font=\tiny},
  xmin=-0.5, xmax=12.5,
  ymin=-0.5, ymax=12.5,
  colormap={ylOrRd}{
    rgb255(0)=(255,255,204)
    rgb255(100)=(254,217,118)
    rgb255(250)=(253,141,60)
    rgb255(420)=(227,26,28)
    rgb255(656)=(128,0,38)
  },
  point meta min=0, point meta max=656,
  title={Kernel-Level API Co-occurrence\\(Number of Kernels with Both APIs)},
]
  \addplot[matrix plot*, mesh/cols=13, point meta=explicit] coordinates {
    (0,0)[656]   (1,0)[nan]   (2,0)[nan]   (3,0)[nan]   (4,0)[nan]   (5,0)[nan]   (6,0)[nan]   (7,0)[nan]   (8,0)[nan]   (9,0)[nan]   (10,0)[nan]  (11,0)[nan]  (12,0)[nan]
    (0,1)[633]   (1,1)[633]   (2,1)[nan]   (3,1)[nan]   (4,1)[nan]   (5,1)[nan]   (6,1)[nan]   (7,1)[nan]   (8,1)[nan]   (9,1)[nan]   (10,1)[nan]  (11,1)[nan]  (12,1)[nan]
    (0,2)[616]   (1,2)[615]   (2,2)[616]   (3,2)[nan]   (4,2)[nan]   (5,2)[nan]   (6,2)[nan]   (7,2)[nan]   (8,2)[nan]   (9,2)[nan]   (10,2)[nan]  (11,2)[nan]  (12,2)[nan]
    (0,3)[27]    (1,3)[6]     (2,3)[6]     (3,3)[27]    (4,3)[nan]   (5,3)[nan]   (6,3)[nan]   (7,3)[nan]   (8,3)[nan]   (9,3)[nan]   (10,3)[nan]  (11,3)[nan]  (12,3)[nan]
    (0,4)[472]   (1,4)[453]   (2,4)[453]   (3,4)[24]    (4,4)[472]   (5,4)[nan]   (6,4)[nan]   (7,4)[nan]   (8,4)[nan]   (9,4)[nan]   (10,4)[nan]  (11,4)[nan]  (12,4)[nan]
    (0,5)[106]   (1,5)[106]   (2,5)[106]   (3,5)[0]     (4,5)[106]   (5,5)[106]   (6,5)[nan]   (7,5)[nan]   (8,5)[nan]   (9,5)[nan]   (10,5)[nan]  (11,5)[nan]  (12,5)[nan]
    (0,6)[22]    (1,6)[22]    (2,6)[22]    (3,6)[6]     (4,6)[22]    (5,6)[0]     (6,6)[22]    (7,6)[nan]   (8,6)[nan]   (9,6)[nan]   (10,6)[nan]  (11,6)[nan]  (12,6)[nan]
    (0,7)[22]    (1,7)[22]    (2,7)[22]    (3,7)[6]     (4,7)[22]    (5,7)[0]     (6,7)[22]    (7,7)[22]    (8,7)[nan]   (9,7)[nan]   (10,7)[nan]  (11,7)[nan]  (12,7)[nan]
    (0,8)[6]     (1,8)[6]     (2,8)[6]     (3,8)[6]     (4,8)[6]     (5,8)[0]     (6,8)[6]     (7,8)[6]     (8,8)[6]     (9,8)[nan]   (10,8)[nan]  (11,8)[nan]  (12,8)[nan]
    (0,9)[22]    (1,9)[22]    (2,9)[22]    (3,9)[6]     (4,9)[22]    (5,9)[0]     (6,9)[22]    (7,9)[22]    (8,9)[6]     (9,9)[22]    (10,9)[nan]  (11,9)[nan]  (12,9)[nan]
    (0,10)[6]    (1,10)[6]    (2,10)[6]    (3,10)[6]    (4,10)[6]    (5,10)[0]    (6,10)[6]    (7,10)[6]    (8,10)[6]    (9,10)[6]    (10,10)[6]   (11,10)[nan] (12,10)[nan]
    (0,11)[6]    (1,11)[6]    (2,11)[6]    (3,11)[6]    (4,11)[6]    (5,11)[0]    (6,11)[6]    (7,11)[6]    (8,11)[6]    (9,11)[6]    (10,11)[6]   (11,11)[6]   (12,11)[nan]
    (0,12)[106]  (1,12)[106]  (2,12)[106]  (3,12)[0]    (4,12)[106]  (5,12)[106]  (6,12)[0]    (7,12)[0]    (8,12)[0]    (9,12)[0]    (10,12)[0]   (11,12)[0]   (12,12)[106]
  };
  \addplot[
    only marks, mark=none, point meta=explicit,
    nodes near coords={\pgfmathfloattoint{\pgfplotspointmeta}\pgfmathresult},
    nodes near coords align={center},
    every node near coord/.style={font=\footnotesize\bfseries, text=white, inner sep=0pt}
  ] coordinates {
    (0,0)[656]  (0,1)[633]  (1,1)[633]  (0,2)[616]  (1,2)[615]  (2,2)[616]
    (0,4)[472]  (1,4)[453]  (2,4)[453]  (4,4)[472]
  };
  \addplot[
    only marks, mark=none, point meta=explicit,
    nodes near coords={\pgfmathfloattoint{\pgfplotspointmeta}\pgfmathresult},
    nodes near coords align={center},
    every node near coord/.style={font=\footnotesize\bfseries, text=black, inner sep=0pt}
  ] coordinates {
    (0,3)[27]    (1,3)[6]     (2,3)[6]     (3,3)[27]
    (3,4)[24]
    (0,5)[106]   (1,5)[106]   (2,5)[106]   (3,5)[0]     (4,5)[106]   (5,5)[106]
    (0,6)[22]    (1,6)[22]    (2,6)[22]    (3,6)[6]     (4,6)[22]    (5,6)[0]     (6,6)[22]
    (0,7)[22]    (1,7)[22]    (2,7)[22]    (3,7)[6]     (4,7)[22]    (5,7)[0]     (6,7)[22]    (7,7)[22]
    (0,8)[6]     (1,8)[6]     (2,8)[6]     (3,8)[6]     (4,8)[6]     (5,8)[0]     (6,8)[6]     (7,8)[6]     (8,8)[6]
    (0,9)[22]    (1,9)[22]    (2,9)[22]    (3,9)[6]     (4,9)[22]    (5,9)[0]     (6,9)[22]    (7,9)[22]    (8,9)[6]     (9,9)[22]
    (0,10)[6]    (1,10)[6]    (2,10)[6]    (3,10)[6]    (4,10)[6]    (5,10)[0]    (6,10)[6]    (7,10)[6]    (8,10)[6]    (9,10)[6]    (10,10)[6]
    (0,11)[6]    (1,11)[6]    (2,11)[6]    (3,11)[6]    (4,11)[6]    (5,11)[0]    (6,11)[6]    (7,11)[6]    (8,11)[6]    (9,11)[6]    (10,11)[6]   (11,11)[6]
    (0,12)[106]  (1,12)[106]  (2,12)[106]  (3,12)[0]    (4,12)[106]  (5,12)[106]  (6,12)[0]    (7,12)[0]    (8,12)[0]    (9,12)[0]    (10,12)[0]   (11,12)[0]   (12,12)[106]
  };
\end{axis}
\end{tikzpicture}
    \caption{Kernel-level API co-occurrence matrix. Each cell counts the number of individual kernels with implementations in both the row and column APIs. The CUDA--HIP pair leads with 633 shared kernels, followed by CUDA--SYCL (616), HIP--SYCL (615), and CUDA--OpenMP (472). OpenCL has dramatically lower kernel-level coverage: only 27 kernels share implementations with CUDA, and 24 with OpenMP. OpenMP Target provides 106 shared kernels with each of CUDA, HIP, SYCL, and OpenMP, concentrated entirely in RAJAPerf. Sequential reference implementations exist for 106 kernels.}
    \label{fig:kernel-cooccurrence}
\end{figure}
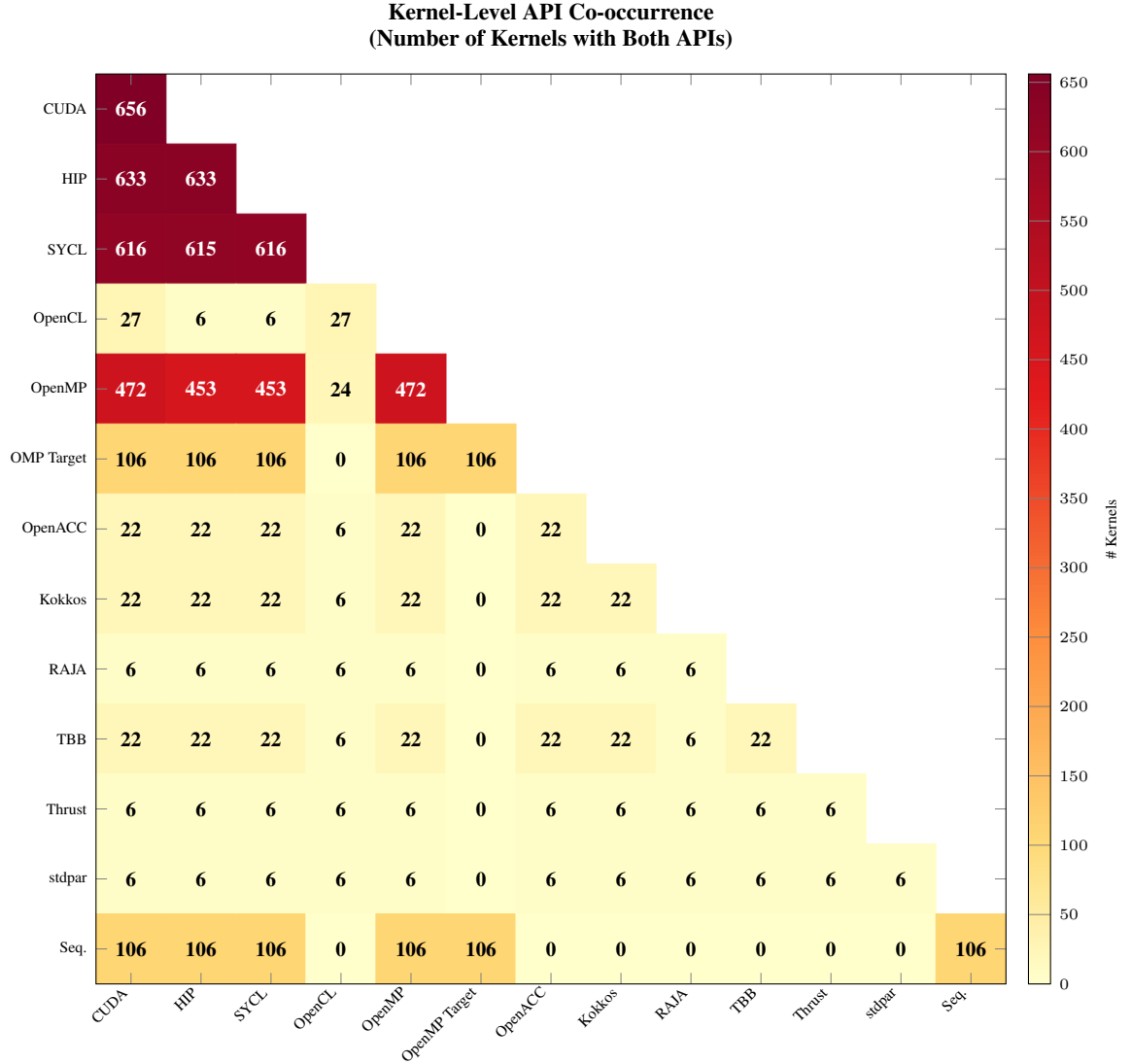

Figure~\ref{fig:kernel-cooccurrence} provides the strongest quantitative evidence for API selection. The kernel-level data reveals a clear hierarchy:

\begin{itemize}
    \item \textbf{Tier~1 ($>$600 kernel pairs):} CUDA--HIP (633), CUDA--SYCL (616), HIP--SYCL (615). These pairs are dominated by HeCBench's systematic multi-API coverage.
    \item \textbf{Tier~2 (${\sim}$450 kernel pairs):} CUDA--OpenMP (472), HIP--OpenMP (453), SYCL--OpenMP (453). OpenMP's CPU threading model provides the deepest pool for \emph{paradigm-crossing} translation evaluation.
    \item \textbf{Tier~3 (${\sim}$100 kernel pairs):} OpenMP Target and Sequential at 106 each, concentrated in RAJAPerf.
    \item \textbf{Tier~4 ($<$30 kernel pairs):} OpenCL at 27 (CUDA--OpenCL) and 24 (OpenMP--OpenCL), with an OpenCL diagonal of 27 total kernels.
\end{itemize}

\begin{figure}[htbp]
    \centering
    \includegraphics[width=\columnwidth]{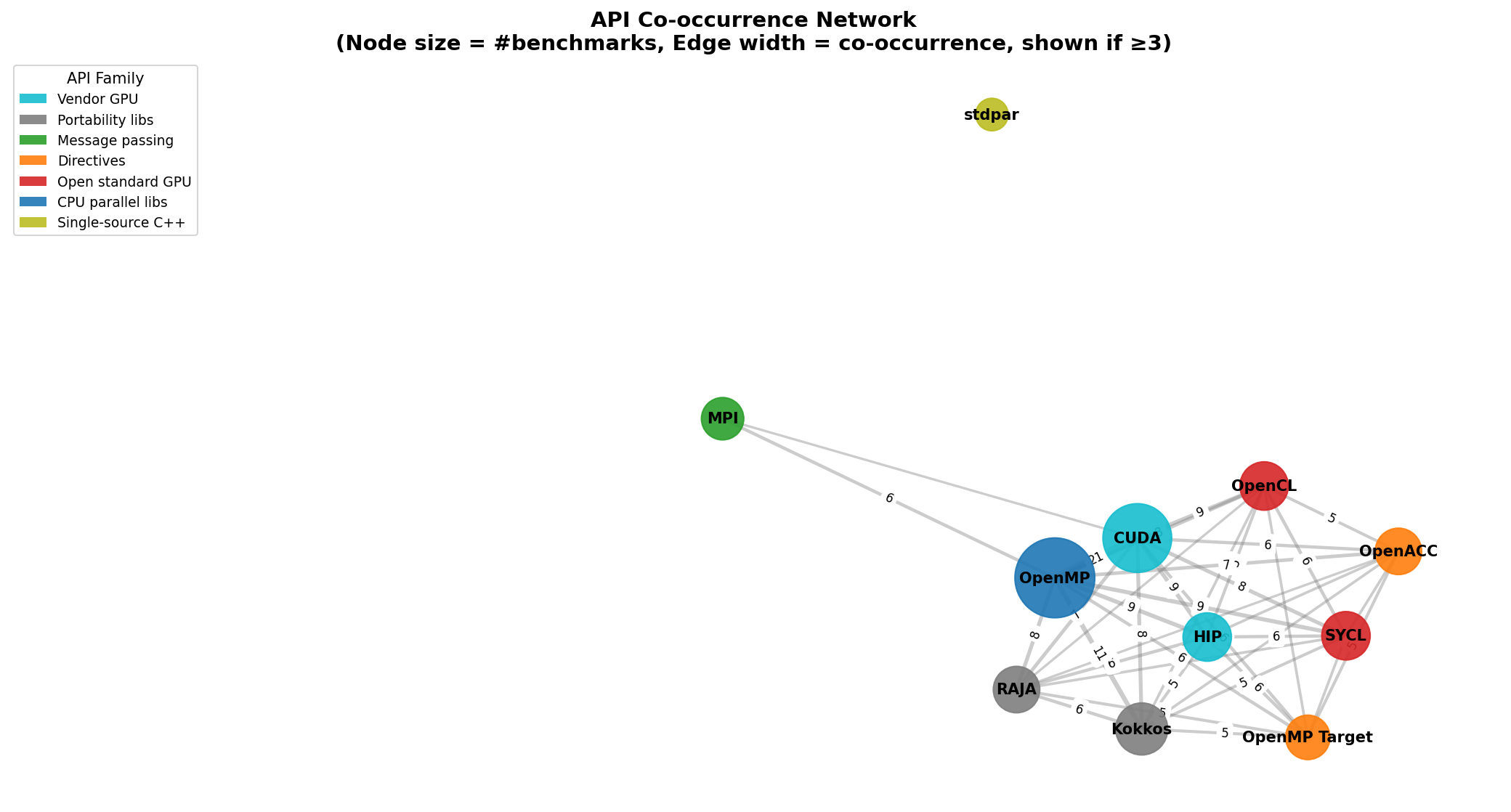}
    \caption{Kernel-level API co-occurrence network. Edges shown for $\geq 20$ kernel pairs; edge width proportional to count. The tight CUDA/HIP/SYCL/OpenMP cluster reflects HeCBench and RAJAPerf's uniform multi-API coverage. OpenCL appears as a peripheral node with weak connections (27 CUDA--OpenCL kernel pairs, just above the 20-kernel threshold). RAJA, stdpar, and Thrust appear as isolated nodes with fewer than 20 kernel-level pairs.}
    \label{fig:kernel-network}
\end{figure}

The network visualization in Figure~\ref{fig:kernel-network} makes this tiered
structure visually apparent.

\begin{figure}[htbp]
\centering
\begin{tikzpicture}
\begin{axis}[
  parbench compact,
  ybar,
  bar width=25pt,
  ymode=log,
  ymin=1, ymax=5000,
  ytick={1,10,100,1000},
  yticklabels={1,10,100,1{,}000},
  log ticks with fixed point,
  xtick={0,1,2},
  xticklabels={CUDA--OpenMP, CUDA--HIP, CUDA--SYCL},
  xticklabel style={font=\footnotesize},
  ylabel={Count (log scale)},
  xmin=-0.6, xmax=2.6,
  legend style={at={(0.5,-0.22)}, anchor=north, legend columns=2},
  ymajorgrids=true, xmajorgrids=false,
  grid style={dashed, draw=gray!40},
  title={Repository vs.\ Kernel Translation Pair Counts},
]
  \addplot[fill=pbOrange!70, draw=pbOrange!80!black, line width=0.8pt, nodes near coords, point meta=explicit symbolic, every node near coord/.style={
  font=\small\bfseries, anchor=south, xshift=-14pt, yshift=1pt, color=pbOrange!80!black}] coordinates {(0,6)[6] (1,3)[3] (2,2)[2]};
  \addlegendentry{Repository Count}
  
  \addplot[fill=pbRodinia, draw=pbRodinia!70!black, line width=0.8pt, nodes near coords, point meta=explicit symbolic, every node near coord/.style={font=\small\bfseries, anchor=south, xshift=12pt, yshift=1pt, color=black}]
  coordinates {(0,472)[472] (1,633)[633] (2,616)[616]};
  \addlegendentry{Kernel Count}

  \node[font=\small\bfseries, color=pbRose, anchor=south]
    at (axis cs:0,1500) {$79\times$};
  \node[font=\small\bfseries, color=pbRose, anchor=south]
    at (axis cs:1,2000) {$211\times$};
  \node[font=\small\bfseries, color=pbRose, anchor=south]
    at (axis cs:2,2000) {$308\times$};
\end{axis}
\end{tikzpicture}
\caption{Repository-level vs.\ kernel-level translation pair counts. Left bars
show the number of repositories containing both APIs; right bars show the number
of independent kernel-level translation pairs. The multipliers
(79$\times$--308$\times$) show why \parbench{} counts kernels rather than
repositories when estimating evaluation opportunity.}
\label{fig:repo-vs-kernel}
\end{figure}
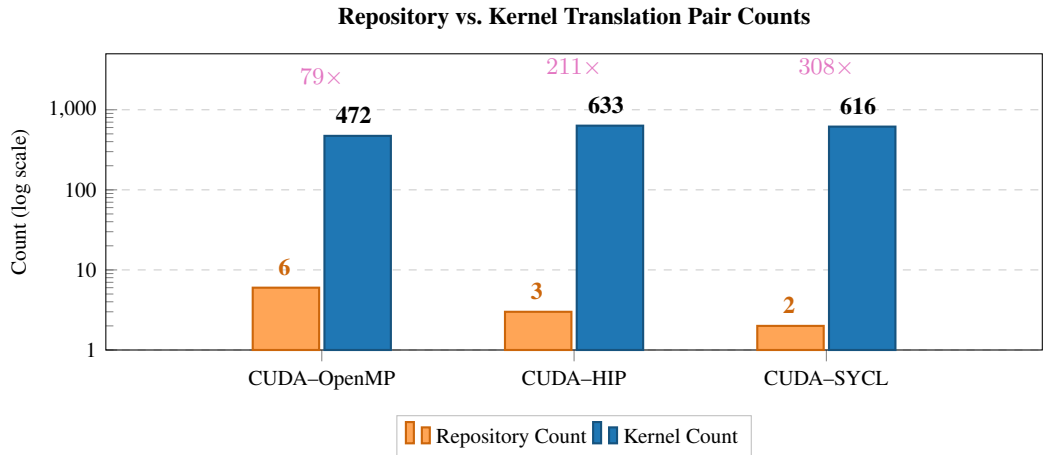

Figure~\ref{fig:repo-vs-kernel} complements the full survey views above by
showing why repository-level counts alone would understate the number of
independent translation opportunities available for evaluation.

\subsection{Translation Difficulty Taxonomy}
\label{sec:appendix-b2}

The preceding analysis establishes which translation directions are feasible at scale. The choice of API pairs for evaluation is motivated not only by data availability (Section~\ref{sec:appendix-b1}) but also by the \emph{structural transformation complexity} each direction demands. We classify translation directions into four categories:

\textbf{Syntactic renaming} (e.g., CUDA to HIP): Near-1:1 API call mapping. \texttt{cudaMalloc} becomes \texttt{hipMalloc}; kernel
launch syntax (\texttt{<<<grid, block>>>}) and thread-index arithmetic (\texttt{threadIdx.x}, \texttt{blockIdx.x}) are identical in
HIP. This category primarily tests API surface adaptation rather than parallel reasoning.

\begin{lstlisting}[language=C, caption={CUDA to HIP: syntactic renaming.}]
// CUDA                          // HIP (syntactic rename)
#include <cuda.h>                #include <hip/hip_runtime.h>
cudaMalloc(&d_a, size);          hipMalloc(&d_a, size);
kernel<<<grid, block>>>(d_a);    kernel<<<grid, block>>>(d_a);
cudaMemcpy(h_a, d_a, ...);       hipMemcpy(h_a, d_a, ...);
\end{lstlisting}

\textbf{Paradigm translation} (e.g., CUDA to OpenMP): SPMD to fork-join model transformation. Explicit GPU thread indexing must be converted to loop-based parallelism with OpenMP directives. Shared memory and synchronization primitives (\texttt{\_\_shared\_\_}, \texttt{\_\_syncthreads()}) must be replaced with stack-allocated arrays and implicit barrier semantics. Memory management (\texttt{cudaMalloc}/\texttt{cudaMemcpy}/\texttt{cudaFree}) is entirely eliminated.

\begin{lstlisting}[language=C, caption={CUDA to OpenMP: paradigm translation.}]
// CUDA                                  // OpenMP
__global__ void kern(float* a, int n) {  void kern(float* a, int n) {
  int i = blockIdx.x * blockDim.x         #pragma omp parallel for
           + threadIdx.x;                  for (int i = 0; i < n; i++) {
  if (i < n) a[i] = a[i] * 2.0f;            a[i] = a[i] * 2.0f;
}                                          }
                                         }
\end{lstlisting}

\textbf{Architecture split} (e.g., CUDA to OpenCL): CUDA's single-source compilation model (host and device code in one \texttt{.cu} file) must be split into separate host code (OpenCL runtime API calls for platform/device/context/queue management) and kernel source (\texttt{.cl} files with OpenCL~C syntax). The memory model changes from CUDA's device-pointer allocation (\texttt{cudaMalloc}/\texttt{cudaMemcpy}) to OpenCL's buffer objects (\texttt{clCreateBuffer}/\texttt{clEnqueueWriteBuffer}). Kernel launch syntax transforms from \texttt{<<<grid, block>>>} to \texttt{clEnqueueNDRangeKernel} with explicit work-group sizing.

\textbf{Directive insertion} (e.g., sequential~C to OpenMP): Adding parallelism to sequential code by identifying parallelizable loops and inserting appropriate \texttt{\#pragma omp} directives with correct data-sharing clauses (\texttt{shared}, \texttt{private}, \texttt{reduction}). This is the inverse of the paradigm translation direction and tests whether the LLM can \emph{discover} parallelism rather than \emph{translate} it.

\parbench{} targets the middle of this difficulty spectrum: paradigm translation and architecture split (CUDA/OpenMP/OpenCL cross-translations). Syntactic renaming (CUDA to HIP) is too easy to be informative, while directive insertion requires sequential baselines outside the current corpus. The six directions among CUDA, OpenMP, and OpenCL span paradigm translation (CUDA$\leftrightarrow$OpenMP) and architecture split (CUDA$\leftrightarrow$OpenCL, OpenMP$\leftrightarrow$OpenCL), maximizing the semantic challenge while remaining tractable for automated verification.

Table~\ref{tab:api-characteristics} summarizes the programming-model
differences that make these directions structurally different translation tasks.

\begin{table*}[htbp]
\centering
\caption{Programming model characteristics of the three primary translation
APIs. These differences define the structural transformation challenges that
LLM-based translation must address.}
\label{tab:api-characteristics}
\small
\resizebox{\textwidth}{!}{%
\begin{tabular}{@{}p{2.6cm}p{3.7cm}p{3.7cm}p{3.7cm}@{}}
\toprule
Dimension & CUDA & OpenMP & OpenCL \\
\midrule
Execution model & GPU SPMD kernels launched from host & CPU fork-join threads via pragmas & Split host/device model with runtime kernel compilation \\
Thread/work-item indexing & Explicit built-ins (\texttt{threadIdx}, \texttt{blockIdx}) & Loop indices and scheduling clauses & Explicit built-ins (\texttt{get\_global\_id}, \texttt{get\_local\_id}) \\
Memory management & Explicit device allocation/copy/free & Shared host memory by default & Buffer objects plus explicit enqueue/write/read \\
Compilation model & Single-source ahead-of-time GPU compilation & Host compiler with pragma lowering & Host code plus JIT-compiled kernel source \\
File structure tendency & Often single-source or host+kernel CUDA files & Usually host-language source only & Separate host source and \texttt{.cl} kernel file \\
Typical translation burden & Remove explicit GPU constructs or introduce them & Introduce or remove compiler directives and loop parallelism & Split or merge host/device logic; satisfy OpenCL runtime constraints \\
\bottomrule
\end{tabular}
}
\end{table*}

\subsection{HIP and SYCL Exclusion Rationale}
\label{sec:appendix-b3}

Despite the large kernel-level co-occurrence counts for CUDA--HIP (633) and CUDA--SYCL (616), these pairs were excluded from the primary evaluation for distinct reasons.

\textbf{HIP exclusion:} CUDA-to-HIP translation is predominantly a syntactic renaming task. The AMD HIP API was explicitly designed as a drop-in replacement for CUDA, preserving the SPMD programming model, kernel launch syntax, and memory management API. Automated tools such as \texttt{hipify-perl} and \texttt{hipify-clang} achieve near-perfect conversion rates. Evaluating LLMs on CUDA-to-HIP translation would measure API name lookup ability rather than parallel programming reasoning--the core capability \parbench{} is designed to
assess.

\textbf{SYCL exclusion:} While SYCL introduces a genuinely different programming model (single-source C++ with lambda-based kernel dispatch and buffer/accessor memory management), it preserves the GPU execution model. CUDA-to-SYCL translation is more complex than CUDA-to-HIP but remains within the same architectural paradigm (GPU offload). Furthermore, SYCL benchmark coverage is concentrated predominantly in HeCBench (approximately 80\% of kernel-level instances), limiting the diversity of evaluation instances. The CUDA-to-OpenMP direction provides a more fundamental paradigm crossing (GPU SPMD to CPU fork-join) with better benchmark diversity across Rodinia, HeCBench, XSBench, and RSBench.

\subsection{OpenACC Exclusion}
\label{sec:appendix-b4}

OpenACC was excluded despite appearing in 17 benchmarks in the repository-level survey (Figure~\ref{fig:api-coverage}), of which 12 also provide both CUDA and OpenMP implementations. 

Three factors motivated this decision:

\begin{enumerate}
    \item \textbf{Low kernel-level coverage}: At the kernel level, only 22 kernels provide OpenACC implementations across the surveyed repositories (Figure~\ref{fig:kernel-cooccurrence})--fewer than OpenCL (27) and substantially fewer than the primary APIs (CUDA: 656, OpenMP: 472). While repository-level co-occurrence with CUDA and OpenMP is high, the kernel-level material is insufficient for the multi-direction evaluation design that \parbench{} requires.
    
    \item \textbf{Paradigm overlap with OpenMP target}: OpenACC's directive-based model (\texttt{\#pragma acc parallel loop}) occupies a similar conceptual niche to OpenMP's target offload directives (\texttt{\#pragma omp target teams distribute}). Including both would provide diminishing returns in programming-model diversity. OpenCL, despite comparable kernel counts, is retained because it exercises a qualitatively distinct programming model (explicit host/kernel separation and JIT compilation) not covered by any other included API.
    
    \item \textbf{Compiler availability}: OpenACC compilation requires the NVIDIA HPC SDK (\texttt{nvc/nvc++}) or GCC with \texttt{-fopenacc}. This is a less universally available toolchain than CUDA (\texttt{nvcc}) or OpenMP (any modern C/C++ compiler), introducing a confounding variable between compiler availability and LLM translation capability.
\end{enumerate}

\subsection{OpenMP Target as Case Study}
\label{sec:appendix-b5}

OpenMP target offload (\texttt{\#pragma omp target}) occupies a middle ground between CPU OpenMP and GPU-native APIs. It uses the OpenMP directive model but targets GPU execution, requiring the programmer to specify data mapping (\texttt{\#pragma omp target data map(...)}) and work distribution (\texttt{\#pragma omp teams distribute parallel for}) explicitly.

\parbench{} evaluates OpenMP target as a case study rather than a primary direction
for three reasons:

\begin{enumerate}
    \item \textbf{Compiler requirement}: OpenMP target compilation for NVIDIA GPUs requires the NVIDIA HPC compiler (\texttt{nvc/nvc++}, part of the NVIDIA HPC SDK~24.3). Default GCC and Clang installations support only CPU-threaded OpenMP; GPU offloading requires custom builds with target-offload support enabled. This introduces a confounding variable: a failed build may indicate compiler unavailability rather than translation quality.
    
    \item \textbf{Limited benchmark coverage}: The kernel-level survey identifies 106 kernels with both OpenMP target and CUDA implementations (Figure~\ref{fig:kernel-cooccurrence}), concentrated entirely in RAJAPerf. Within \parbench{}'s curated corpus, OpenMP target implementations are available for 12~kernels: 10 from HeCBench, 1 from XSBench, and 1 from RSBench. Rodinia does not provide OpenMP target variants.
    
    \item \textbf{Compilation model difference}: OpenMP target generates GPU offload code through the compiler, while CPU OpenMP generates threaded CPU code. A translation from CUDA to OpenMP target preserves GPU execution semantics but changes the syntax entirely; a translation from CUDA to CPU OpenMP changes both the execution model and the syntax. These are qualitatively different evaluation targets that should not be conflated in
    aggregate statistics.
\end{enumerate}

The case study results for OpenMP target are reported separately in the direction analysis (Appendix Table~\ref{tab:direction-rates}) to enable direct comparison with the primary CUDA/OpenMP/OpenCL directions without confounding the aggregate pass rates.
  

\section{Benchmark Kernel Survey}
\label{sec:appendix-c}

This appendix documents the systematic survey and multi-stage selection process that produced \parbench{}'s evaluation corpus. The main paper (Section~\ref{sec:benchmark-curation}) summarizes this process; here we provide the complete data, selection criteria,
and exclusion rationale.

\subsection{Full Repository Inventory}
\label{sec:appendix-c1}

\begin{figure}[htbp]
    \centering
\begin{tikzpicture}
\begin{axis}[
  ybar,
  bar width=18pt,
  height=5.5cm,
  width=\columnwidth,
  xmin=-0.6, xmax=4.6,
  ylabel={\#Repositories},
  ymin=0, ymax=15,
  ytick={0,5,10,15},
  xtick={0,1,2,3,4},
  xticklabels={Suite, Miniapp, Library, {Proxy App}, Application},
  xticklabel style={font=\small},
  tick label style={font=\small},
  label style={font=\small},
  title style={font=\normalsize\bfseries},
  ymajorgrids=true,
  major grid style={dashed, draw=gray!40},
  title={Distribution of Benchmark Types (35 Repositories)},
  nodes near coords,
  nodes near coords style={font=\small\bfseries, anchor=south},
  clip=false,
]
  \addplot[fill=pbTealDark, draw=pbTealDark!70!black, line width=0.5pt]
    coordinates {(0,13) (1,12) (2,4) (3,4) (4,2)};
\end{axis}
\end{tikzpicture}
    \caption{Distribution of benchmark types across the 35 repositories that met inclusion criteria. Suites (13, 37\%) and mini-applications (12, 34\%) dominate, accounting for 71\% of included repositories. Proxy applications (4, 11\%), libraries (4, 11\%), and full applications (2, 6\%) complete the distribution. Suites and mini-applications yield the richest kernel-level material for translation evaluation: they contain multiple independent computational kernels with self-checking verification.}
    \label{fig:benchmark-types}
\end{figure}
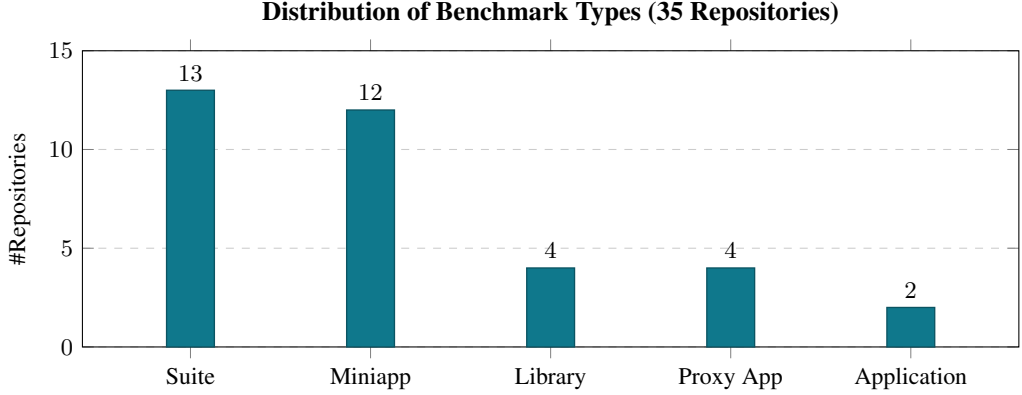

The benchmark landscape survey covered 71 open-source HPC repositories (Appendix~\ref{sec:appendix-b1}). After applying the inclusion criteria described in Section~\ref{sec:appendix-c2}, 35 repositories remained as candidates for kernel extraction, spanning five categories of HPC benchmark software (Figure~\ref{fig:benchmark-types}).

API coverage varies widely across repositories: some implement a single API (e.g., SPEC OMP, LonestarGPU, STREAM), while others provide implementations across 10+ APIs (BabelStream and miniBUDE each with~12).

\begin{table*}[!htbp]
\centering
\caption{Quantitative characterization of the evaluation corpus (35 kernels, 96 specs across 5 suites). SLoC counts physical source lines (non-blank, non-comment) in each kernel's CUDA prompt payload, the code provided to the LLM as translation input. Multi-File gives specs requiring translation of $>$1 source file. Std.\ shows the dominant language standard per suite.}
\label{tab:benchmark-characterization}
\small
\begin{tabular}{@{}lcrcl@{}}
\toprule
Suite & SLoC Range & Med. & Multi-File & Std. \\
\midrule
Rodinia            & 195--3{,}304       & 334 & 21/60 (35\%) & C++14 \\
HeCBench (curated) &    80--235         & 180 &  0/25 ~(0\%) & C++17 \\
XSBench            &    1{,}390         &  -- &  2/4~~(50\%) & C11   \\
RSBench            &    1{,}016         &  -- &  1/4~~(25\%) & C11   \\
mixbench           &       312          &  -- &  0/3~~~(0\%) & C++11 \\
\midrule
\textbf{Total} & \textbf{80--3{,}304} & \textbf{271} & \textbf{24/96 (25\%)} & \\
\bottomrule
\end{tabular}
\end{table*}

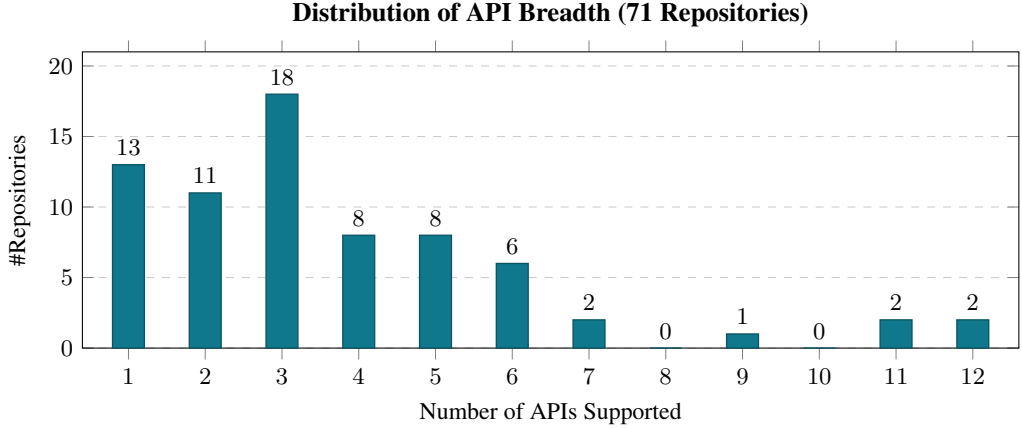
\begin{figure}[htbp]
    \centering
\begin{tikzpicture}
\begin{axis}[
  ybar,
  bar width=12pt,
  height=5.5cm,
  width=\columnwidth,
  xmin=0.4, xmax=12.6,
  ylabel={\#Repositories},                                                                                                                                                                                                                                                                                                                                                                                                                                                                                                                                                                                                                                                                                                                                                                                                                                                                      
  xlabel={Number of APIs Supported},
  ymin=0, ymax=21,
  ytick={0,5,10,15,20},
  xtick={1,2,3,4,5,6,7,8,9,10,11,12},
  tick label style={font=\small},
  label style={font=\small},
  title style={font=\normalsize\bfseries},
  ymajorgrids=true,
  major grid style={dashed, draw=gray!40},
  title={Distribution of API Breadth (71 Repositories)},
  nodes near coords,
  nodes near coords style={font=\small\bfseries, anchor=south},
  clip=false,
]
  \addplot[fill=pbTealDark, draw=pbTealDark!70!black, line width=0.5pt]
    coordinates {(1,13) (2,11) (3,18) (4,8) (5,8) (6,6) (7,2) (8,0) (9,1) (10,0) (11,2) (12,2)};
\end{axis}
\end{tikzpicture}
    \caption{Distribution of API breadth across all 71 surveyed repositories. The mode is 3~APIs (18~repositories); a long tail extends to 12~APIs (BabelStream and miniBUDE). 24~repositories support only 1--2~APIs and are unsuitable for cross-API translation evaluation. Repositories supporting $\geq$3~APIs constitute the candidate pool for multi-direction evaluation.}
    \label{fig:api-count-hist}
\end{figure}

The API breadth distribution across the full 71-repository survey reveals the selection opportunity (Figure~\ref{fig:api-count-hist}).

Taken together, these distributions show that kernel-level translation material is concentrated in a small number of large suites with broad API coverage--a Pareto distribution that guided suite selection. HeCBench and Rodinia were selected as the primary contributors due to their kernel richness and three-API coverage; XSBench, RSBench, and mixbench were added to extend domain coverage to nuclear physics simulation and GPU roofline characterization.

\subsection{Quality Tier Classification}
\label{sec:appendix-c2}

Of the 71 surveyed repositories, 36 were excluded during initial screening: 31 lacked multi-API HPC kernel content or fell outside the CUDA/OpenMP/OpenCL scope, and 5 had inaccessible downloads, missing documentation, or duplicate content. The remaining 35 repositories were classified into two quality tiers based on three criteria: build documentation, verification capability, and maintenance status.

\textbf{Tier~A criteria} (all three required):
\begin{itemize}
    \item Documented build process (Makefile, CMake, or equivalent with clear instructions)
    \item Automated verification: self-checking output (print PASS/FAIL, compute checksum, compare against reference) or reference output file comparison
    \item Active maintenance: commits within 2~years of the survey date, or stable release with no known build failures on modern toolchains
\end{itemize}

\textbf{Tier~B criteria} (any one triggers downgrade):
\begin{itemize}
    \item Partial verification only (e.g., visual output inspection, manual comparison)
    \item Limited API coverage (fewer than three of the target APIs, reducing cross-translation utility)
    \item Unmaintained but still buildable (no commits in 2+ years, but Makefiles still produce working binaries)
\end{itemize}

Tier~B repositories were not excluded from the survey but were deprioritized in kernel selection. Their kernels were considered only when no Tier~A alternative existed for a given computational domain.

\subsection{Candidate Ranking and HeCBench Selection Funnel}
\label{sec:appendix-c3}

The selection of individual kernels for \parbench{}'s evaluation corpus required balancing two competing objectives: \emph{kernel richness} (maximizing the number of independent evaluation instances) and \emph{API breadth} (maximizing the number of translation directions per kernel). Figure~\ref{fig:kernel-api-bubble} plots these two dimensions for all surveyed repositories.

\begin{figure}[htbp]
    \centering
\begin{tikzpicture}
\begin{axis}[
  parbench compact,
  width=\columnwidth,
  height=0.7\columnwidth,
  xlabel={Number of APIs Supported},
  ylabel={Number of Kernels (capped at 100)},
  xmin=0, xmax=13,
  ymin=-5, ymax=110,
  xtick={1,2,3,4,5,6,7,8,9,10,11,12},
  grid=major,
  grid style={dashed, draw=gray!30},
  title={Benchmark Kernel Richness vs API Breadth\\(Bubble size $=$ kernel count)},
  title style={font=\small\bfseries, align=center},
  legend style={at={(1.0,1.0)}, anchor=north east, font=\tiny, draw=gray!50},
  clip=false,
]
\addplot[only marks, mark=*,
  fill=pbOrange!80, draw=pbOrange!80!black, fill opacity=0.7,
  visualization depends on={\thisrow{size} \as \perpointmarksize},
  scatter/@pre marker code/.append style={/tikz/mark size=\perpointmarksize},
] table[x=x, y=y, meta=size] {figures/figures_tek_version/kernel_api_bubble_microbenchmarks.dat};
\addlegendentry{Microbenchmarks}
\addplot[only marks, mark=*,
  fill=pbPass!80, draw=pbPass!80!black, fill opacity=0.7,
  visualization depends on={\thisrow{size} \as \perpointmarksize},
  scatter/@pre marker code/.append style={/tikz/mark size=\perpointmarksize},
] table[x=x, y=y, meta=size] {figures/figures_tek_version/kernel_api_bubble_suites.dat};
\addlegendentry{Suites}
\addplot[only marks, mark=*,
  fill=cyan!60, draw=cyan!60!black, fill opacity=0.7,
  visualization depends on={\thisrow{size} \as \perpointmarksize},
  scatter/@pre marker code/.append style={/tikz/mark size=\perpointmarksize},
] table[x=x, y=y, meta=size] {figures/figures_tek_version/kernel_api_bubble_miniapps.dat};
\addlegendentry{Miniapps}
\addplot[only marks, mark=*,
  fill=blue!20, draw=blue!40, fill opacity=0.7,
  visualization depends on={\thisrow{size} \as \perpointmarksize},
  scatter/@pre marker code/.append style={/tikz/mark size=\perpointmarksize},
] table[x=x, y=y, meta=size] {figures/figures_tek_version/kernel_api_bubble_proxy_apps.dat};
\addlegendentry{Proxy apps}
\addplot[only marks, mark=*,
  fill=orange!50, draw=orange!70!black, fill opacity=0.7,
  visualization depends on={\thisrow{size} \as \perpointmarksize},
  scatter/@pre marker code/.append style={/tikz/mark size=\perpointmarksize},
] table[x=x, y=y, meta=size] {figures/figures_tek_version/kernel_api_bubble_libraries.dat};
\addlegendentry{Libraries}
\addplot[only marks, mark=*,
  fill=blue!60, draw=blue!80!black, fill opacity=0.7,
  visualization depends on={\thisrow{size} \as \perpointmarksize},
  scatter/@pre marker code/.append style={/tikz/mark size=\perpointmarksize},
] table[x=x, y=y, meta=size] {figures/figures_tek_version/kernel_api_bubble_applications.dat};
\addlegendentry{Applications}
\node[font=\tiny, anchor=south east] at (axis cs:6,100) {HeCBench};
\node[font=\tiny, anchor=south east] at (axis cs:5,80) {RAJAPerf};
\node[font=\tiny, anchor=south east] at (axis cs:11,23) {CloverLeaf};
\node[font=\tiny, anchor=south west] at (axis cs:12,5) {BabelStream};
\node[font=\tiny, anchor=north west] at (axis cs:12,1) {miniBUDE};
\end{axis}
\end{tikzpicture}
    \caption{Benchmark kernel richness (y-axis, capped at 100 for readability) vs.\
    API breadth (x-axis, number of APIs supported). Bubble size proportional to total kernel count; color indicates benchmark type. HeCBench (522 kernels, 6~APIs\protect\footnote{HeCBench's 6-API count treats OpenMP threading and OpenMP-target offloading as distinct APIs and includes limited MPI support; the four primary implementation directories are CUDA, HIP, SYCL, and OpenMP.}) dominates the upper region. RAJAPerf (80 kernels, 5~APIs) is the second-largest. CloverLeaf (23 kernels, 11~APIs) and BabelStream (5 kernels, 12~APIs) occupy the high-API-breadth region but with far fewer kernels. The optimal candidates for corpus construction lie in the high-kernel, moderate-API quadrant (upper-center).}
  \label{fig:kernel-api-bubble}
\end{figure}

\begin{figure}[htbp]
    \centering
\begin{tikzpicture}
\begin{axis}[
  parbench compact,
  ybar,
  bar width=8pt,
  height=6cm,
  ylabel={Count},
  xtick={0,1,2,3,4,5,6,7},
  xticklabels={HeCBench, BabelStream, RAJAPerf, miniBUDE, CloverLeaf, Kokkos, PRK, Rodinia},
  xticklabel style={font=\tiny, rotate=45, anchor=east},
  xmin=-0.6, xmax=7.6,
  ymin=0, ymax=18,
  ytick={0,3,6,9,12,15},
  legend style={at={(0.5,0.98)}, anchor=north, font=\footnotesize}, legend columns=2,
  ymajorgrids=true,
  grid style={dashed, draw=gray!20},
  title={Top Multi-API Repository Candidates},
  clip=false,
]
  \addplot[fill=pbRodinia, draw=pbRodinia!70!black, line width=0.6pt]
    coordinates {(0,14.9) (1,0.14) (2,2.3) (3,0.03) (4,0.66) (5,1.4) (6,0.34) (7,0.63)};
  \addlegendentry{Kernels (scaled)}

  \addplot[fill=pbOrange, draw=pbOrange!70!black, line width=0.6pt]
    coordinates {(0,6) (1,12) (2,5) (3,12) (4,11) (5,4) (6,9) (7,4)};
  \addlegendentry{APIs}

  \node[above, font=\tiny\bfseries, color=pbRodinia!80!black] at (axis cs:-0.13,14.9) {522};
  \node[above, font=\tiny\bfseries, color=pbRodinia!80!black] at (axis cs:0.87,0.14) {5};
  \node[above, font=\tiny\bfseries, color=pbRodinia!80!black] at (axis cs:1.87,2.3) {80};
  \node[above, font=\tiny\bfseries, color=pbRodinia!80!black] at (axis cs:2.87,0.03) {1};
  \node[above, font=\tiny\bfseries, color=pbRodinia!80!black] at (axis cs:3.87,0.66) {23};
  \node[above, font=\tiny\bfseries, color=pbRodinia!80!black] at (axis cs:4.87,1.4) {50};
  \node[above, font=\tiny\bfseries, color=pbRodinia!80!black] at (axis cs:5.87,0.34) {12};
  \node[above, font=\tiny\bfseries, color=pbRodinia!80!black] at (axis cs:6.87,0.63) {22};

  \node[above, font=\tiny\bfseries, color=pbOrange!80!black] at (axis cs:0.13,6) {6};
  \node[above, font=\tiny\bfseries, color=pbOrange!80!black] at (axis cs:1.13,12) {12};
  \node[above, font=\tiny\bfseries, color=pbOrange!80!black] at (axis cs:2.13,5) {5};
  \node[above, font=\tiny\bfseries, color=pbOrange!80!black] at (axis cs:3.13,12) {12};
  \node[above, font=\tiny\bfseries, color=pbOrange!80!black] at (axis cs:4.13,11) {11};
  \node[above, font=\tiny\bfseries, color=pbOrange!80!black] at (axis cs:5.13,4) {4};
  \node[above, font=\tiny\bfseries, color=pbOrange!80!black] at (axis cs:6.13,9) {9};
  \node[above, font=\tiny\bfseries, color=pbOrange!80!black] at (axis cs:7.13,4) {4};
\end{axis}
\end{tikzpicture}
    \caption{Top multi-API repository candidates ranked by kernel count (blue bars, scaled for readability; true counts annotated above) and API breadth (orange bars). HeCBench provides the largest kernel pool (522 kernels, 6~APIs spanning CUDA, HIP, SYCL, OpenMP, OpenMP-target offloading, and MPI). BabelStream and miniBUDE provide the broadest API coverage (12~APIs each) but with few kernels. RAJAPerf (80 kernels, 5~APIs) offers a balanced middle ground. Rodinia (22 kernels, 4~APIs including a community SYCL port) provides established, well-understood benchmarks with strong community recognition.}
    \label{fig:top-candidates}
\end{figure}
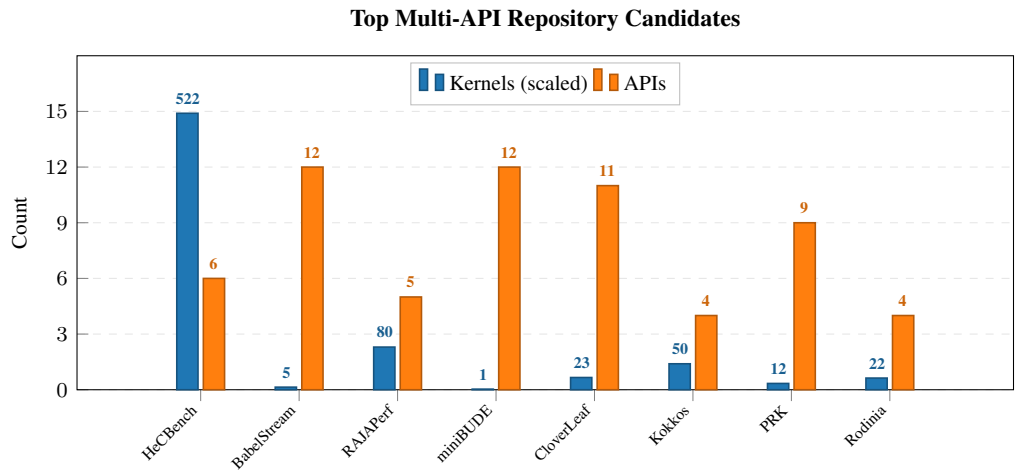

Figure~\ref{fig:top-candidates} presents the top candidates ranked by their combined kernel-count and API-breadth scores, revealing the fundamental tradeoff in corpus construction.

\subsection{HeCBench Selection Funnel}

HeCBench's 522 kernels were filtered through a five-stage funnel to identify the curated evaluation set:

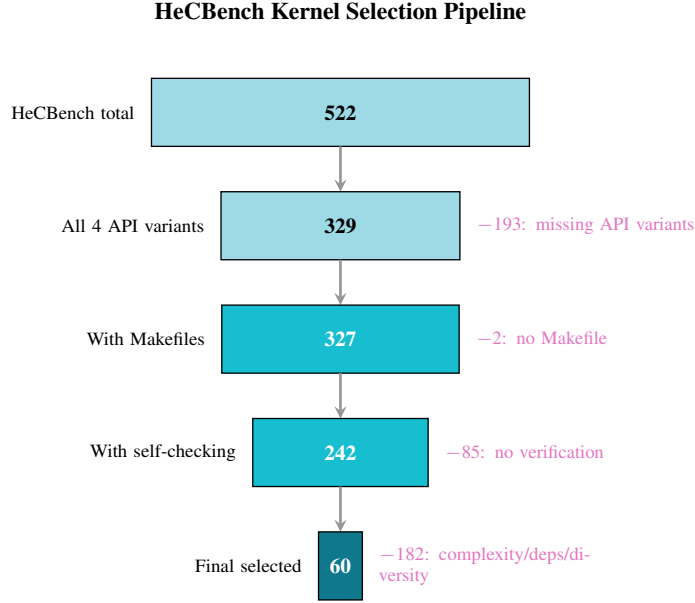
\begin{figure}[htbp]
  \centering
\begin{tikzpicture}[x=1cm, y=1cm]
\node[font=\small\bfseries, anchor=south] at (2.5,7.50) {HeCBench Kernel Selection Pipeline};
  \fill[pbTealLight, draw=black, line width=0.5pt] (0.00,6.00) rectangle (5.00,6.90);
  \node[font=\footnotesize\bfseries, text=black] at (2.50,6.45) {522};
  \node[font=\scriptsize, anchor=east, align=right, text width=3.0cm] at (-0.10,6.45) {HeCBench total};
  \draw[->, >=stealth, color=pbGray, line width=1pt] (2.50,6.00) -- (2.50,5.40);
  \fill[pbTealLight, draw=black, line width=0.5pt] (0.92,4.50) rectangle (4.08,5.40);
  \node[font=\footnotesize\bfseries, text=black] at (2.50,4.95) {329};
  \node[font=\scriptsize, anchor=east, align=right, text width=3.0cm] at (0.82,4.95) {All 4 API variants};
  \node[font=\scriptsize, anchor=west, color=pbRose, align=left, text width=3.0cm] at (4.18,4.95) {$-$193: missing API variants};
  \draw[->, >=stealth, color=pbGray, line width=1pt] (2.50,4.50) -- (2.50,3.90);
  \fill[pbTeal, draw=black, line width=0.5pt] (0.93,3.00) rectangle (4.07,3.90);
  \node[font=\footnotesize\bfseries, text=white] at (2.50,3.45) {327};
  \node[font=\scriptsize, anchor=east, align=right, text width=3.0cm] at (0.83,3.45) {With Makefiles};
  \node[font=\scriptsize, anchor=west, color=pbRose, align=left, text width=3.0cm] at (4.17,3.45) {$-$2: no Makefile};
  \draw[->, >=stealth, color=pbGray, line width=1pt] (2.50,3.00) -- (2.50,2.40);
  \fill[pbTeal, draw=black, line width=0.5pt] (1.34,1.50) rectangle (3.66,2.40);
  \node[font=\footnotesize\bfseries, text=white] at (2.50,1.95) {242};
  \node[font=\scriptsize, anchor=east, align=right, text width=3.0cm] at (1.24,1.95) {With self-checking};
  \node[font=\scriptsize, anchor=west, color=pbRose, align=left, text width=3.0cm] at (3.76,1.95) {$-$85: no verification};
  \draw[->, >=stealth, color=pbGray, line width=1pt] (2.50,1.50) -- (2.50,0.90);
  \fill[pbTealDark, draw=black, line width=0.5pt] (2.21,0.00) rectangle (2.79,0.90);
  \node[font=\footnotesize\bfseries, text=white] at (2.50,0.45) {60};
  \node[font=\scriptsize, anchor=east, align=right, text width=3.0cm] at (2.11,0.45) {Final selected};
  \node[font=\scriptsize, anchor=west, color=pbRose, align=left, text width=3.0cm] at (2.89,0.45) {$-$182: complexity/deps/diversity};
\end{tikzpicture}
  \caption{HeCBench kernel selection funnel. Starting from 522 kernels spanning
    CUDA, HIP, SYCL, and OpenMP at commit \texttt{22785cdd}, successive filters
    for API completeness (329 with all 4~variants), build infrastructure (327),
    self-checking output patterns (242), and complexity bounds yield a 60-kernel
    working set. Manual curation further narrows this to 10~evaluation kernels
    (25~specs; see Section~\ref{sec:benchmark-curation}).}
  \label{fig:selection-funnel}
\end{figure}

\begin{itemize}
    \item \textbf{Stage~1 (API completeness):} Of 522 total kernels in the HeCBench repository, 329 provide implementations in all 4~primary APIs (CUDA, HIP, SYCL, OpenMP).
    \item \textbf{Stage~2 (Build infrastructure and verification):} Of these 329 kernels, 2~lack Makefiles and were excluded. Of the remaining 327, only 242 produce deterministic, machine-parseable self-checking output--printing PASS/FAIL verdicts, computing checksums, or comparing numerical results against reference values--as determined by manual inspection of the CUDA source at commit \texttt{22785cdd}. These 242 candidates proceed to complexity filtering.
    \item \textbf{Stage~3 (Complexity filter):} The 242 candidates were filtered for evaluation tractability. Kernels with more than 15 source files or fewer than 2 were excluded (the former are too complex for single-prompt translation; the latter are trivial boilerplate). Kernels requiring external input data files were also excluded to ensure self-contained evaluation.
    \item \textbf{Stage~4 (Domain diversity):} From the complexity-filtered set, 60 kernels were manually selected to ensure coverage across 10 computational categories: linear algebra, stencil computation, graph algorithms, machine learning, physics simulation, sorting, reduction, scan, image processing, and cryptography.
    \item \textbf{Stage~5 (Curation):} The 60-kernel working set was narrowed to a final curated set of 20 kernels based on manual inspection of build reliability, output determinism, and domain representativeness. Of these, 10 kernels were included in the current \parbench{} evaluation corpus, with the remaining~10 reserved for future expansion.
\end{itemize}

The 10 curated kernels included in the current corpus are: \texttt{convolution1d}, \texttt{floydwarshall}, \texttt{heat2d}, \texttt{iso2dfd}, \texttt{jacobi}, \texttt{md}, \texttt{nqueen}, \texttt{page-rank}, \texttt{scan}, and \texttt{stencil1d}. Each provides a CUDA implementation paired with one or more OpenMP/OpenMP-target variants, totaling 25~specs. Two of these are \knownfail{}: \texttt{stencil1d-omp\_target} (build failure due to target-offloading compilation) and \texttt{scan-omp\_target} (output mismatch on the CPU offload target), leaving 23 in the active evaluation corpus.

This funnel illustrates a general challenge in constructing LLM code translation
benchmarks: the vast majority of available HPC kernels are unsuitable for
automated evaluation due to missing API variants, non-deterministic output, or
excessive complexity. Only 3.8\% of HeCBench's 522 kernels (20/522) survive the
full selection pipeline, highlighting the gap between ``available parallel code''
and ``evaluable parallel code.''

\subsection{Rodinia Kernel Inventory}
\label{sec:appendix-c4}

Rodinia~\cite{Rodinia2009} provides the foundation of \parbench{}'s evaluation corpus. As one of the most widely cited heterogeneous computing benchmark suites, it offers well-understood computational characteristics across CUDA, OpenMP, and OpenCL. \parbench{} includes all 22 Rodinia kernels, yielding 60~total specs; five kernels lack one or more API variants because the upstream repository does not provide implementations for all three APIs (six API variants absent in total, as one kernel--\texttt{huffman}--lacks both OpenMP and OpenCL). Of these 60 specs:

\begin{itemize}
    \item \textbf{53 \pass{}}: Each verified via conjunctive application of all its oracle strategies--every strategy must pass for the spec to pass. 48~of the 53 use exit-code-zero \emph{and} stdout pattern matching. The remaining five employ stronger oracle methods: three (\texttt{hotspot3d}) augment the conjunctive check with numeric comparison against baseline output within a fixed tolerance,
    \item \textbf{7 \knownfail{}}: Excluded from evaluation due to pre-existing platform failures unrelated to LLM translation quality:
    2~specs use deprecated CUDA \texttt{texture<>} references removed in CUDA~12 (\texttt{kmeans-cuda}, \texttt{mummergpu-cuda});
    1~requires OpenGL (\texttt{hybridsort-cuda});
    1~has CUDA API dependencies in its OpenMP variant (\texttt{mummergpu-omp});
    2~exhibit pre-existing OpenCL segmentation faults (\texttt{nn-opencl}, \texttt{kmeans-opencl}); and
    1~has a silent \texttt{clEnqueueReadBuffer} error masked by exit code~0 (\texttt{backprop-opencl}).
\end{itemize}

\subsection{XSBench, RSBench, and mixbench Profiles}
\label{sec:appendix-c5}

Three additional benchmark suites complement Rodinia and HeCBench in the evaluation corpus:

\textbf{XSBench}~\cite{XSBench2014} (4~specs: CUDA, OpenMP, OpenCL, OpenMP Target): A Monte Carlo neutron cross-section lookup proxy application representative of nuclear reactor simulation workloads. XSBench implements a unionized energy grid algorithm; the OpenMP variant runs in history-based mode (checksum 941{,}535) while the CUDA, OpenCL, and OpenMP Target variants run in event-based mode (checksum 945{,}990). Both checksums are self-verified as valid by XSBench's built-in verification. This mode asymmetry reflects algorithmic differences between CPU and GPU scheduling strategies rather than correctness
issues. All 4~API variants pass baseline verification.

\textbf{RSBench}~\cite{RSBench2015} (4~specs: CUDA, OpenMP, OpenCL, OpenMP Target): A proxy application for the multipole method of computing neutron cross sections on-the-fly, representing a complementary algorithm to XSBench's unionized energy grid approach. Like XSBench, the OpenMP variant defaults to history-based mode (checksum 879{,}693) while the CUDA, OpenCL, and OpenMP Target variants run in event-based mode (checksum 880{,}018), with both checksums self-verified as valid. Including both suites enables controlled comparison within the same computational domain (Monte Carlo neutron transport) at different algorithmic complexity levels. All 4~API variants pass baseline verification.

\textbf{mixbench}~\cite{mixbench2017} (3~specs: CUDA, OpenMP, OpenCL): A micro-benchmark that sweeps a range of compute-to-memory operation ratios, mapping the practical roofline boundary of a compute device. Its parameterized kernel structure tests whether LLM translations preserve arithmetic intensity across API boundaries. All 3~API variants pass baseline verification.

\subsection{Exclusion Log and Verification Methods}
  \label{sec:appendix-c6}

\begin{figure}[htbp]
    \centering
    \includegraphics[width=\columnwidth]{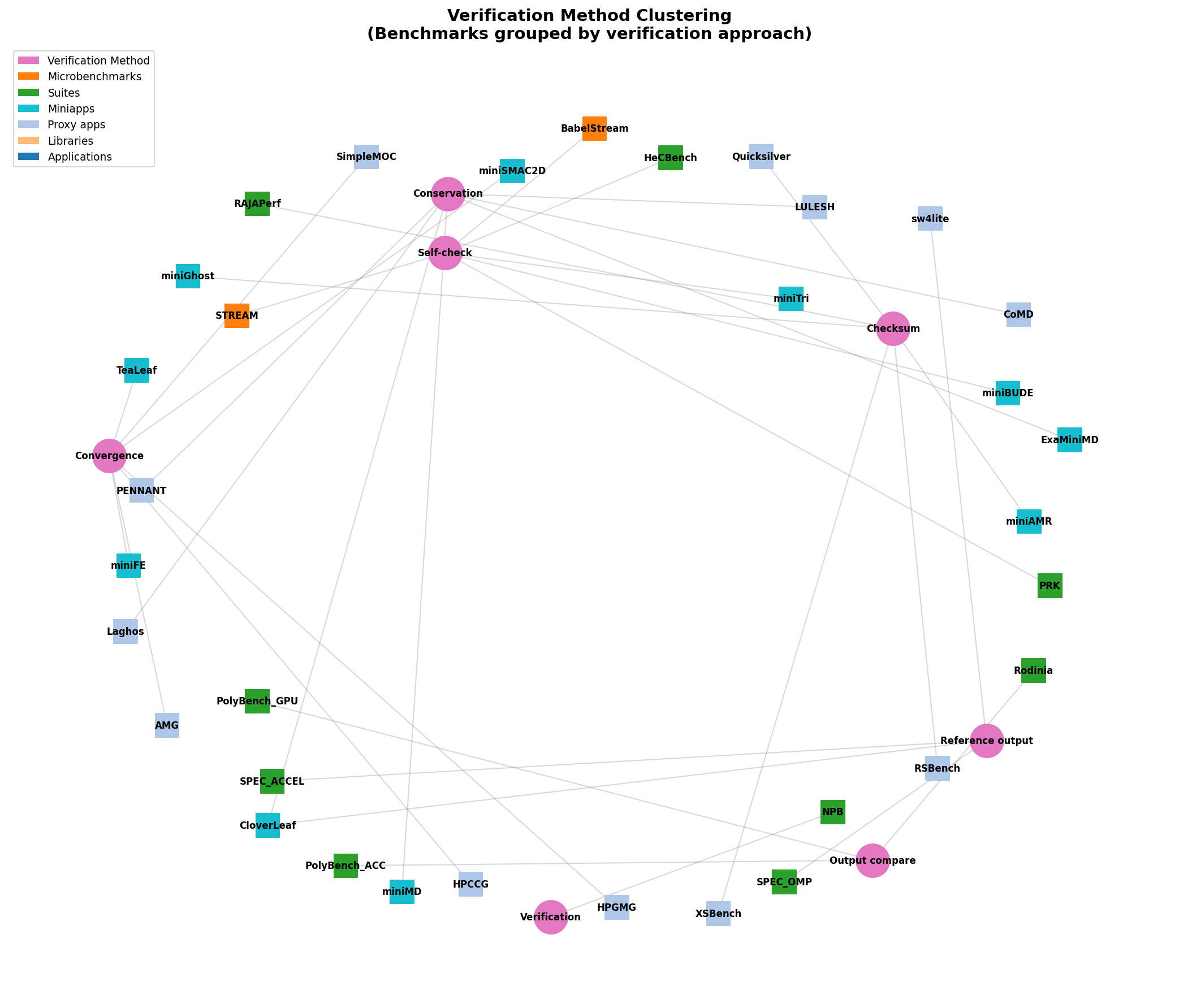}
    \caption{Benchmarks grouped by verification method. Large nodes represent
      verification approaches; benchmark nodes are connected to their method(s).
      Benchmarks without automated verification were excluded or downgraded to
      Tier~B (Section~\ref{sec:appendix-c2}). The checksum and reference output
      clusters contain the majority of Tier~A benchmarks.}
    \label{fig:verification-network}
  \end{figure}

  Figure~\ref{fig:verification-network} groups the surveyed benchmarks by their
  verification approach. The dominant clusters--checksum validation and reference
  output comparison--contain the majority of Tier~A benchmarks
  (Section~\ref{sec:appendix-c2}), while benchmarks relying solely on visual
  inspection or manual comparison were excluded or downgraded to Tier~B.

  \subsubsection*{Repository-Level Exclusions}

  Of the 71 surveyed repositories, 36 were excluded in two stages. First, 31 were
  removed during initial screening (fewer than three API implementations, no HPC
  kernel content, or outside the CUDA/OpenMP/OpenCL scope), leaving
  40~candidates. Of these, five were excluded during detailed evaluation:

  \begin{itemize}
    \item \textbf{Download failures} (2): Repository URLs were dead or
      access-restricted at the time of survey.
    \item \textbf{Insufficient documentation} (2): No build instructions, no
      Makefiles, or build system required proprietary dependencies not available on
      the evaluation platform.
    \item \textbf{Duplicate content} (1): One repository was a fork of an
      already-surveyed suite with no additional API variants.
  \end{itemize}

  \subsubsection*{Kernel-Level Exclusions}

  Within the selected suites, individual kernels were excluded for the following
  reasons:
  \begin{itemize}
    \item \textbf{Insufficient API coverage}: Kernels lacking the API variants
      needed for multi-direction translation evaluation. The completeness
      threshold is suite-specific: HeCBench required all four primary APIs
      (Stage~1 of the selection funnel, Section~\ref{sec:appendix-c3}); Rodinia
      required at least two of CUDA, OpenMP, and OpenCL.
    \item \textbf{No self-checking output}: Kernels that produce graphical output,
      require manual inspection, or have no deterministic expected output.
    \item \textbf{Excessive complexity}: Kernels requiring more than 15 source
      files (the HeCBench selection filter), external data files not included in
      the repository (e.g., Rodinia's \texttt{leukocyte} requires a separately
      downloaded video dataset and 102 external library files), or runtime
      dependencies (e.g., OpenGL for Rodinia's \texttt{hybridsort}, MPI
      multi-node) not available on the single-GPU evaluation platform.
    \item \textbf{Non-deterministic output}: Kernels whose output depends on
      execution order, random seeds without fixed initialization, or
      floating-point non-associativity that prevents consistent cross-API
      baseline matching. (Retained kernels with borderline FP divergence--such as
      \texttt{cfd}, \texttt{hotspot}, and \texttt{myocyte}--use relaxed
      stdout-pattern oracles rather than hash-based verification; see
      Section~\ref{sec:appendix-c4}.)
  \end{itemize}

  For kernels retained in the corpus but failing baseline verification
  (\knownfail{} specs), the specific failure reason and build output are recorded
  in each spec's \texttt{baseline\_results} field. For kernels excluded before
  spec creation (e.g., Rodinia's \texttt{leukocyte}), the exclusion is
  documented in the survey data and selection funnel
  (Sections~\ref{sec:appendix-c3}--\ref{sec:appendix-c4}).

\section{Augmentation Protocol and Extended Evaluation Results}
\label{sec:appendix-e}

This appendix collects the augmentation protocol details that would interrupt the
main text, then groups the result-heavy supporting evidence in the same order as
the main paper's empirical claims. The first subsection defines the L0--L4
levels used throughout the robustness analysis. The next subsection documents
transform frequency and baseline validation limits. The remaining subsections
cover a representative multi-file failure case and the extended result tables
that support the pass@$k$, augmentation, and kernel-difficulty claims in
Section~\ref{sec:results}.

\subsection{Augmentation Levels}
\label{sec:appendix-a-augmentation}

Table~\ref{tab:augmentation-levels} gives the exact L0--L4 augmentation
definitions used in the robustness protocol.

\begin{table}[htbp]
\centering
\caption{Augmentation level definitions.}
\label{tab:augmentation-levels}
\begin{tabular}{@{}cccp{4.2cm}@{}}
\toprule
Level & Transform selection & Candidate fraction & Description \\
\midrule
L0 & None & 0\% & Unmodified source code baseline \\
L1 & 1 transform & 1 site & Minimal source perturbation \\
L2 & 33\% of transforms & 33\% of candidates & Moderate perturbation \\
L3 & 66\% of transforms & 66\% of candidates & Heavy perturbation \\
L4 & All transforms & 100\% of candidates & Maximum configured perturbation \\
\bottomrule
\end{tabular}
\end{table}

\subsection{Augmentation Transform Frequency And Baseline Validation}
\label{sec:appendix-e1}

\begin{figure}[htbp]
  \centering
\begin{tikzpicture}
\begin{axis}[
  width=0.92\columnwidth,
  height=9cm,
  enlargelimits=false,
  axis on top,
  y dir=reverse,
  tick label style={font=\tiny},
  colorbar,
  colorbar style={
    font=\tiny, width=0.3cm,
    ylabel={\# Applications},
    ylabel style={font=\tiny},
  },
  colormap={ylOrRd}{
    rgb255(0)=(255,255,204)
    rgb255(14)=(254,217,118)
    rgb255(28)=(253,141,60)
    rgb255(42)=(227,26,28)
    rgb255(56)=(128,0,38)
  },
  point meta min=0, point meta max=48,
  xtick={0,...,5},
  xticklabels={Arith,ChgFnNames,ChgNames,{PtrArr{\ensuremath{\to}}Idx},SwapCond,TypedefExp},
  xticklabel style={font=\tiny, rotate=45, anchor=east},
  ytick={0,...,21},
  yticklabels={backprop,bfs,bptree,cfd,dwt2d,gaussian,heartwall,hotspot,hotspot3d,huffman,hybridsort,kmeans,lavamd,lud,mummergpu,myocyte,nn,nw,particlefilter,pathfinder,srad,streamcluster},
  yticklabel style={font=\tiny},
  xmin=-0.5, xmax=5.5,
  ymin=-0.5, ymax=21.5,
  title={Transform Application Frequency Per Kernel},
  title style={font=\small\bfseries},
]
  \addplot[matrix plot*, mesh/cols=6, point meta=explicit] coordinates {
    (0,0)[4]  (1,0)[0]  (2,0)[2]  (3,0)[0]  (4,0)[15]  (5,0)[0]
    (0,1)[1]  (1,1)[0]  (2,1)[4]  (3,1)[0]  (4,1)[12]  (5,1)[0]
    (0,2)[4]  (1,2)[0]  (2,2)[6]  (3,2)[0]  (4,2)[29]  (5,2)[0]
    (0,3)[13]  (1,3)[0]  (2,3)[0]  (3,3)[0]  (4,3)[21]  (5,3)[0]
    (0,4)[10]  (1,4)[0]  (2,4)[2]  (3,4)[0]  (4,4)[20]  (5,4)[0]
    (0,5)[5]  (1,5)[0]  (2,5)[3]  (3,5)[4]  (4,5)[12]  (5,5)[0]
    (0,6)[2]  (1,6)[0]  (2,6)[1]  (3,6)[0]  (4,6)[12]  (5,6)[0]
    (0,7)[0]  (1,7)[0]  (2,7)[1]  (3,7)[0]  (4,7)[10]  (5,7)[4]
    (0,8)[7]  (1,8)[0]  (2,8)[1]  (3,8)[0]  (4,8)[11]  (5,8)[0]
    (0,9)[6]  (1,9)[1]  (2,9)[6]  (3,9)[0]  (4,9)[14]  (5,9)[0]
    (0,10)[15]  (1,10)[0]  (2,10)[4]  (3,10)[0]  (4,10)[22]  (5,10)[0]
    (0,11)[6]  (1,11)[0]  (2,11)[13]  (3,11)[0]  (4,11)[26]  (5,11)[0]
    (0,12)[4]  (1,12)[0]  (2,12)[3]  (3,12)[0]  (4,12)[12]  (5,12)[0]
    (0,13)[10]  (1,13)[0]  (2,13)[6]  (3,13)[0]  (4,13)[26]  (5,13)[3]
    (0,14)[10]  (1,14)[0]  (2,14)[5]  (3,14)[2]  (4,14)[20]  (5,14)[0]
    (0,15)[7]  (1,15)[0]  (2,15)[10]  (3,15)[0]  (4,15)[48]  (5,15)[0]
    (0,16)[3]  (1,16)[0]  (2,16)[5]  (3,16)[0]  (4,16)[12]  (5,16)[0]
    (0,17)[3]  (1,17)[0]  (2,17)[1]  (3,17)[0]  (4,17)[10]  (5,17)[0]
    (0,18)[9]  (1,18)[0]  (2,18)[1]  (3,18)[0]  (4,18)[15]  (5,18)[0]
    (0,19)[0]  (1,19)[0]  (2,19)[2]  (3,19)[0]  (4,19)[9]  (5,19)[0]
    (0,20)[10]  (1,20)[0]  (2,20)[10]  (3,20)[0]  (4,20)[22]  (5,20)[0]
    (0,21)[5]  (1,21)[2]  (2,21)[2]  (3,21)[0]  (4,21)[10]  (5,21)[0]
  };
  \addplot[
    only marks, mark=none, point meta=explicit,
    nodes near coords={\pgfmathfloattoint{\pgfplotspointmeta}\pgfmathresult},
    nodes near coords align={center},
    every node near coord/.style={font=\tiny\bfseries, text=black, inner sep=0pt}
  ] coordinates {
    (0,0)[4]  (1,0)[0]  (2,0)[2]  (3,0)[0]  (4,0)[15]  (5,0)[0]
    (0,1)[1]  (1,1)[0]  (2,1)[4]  (3,1)[0]  (4,1)[12]  (5,1)[0]
    (0,2)[4]  (1,2)[0]  (2,2)[6]  (3,2)[0]  (4,2)[29]  (5,2)[0]
    (0,3)[13]  (1,3)[0]  (2,3)[0]  (3,3)[0]  (4,3)[21]  (5,3)[0]
    (0,4)[10]  (1,4)[0]  (2,4)[2]  (3,4)[0]  (4,4)[20]  (5,4)[0]
    (0,5)[5]  (1,5)[0]  (2,5)[3]  (3,5)[4]  (4,5)[12]  (5,5)[0]
    (0,6)[2]  (1,6)[0]  (2,6)[1]  (3,6)[0]  (4,6)[12]  (5,6)[0]
    (0,7)[0]  (1,7)[0]  (2,7)[1]  (3,7)[0]  (4,7)[10]  (5,7)[4]
    (0,8)[7]  (1,8)[0]  (2,8)[1]  (3,8)[0]  (4,8)[11]  (5,8)[0]
    (0,9)[6]  (1,9)[1]  (2,9)[6]  (3,9)[0]  (4,9)[14]  (5,9)[0]
    (0,10)[15]  (1,10)[0]  (2,10)[4]  (3,10)[0]  (4,10)[22]  (5,10)[0]
    (0,11)[6]  (1,11)[0]  (2,11)[13]  (3,11)[0]  (4,11)[26]  (5,11)[0]
    (0,12)[4]  (1,12)[0]  (2,12)[3]  (3,12)[0]  (4,12)[12]  (5,12)[0]
    (0,13)[10]  (1,13)[0]  (2,13)[6]  (3,13)[0]  (4,13)[26]  (5,13)[3]
    (0,14)[10]  (1,14)[0]  (2,14)[5]  (3,14)[2]  (4,14)[20]  (5,14)[0]
    (0,15)[7]  (1,15)[0]  (2,15)[10]  (3,15)[0]  (5,15)[0]
    (0,16)[3]  (1,16)[0]  (2,16)[5]  (3,16)[0]  (4,16)[12]  (5,16)[0]
    (0,17)[3]  (1,17)[0]  (2,17)[1]  (3,17)[0]  (4,17)[10]  (5,17)[0]
    (0,18)[9]  (1,18)[0]  (2,18)[1]  (3,18)[0]  (4,18)[15]  (5,18)[0]
    (0,19)[0]  (1,19)[0]  (2,19)[2]  (3,19)[0]  (4,19)[9]  (5,19)[0]
    (0,20)[10]  (1,20)[0]  (2,20)[10]  (3,20)[0]  (4,20)[22]  (5,20)[0]
    (0,21)[5]  (1,21)[2]  (2,21)[2]  (3,21)[0]  (4,21)[10]  (5,21)[0]
  };
  \addplot[
    only marks, mark=none, point meta=explicit,
    nodes near coords={\pgfmathfloattoint{\pgfplotspointmeta}\pgfmathresult},
    nodes near coords align={center},
    every node near coord/.style={font=\tiny\bfseries, text=white, inner sep=0pt}
  ] coordinates {
    (4,15)[48]
  };
\end{axis}
\end{tikzpicture}
  \caption{Per-kernel transform application frequency across the Rodinia corpus.
    Rows represent the 22~Rodinia kernels; columns represent the six AST-level
    transforms. Cell values give the total number of times each transform was
    applied to files belonging to that kernel, aggregated across all
    augmentation levels and API variants. The distribution is highly
    non-uniform: multi-file kernels such as \texttt{myocyte} (16 source files)
    and \texttt{bptree} accumulate many SwapCondition applications, while
    smaller kernels (e.g., \texttt{pathfinder}, \texttt{hotspot}) show lower
    counts. ChangeFunctionNames applies to only two kernels
    (\texttt{huffman}, \texttt{streamcluster}).}
  \label{fig:transform-freq}
\end{figure}
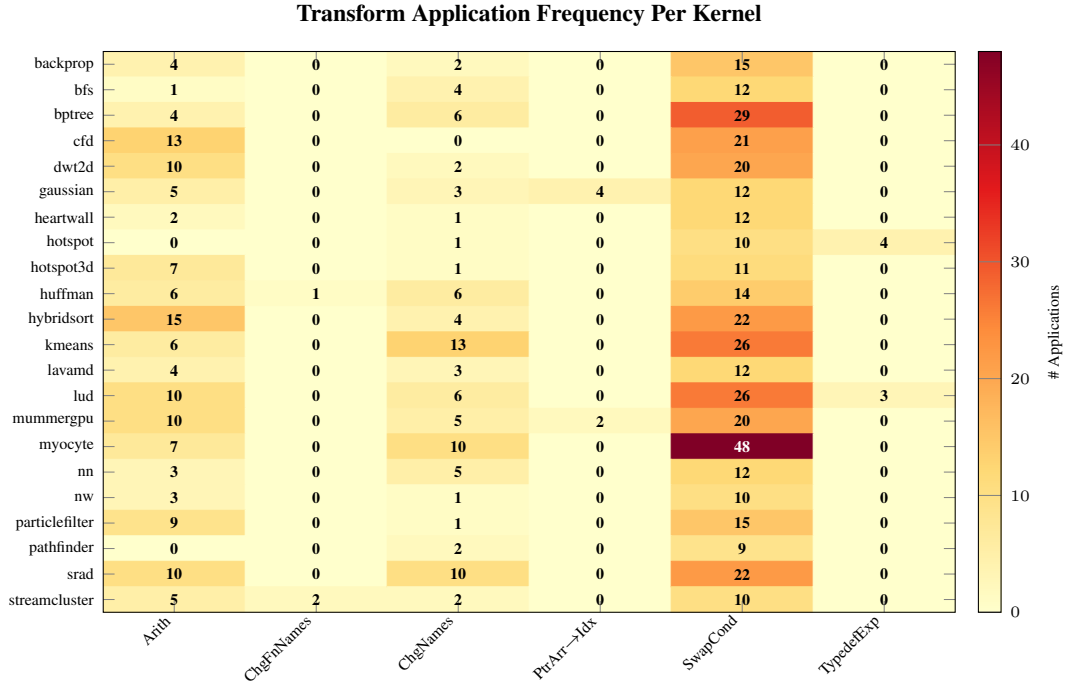

\parbench{}'s code augmentation pipeline applies six AST-level transforms at four
intensity levels (L1--L4). The transforms do not apply uniformly: some kernels
have many candidate sites for a given transform, while others have few or none.
Figure~\ref{fig:transform-freq} presents the per-kernel, per-transform frequency
heatmap.

Aggregate transform frequencies across the corpus reveal a clear dominance hierarchy. Below we report the number of Rodinia specs (out of 60) to which each transform applies at L4 (full intensity), followed by the total application count across all 240 augmentation tasks (60~specs $\times$ 4~levels):
  
\begin{itemize}
  \item \textbf{SwapCondition} (59/60 specs at L4; 162 applications across all levels): The most broadly applicable transform. Swaps operands of comparison operators (e.g., \texttt{a < b} $\rightarrow$ \texttt{b > a}), with guards against expressions containing side effects. Its near-universal coverage reflects the ubiquity of comparison-based control flow in HPC kernels--boundary checks, convergence tests, and branch-based algorithm selection all contribute candidate sites.
  
  \item \textbf{ArithmeticTransform} (45/60 specs at L4; 69 applications): Converts compound assignment operators to their expanded forms and vice versa (e.g., \texttt{x += expr} $\leftrightarrow$ \texttt{x = x + (expr)}). Expansion applies to integer and floating-point scalar targets; folding back to compound form is restricted to integer-typed scalars to avoid floating-point precedence issues. For-loop incrementors are excluded to avoid disrupting iteration patterns. Less broadly applicable than SwapCondition because it requires compound-assignment or assignment-with-matching-LHS patterns that not all kernels exhibit.
  \item \textbf{ChangeNames} (32/60 specs at L4; 55 applications): Consistently renames local variables and function parameters using symbol-aware reference collection, with behavior-preserving intent validated against the retained baseline subset. Coverage is bounded by the number of renameable local declarations per kernel.
  
  \item \textbf{TypedefExpansion} (3/60 specs at L4; 7 applications): Inlines \texttt{typedef} aliases to their underlying types. Rare because most Rodinia kernels use primitive types directly rather than defining type aliases.
  
  \item \textbf{PointerArithmeticToArrayIndex} (3/60 specs at L4; 6 applications): Rewrites \texttt{*(ptr + i)} to \texttt{ptr[i]} and vice versa, with precedence-aware handling of struct member access. Limited by the prevalence of pointer arithmetic in the corpus--most kernels use array indexing natively.
  
  \item \textbf{ChangeFunctionNames} (2/60 specs at L4; 2 applications): Renames user-defined functions and updates all call sites, skipping \texttt{main}, GPU kernel entry points, and identifiers appearing in string literals. The rarest transform, reflecting the small number of user-defined functions in single-kernel files.
\end{itemize}

The level-invariance property supports that augmentation introduces no new baseline failures on the retained Rodinia subset: all 53 non-\knownfail{} Rodinia specs pass at every augmentation level L1--L4. (Note that the per-transform applicability fractions above, e.g., 59/60 for SwapCondition, are computed over all 60~Rodinia specs including the 7~\knownfail{} specs, since augmentation is applied to source files regardless of baseline status; baseline validation uses only the 53 non-\knownfail{} specs.) All transforms are designed to preserve behavior, supported by 15~unit tests covering each transform in isolation and by the end-to-end augmentation sweep across the full Rodinia corpus.

Across all five suites, 80 of 87 non-\knownfail{} specs pass at all levels L1--L4, with the 7~failures confined to \texttt{omp\_target} variants. Five of these (all HeCBench kernels) fail at Level~3 with SwapCondition as the sole active transform, directly implicating condition operand swapping (e.g., \texttt{a~<~b}~$\to$~\texttt{b~>~a}) in causing the GPU-offloading compiler to generate numerically divergent device code. The remaining two (\texttt{rsbench}, \texttt{xsbench}) fail only at Level~4 where both increased SwapCondition intensity and the addition of ArithmeticTransform coincide, making precise attribution ambiguous. Variable renaming is separately disabled for any file containing \texttt{\#pragma~omp} directives, so it plays no role in these failures. All 87 specs pass at L1--L2. Because these baseline failures are specific to the \texttt{omp\_target} toolchain, the affected variants are excluded from robustness claims about \texttt{omp\_target} baselines and still serve only as prompt inputs when translating to other targets.

\subsection{Case Study: Multi-File Translation Consistency Failure}
\label{sec:appendix-e3}

Multi-file kernel structures expose direction-dependent translation failures that single-file kernels do not exhibit. We document these through a detailed case study of the \texttt{rodinia-lavamd} kernel, a molecular dynamics simulation with a multi-file build structure (kernel, wrapper, and supporting headers). \parbench{} evaluates \texttt{lavamd} in both full-program mode (CUDA/OMP targets, where the LLM rewrites kernel and host code) and kernel-only mode (OpenCL targets, where only the \texttt{.cl} compute kernel is translated), revealing distinct failure mechanisms in each.

\subsubsection*{Observation}

Across all six standard translation directions, \qwenshort{} fails every \texttt{lavamd} translation: 0 of 18~records (6~directions $\times$ 3~independent samples at temperature~0.7) produce a \pass{} outcome. The failure mode is not uniform: it varies systematically by translation direction rather than by stochastic sample, revealing that the failure mechanism is structurally determined by the source--target API pairing.

\subsubsection*{Per-Direction Error Taxonomy}

\begin{table}[htbp]
  \centering
  \caption{Translation error taxonomy for \texttt{lavamd} across all
  six standard directions (\qwenshort{}, L0, 3~samples each). Full-program
  directions (CUDA/OMP targets) fail deterministically at build time;
  kernel-only directions (OpenCL targets) fail at runtime. Five of six
  directions produce identical failure modes across all three samples.}
  \label{tab:lavamd-errors}
  \small
  \begin{tabular}{@{}llp{4.5cm}@{}}
    \toprule
    Direction & Status ($\times$3) & Primary Failure Mode \\
    \midrule
    OpenCL$\to$CUDA & \buildfail{} & Phantom include (\texttt{timing.h}) \\
    OMP$\to$CUDA    & \buildfail{} & Phantom include (\texttt{./util/timer/timer.h}) \\
    CUDA$\to$OMP    & \buildfail{} & Phantom include (\texttt{./main.h}) \\
    CUDA$\to$OpenCL & \runfail{}   & OpenCL kernel compilation failure \\
    OMP$\to$OpenCL  & \runfail{}   & OpenCL kernel compilation failure \\
    OpenCL$\to$OMP  & Mixed        & 2~\buildfail{} + 1~\verifyfail{} \\
    \bottomrule
  \end{tabular}
\end{table}

Table~\ref{tab:lavamd-errors} reveals two structural patterns corresponding to \parbench{}'s translation modes. First, three of four full-program directions (targeting CUDA or OMP, where the LLM rewrites both kernel and host code) fail uniformly at build time due to phantom include paths: the model generates headers (\texttt{timing.h}, \texttt{./main.h}, \texttt{./util/timer/timer.h}) that exist in the source program's broader directory tree but are unavailable in the translated program's build context. All three samples within each of these directions fail on the same phantom header, indicating a deterministic error rather than stochastic variation. The fourth full-program direction (OpenCL$\to$OMP) exhibits mixed failure modes across its three samples--including a phantom include, an OpenMP barrier-nesting violation, and a verification failure where the program runs to completion (exit code~0) but produces only configuration output, no computation results. Second, the two kernel-only directions (targeting OpenCL, where only the \texttt{.cl} kernel file is translated while the host driver is untouched) pass host-code compilation but fail at runtime when \texttt{clBuildProgram} attempts to compile the translated kernel source (exit code~245 = \texttt{CL\_BUILD\_PROGRAM\_FAILURE}). The runtime compilation errors include incorrect \texttt{\_\_local} variable scoping, undeclared identifiers (e.g., \texttt{fp}, \texttt{NUMBER\_THREADS}), and missing address space qualifiers on kernel pointer parameters--API-specific constraints that the model fails to satisfy. Cross-model comparison reveals that these barriers are not uniformly intractable: \gptnew{} passes 5 of 18~records by inlining header definitions (avoiding phantom includes) and producing harness-passing OMP$\to$OpenCL kernels, while \codex{} passes 3 of 18 on the same kernel-only direction. The full-program CUDA-targeting directions, however, remain unsolved across all three models.

\subsubsection*{Cross-Model Persistence}
The direction-dependent failure pattern persists across models. Only two of
six directions yield any \pass{} outcomes across the three models:
OMP$\to$OpenCL (\gptnew{} and \codex{} each 3/3~\pass{}, \qwenshort{}
0/3) and CUDA$\to$OMP (\gptnew{} 2/3~\pass{}, \qwenshort{} and \codex{}
0/3). The remaining four directions are universally failed across all models
and samples (0/36~\pass{}). OpenCL$\to$CUDA and OMP$\to$CUDA fail at build
time due to phantom includes (18/18~\buildfail{}), CUDA$\to$OpenCL fails at
runtime due to OpenCL kernel compilation errors (9/9~\runfail{}), and
OpenCL$\to$OMP fails with a mix of phantom includes and output mismatches.
These asymmetries indicate that multi-file coordination difficulty is a
property of the translation direction--not solely of the kernel complexity
or the model.

\subsubsection*{Implications} 
This case study identifies three distinct error subcategories within the ``multi-file translation consistency failure'' class:
\begin{enumerate}
  \item \textbf{Phantom dependencies}: The model hallucinates include files (\texttt{timing.h}, \texttt{./main.h}) or helper functions (\texttt{checkCUDAError}, \texttt{setdevice}) that may exist in the model's training data but are absent from the target program's file structure. This is the dominant build-time failure mode for CUDA-targeting directions across all three models.
    
    \item \textbf{Cross-file identifier inconsistency}: In some translations (observed in \codex{} and \gptnew{} OpenCL$\to$CUDA attempts), the kernel file uses the target API's naming convention while the wrapper file retains the source API's function name, producing linker errors (\texttt{undefined reference to `kernel\_gpu\_cuda'}).
    
    \item \textbf{Runtime kernel compilation failure}: For OpenCL-targeting directions, the host C/C++ code compile successfully, but the OpenCL kernel source, compiled at runtime by \texttt{clBuildProgram}, contains errors such as incorrect \texttt{\_\_local} variable scoping and undeclared type aliases. This represents a disconnect between the C/C++ compiler's view of the host code and the OpenCL runtime compiler's requirements for device code, a failure mode invisible to static build checks.
\end{enumerate}

These findings suggest that multi-file translation failures are not uniformly distributed across directions. Directions requiring the model to generate host-side boilerplate for a fundamentally different runtime (e.g., CUDA driver API, OpenCL \texttt{cl*} calls) are systematically harder than directions where the host code structure is similar between source and target. A hybrid pipeline combining LLM translation for computational kernels with template-based scaffold generation for host infrastructure may address subcategories~1 and~2 but would not resolve the runtime compilation failures in subcategory~3, which require translated API-specific device-code constraints to satisfy the runtime compiler.

\subsection{Extended Results Details}
\label{sec:appendix-e4}

This section expands on the results presented in Section~\ref{sec:results}.
Across all three models, we collected 2,262 valid translation records: 626 for \qwenshort{} (426~L0 + 200~augmentations, after excluding 82 records involving 9~\knownfail{} specs\footnote{The \qwenshort{} total is 626 rather than 630 because four \texttt{rodinia-backprop-opencl} L1--L4 augmentation records are also excluded via the \knownfail{} policy (the OpenCL baseline is silently broken).}), 822 for \gptnew{} (426~L0 + 396~augmentations), and 814 for \codex{} (426~L0 + 388~augmentations). \knownfail{} specs were pre-excluded from the \gptnew{} and \codex{} evaluation batches.

\paragraph{Record-Level Diagnostic Summary.}
\begin{table*}[h]
\centering
\caption{Record-level diagnostic pass rates across three models and 2,262 valid translation records. Records include L0 samples and L1--L4 augmentation runs for L0-passing pairs, so this table mixes canonical L0 records with model-specific L0-conditional augmentation records. Wilson intervals are descriptive because records are clustered by task; inferential comparison uses the balanced 142 L0 pairs in Section~\ref{sec:passk-analysis}.}
\label{tab:overall-pass}
\small
\resizebox{\textwidth}{!}{%
\begin{tabular}{@{}lrrrrrrl@{}}
\toprule
Model & \pass{} & \buildfail{} & \runfail{} & \verifyfail{} & \extractionfail{} & Total & Rate [95\% Wilson CI] \\
\midrule
\qwen{} & 230 & 245 & 121 & 29 & 1 & 626 & 36.7\% [33.1\%, 40.6\%] \\
\gptnew{} & 621 & 123 & 43 & 32 & 3 & 822 & 75.5\% [72.5\%, 78.4\%] \\
\codex{} & 604 & 139 & 44 & 27 & 0 & 814 & 74.2\% [71.1\%, 77.1\%] \\
\bottomrule
\end{tabular}%
}
\end{table*}

\paragraph{Direction-Level Pass Rates.}
\begin{table*}[h]
\centering
\caption{Pass rates by translation direction (L0 records, all three samples per task). Six standard directions form the core matrix; four OMP-target directions are case studies with fewer participating kernels ($m$). Here $n$ denotes L0 records per model for that direction. Wilson 95\% CIs in brackets.}
\label{tab:direction-rates}
\small
\resizebox{\textwidth}{!}{%
\begin{tabular}{@{}lrrrc@{}}
\toprule
Direction & \qwenshort{} & \gptnew{} & \codex{} & $n$ \\
\midrule
CUDA$\to$OMP & 40.3\% [29.7\%, 51.8\%] & 83.3\% [73.1\%, 90.2\%] & 76.4\% [65.4\%, 84.7\%] & 72 \\
OMP$\to$OpenCL & 33.3\% [22.0\%, 47.0\%] & 72.5\% [59.1\%, 82.9\%] & 82.4\% [69.8\%, 90.4\%] & 51 \\
OMP$\to$CUDA & 25.0\% [16.4\%, 36.1\%] & 55.6\% [44.1\%, 66.5\%] & 55.6\% [44.1\%, 66.5\%] & 72 \\
OpenCL$\to$OMP & 9.8\% [4.3\%, 21.0\%] & 41.2\% [28.8\%, 54.8\%] & 39.2\% [27.0\%, 52.9\%] & 51 \\
CUDA$\to$OpenCL & 7.0\% [2.8\%, 16.7\%] & 59.7\% [46.7\%, 71.4\%] & 57.9\% [45.0\%, 69.8\%] & 57 \\
OpenCL$\to$CUDA & 0.0\% [0.0\%, 6.3\%] & 19.3\% [11.1\%, 31.3\%] & 19.3\% [11.1\%, 31.3\%] & 57 \\
\midrule
OMP-target$\to$OMP ($m{=}3$) & 100\% [70.1\%, 100\%] & 100\% [70.1\%, 100\%] & 100\% [70.1\%, 100\%] & 9 \\
OMP-target$\to$CUDA ($m{=}8$) & 66.7\% [46.7\%, 82.0\%] & 95.8\% [79.8\%, 99.3\%] & 100\% [86.2\%, 100\%] & 24 \\
OMP$\to$OMP-target ($m{=}3$) & 44.4\% [18.9\%, 73.3\%] & 100\% [70.1\%, 100\%] & 100\% [70.1\%, 100\%] & 9 \\
CUDA$\to$OMP-target ($m{=}8$) & 0.0\% [0.0\%, 13.8\%] & 95.8\% [79.8\%, 99.3\%] & 100\% [86.2\%, 100\%] & 24 \\
\bottomrule
\end{tabular}%
}
\end{table*}

\paragraph{pass@k Analysis.}
\begin{table}[!htbp]
\centering
\caption{pass@$k$ results for \qwenshort{} (L0 only, temperature~0.7, three independent samples per task, 142 unique source-target pairs). CIs are normal-approximation intervals on the macro-averaged Chen et al.\ estimator (distinct from the Wilson binomial CIs used for per-record pass rates in Table~\ref{tab:overall-pass})}
\label{tab:pass-at-k-main}
\small
\begin{tabular}{@{}lr@{}}
\toprule
Metric & Value [95\% CI] \\
\midrule
pass@1 & 23.9\% [17.9\%, 30.0\%] \\
pass@3 & 35.2\% [27.3\%, 43.1\%] \\
\midrule
Always pass (3/3) & 19 tasks (13.4\%) \\
Noisy (1--2/3) & 31 tasks (21.8\%) \\
Hard fail (0/3) & 92 tasks (64.8\%) \\
\bottomrule
\end{tabular}
\end{table}
The 36.7\% per-record rate (Table~\ref{tab:overall-pass}) treats each sample and augmentation level independently. The pass@$k$ analysis (Table~\ref{tab:pass-at-k-main}) restricts to the 142~L0 tasks and macro-averages over tasks to estimate single-sample and three-sample success probabilities, with each task generating three independent samples at temperature~0.7. The gap between pass@1 (23.9\%) and pass@3 (35.2\%) is 11.3\%: only 31 of 142 tasks show within-task variability (1 or 2 of 3 samples pass); none of the 92 hard failures (0/3) are recovered by resampling. This confirms that the majority of failures reflect systematic translation challenges--structural mismatches between parallel programming models--rather than stochastic variation, validating \parbench{}'s design as a benchmark that surfaces genuine capability gaps.

\paragraph{Cross-Model pass@$k$ Summary.}
Table~\ref{tab:pass-at-k} extends the single-model view above to all three model
campaigns on the shared 142-task L0 set.

\begin{table*}[t]
\centering
\caption{Pass@$k$ estimates for the shared L0 task set (142 evaluation-eligible
direction--kernel pairs, three independent samples per pair). Hard Fail = 0/3
samples pass; Always Pass = 3/3; Noisy = 1--2/3.}
\label{tab:pass-at-k}
\small
\begin{tabular}{@{}lrrrrrr@{}}
\toprule
Model & Pairs & pass@1 & pass@3 & Hard Fail & Noisy & Always Pass \\
\midrule
\qwenshort{} & 142 & 23.9\% & 35.2\% & 92/142 (64.8\%) & 31/142 (21.8\%) & 19/142 (13.4\%) \\
\gptnew{} & 142 & 62.7\% & 69.7\% & 43/142 (30.3\%) & 23/142 (16.2\%) & 76/142 (53.5\%) \\
\codex{} & 142 & 62.7\% & 68.3\% & 45/142 (31.7\%) & 16/142 (11.3\%) & 81/142 (57.0\%) \\
\bottomrule
\end{tabular}
\end{table*}

The cross-model summary in Table~\ref{tab:pass-at-k} shows that \gptnew{} and
\codex{} share the same pass@1 while differing slightly in consistency:
\codex{} has more always-pass pairs and fewer noisy pairs, whereas
\qwenshort{} concentrates much more mass in hard failures. Across all three
models, most failed L0 tasks remain unrecovered by resampling.

\paragraph{Representative Per-Kernel Pass Rates.}
Table~\ref{tab:per-kernel-main} highlights the easiest and hardest kernels for \qwenshort{}.
\begin{table}[h]
\centering
\caption{Representative per-kernel pass rates for \qwenshort{} across all directions and augmentation levels (626 records, 31~evaluated kernels--4~curated kernels have no evaluable translation pairs). \texttt{nqueen} is tied with \texttt{page-rank} at 60.0\%.}
\label{tab:per-kernel-main}
\small
\begin{tabular}{@{}llrrrr@{}}
\toprule
Suite & Kernel & Total & \pass{} & Fail & Rate \\
\midrule
\multicolumn{6}{l}{\textit{Easiest kernels}} \\
HeCBench & iso2dfd & 38 & 32 & 6 & 84.2\% \\
HeCBench & floydwarshall & 38 & 29 & 9 & 76.3\% \\
HeCBench & heat2d & 38 & 29 & 9 & 76.3\% \\
HeCBench & stencil1d & 14 & 9 & 5 & 64.3\% \\
HeCBench & page-rank & 10 & 6 & 4 & 60.0\% \\
\midrule
\multicolumn{6}{l}{\textit{Hardest kernels (0\% pass rate, $n \geq 18$)}} \\
Rodinia & heartwall & 18 & 0 & 18 & 0.0\% \\
Rodinia & myocyte & 18 & 0 & 18 & 0.0\% \\
Rodinia & bptree & 18 & 0 & 18 & 0.0\% \\
RSBench & rsbench & 18 & 0 & 18 & 0.0\% \\
Rodinia & lavamd & 18 & 0 & 18 & 0.0\% \\
\bottomrule
\end{tabular}
\end{table}

\paragraph{Augmentation Pass Rates.}
\begin{table}[h]
\centering
\caption{Augmentation pass rates on the common 12-kernel CUDA-to-OpenMP subset present at all five levels. These 12 kernels were selected by the L0-conditional filter ($\geq$1 of 3 L0 samples passes), so L0 rates exceed the overall CUDA-to-OpenMP rate of 40.3\%.}
\label{tab:aug-balanced}
\small
\begin{tabular}{@{}lrrl@{}}
\toprule
Level & Pass / Total & Rate & Kernels with any pass \\
\midrule
L0 & 29/36 & 80.6\% & 12/12 (100\%) \\
L1 &  9/12 & 75.0\% &  9/12 \phantom{0}(75\%) \\
L2 & 10/12 & 83.3\% & 10/12 \phantom{0}(83\%) \\
L3 &  9/12 & 75.0\% &  9/12 \phantom{0}(75\%) \\
L4 & 10/12 & 83.3\% & 10/12 \phantom{0}(83\%) \\
\bottomrule
\end{tabular}
\end{table}

If baseline success were driven primarily by surface-form memorization, augmented source variants should degrade pass rates. On a balanced 12-kernel CUDA-to-OpenMP subset present at all five augmentation levels (Table~\ref{tab:aug-balanced}), pass rates remain between 75\% and 83\% at L1--L4. The L0 rate on this subset (80.6\%) exceeds the overall CUDA-to-OpenMP L0 rate (40.3\%) because the L0-conditional filter selects kernels that already pass at L0, creating a survivorship bias toward easier kernels. Restricting the Cochran--Armitage trend test to L1--L4 only (avoiding the 3:1 L0 sample-size imbalance and L0-conditional ceiling), \qwenshort{} shows $z = -1.84$, $p = 0.065$--suggestive of a decreasing trend but not significant at $\alpha = 0.05$. \gptnew{} and \codex{} show no significant L1--L4 trend (both plateau between 85\% and 91\%). The survivorship selection inherent in L0-conditional filtering still applies; these results characterize the qualifying subset only and do not rule out memorization at higher levels of abstraction (e.g., algorithmic templates).

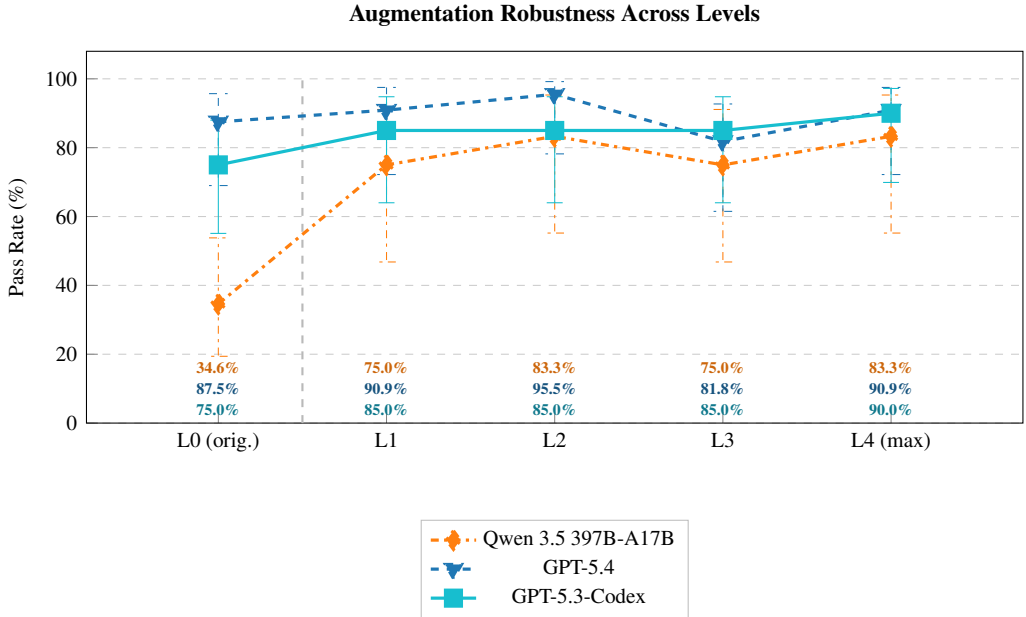
\begin{figure}[t]
\centering
%

\begin{tikzpicture}
\begin{axis}[
  parbench compact,
  width=\columnwidth,
  height=6.5cm,
  ymin=0, ymax=108,
  xmin=-0.4, xmax=4.4,
  xtick={0,1,2,3,4},
  xticklabels={
    {L0 (orig.)},
    {L1},
    {L2},
    {L3},
    {L4 (max)}
  },
  ytick={0,20,40,60,80,100},
  yticklabels={0,20,40,60,80,100},
  ylabel={Pass Rate (\%)},
  ymajorgrids=true,
  grid style={dashed, draw=gray!40},
  tick label style={font=\footnotesize},
  label style={font=\footnotesize},
  title style={font=\small\bfseries, align=center},
  title={Augmentation Robustness Across Levels},
  legend style={
    at={(0.5,-0.26)}, anchor=north,
    legend columns=1,
    font=\footnotesize,
    draw=gray!50,
  },
  clip=false,
]

\draw[dashed, gray!55, line width=0.8pt] (axis cs:0.5, 0) -- (axis cs:0.5, 100);

\addplot[
  color=pbOrange,
  mark=diamond*,
  mark size=3.5pt,
  line width=1.2pt,
  dashdotted,
  error bars/.cd, y dir=both, y explicit,
]
table[x=lv, y=qw, y error plus=qw_hi, y error minus=qw_lo] {figures/figures_tek_version/f7_augmentation_robustness.dat};
\addlegendentry{Qwen~3.5 397B-A17B}

\addplot[
  color=pbRodinia,
  mark=triangle*,
  mark options={rotate=180},
  mark size=3.5pt,
  line width=1.2pt,
  dashed,
  error bars/.cd, y dir=both, y explicit,
]
table[x=lv, y=g54, y error plus=g54_hi, y error minus=g54_lo] {figures/figures_tek_version/f7_augmentation_robustness.dat};
\addlegendentry{GPT-5.4}

\addplot[
  color=pbTeal,
  mark=square*,
  mark size=3pt,
  line width=1.2pt,
  error bars/.cd, y dir=both, y explicit,
]
table[x=lv, y=cdx, y error plus=cdx_hi, y error minus=cdx_lo] {figures/figures_tek_version/f7_augmentation_robustness.dat};
\addlegendentry{GPT-5.3-Codex}

\node[font=\tiny\bfseries, color=pbOrange!80!black,  anchor=center] at (axis cs:0, 16) {34.6\%};
\node[font=\tiny\bfseries, color=pbOrange!80!black,  anchor=center] at (axis cs:1, 16) {75.0\%};
\node[font=\tiny\bfseries, color=pbOrange!80!black,  anchor=center] at (axis cs:2, 16) {83.3\%};
\node[font=\tiny\bfseries, color=pbOrange!80!black,  anchor=center] at (axis cs:3, 16) {75.0\%};
\node[font=\tiny\bfseries, color=pbOrange!80!black,  anchor=center] at (axis cs:4, 16) {83.3\%};

\node[font=\tiny\bfseries, color=pbRodinia!70!black, anchor=center] at (axis cs:0, 10) {87.5\%};
\node[font=\tiny\bfseries, color=pbRodinia!70!black, anchor=center] at (axis cs:1, 10) {90.9\%};
\node[font=\tiny\bfseries, color=pbRodinia!70!black, anchor=center] at (axis cs:2, 10) {95.5\%};
\node[font=\tiny\bfseries, color=pbRodinia!70!black, anchor=center] at (axis cs:3, 10) {81.8\%};
\node[font=\tiny\bfseries, color=pbRodinia!70!black, anchor=center] at (axis cs:4, 10) {90.9\%};

\node[font=\tiny\bfseries, color=pbTealDark,         anchor=center] at (axis cs:0, 4) {75.0\%};
\node[font=\tiny\bfseries, color=pbTealDark,         anchor=center] at (axis cs:1, 4) {85.0\%};
\node[font=\tiny\bfseries, color=pbTealDark,         anchor=center] at (axis cs:2, 4) {85.0\%};
\node[font=\tiny\bfseries, color=pbTealDark,         anchor=center] at (axis cs:3, 4) {85.0\%};
\node[font=\tiny\bfseries, color=pbTealDark,         anchor=center] at (axis cs:4, 4) {90.0\%};

\end{axis}
\end{tikzpicture}
\caption{L0-conditional augmentation outcomes across all three models for the
CUDA-to-OpenMP direction. L0 uses first-sample results across all kernels;
L1--L4 are restricted to each model's qualifying subset. Error bars show Wilson
95\% intervals and are descriptive because the subset differs by model.}
\label{fig:augmentation}
\end{figure}

\begin{figure}[t]
  \centering
  \input{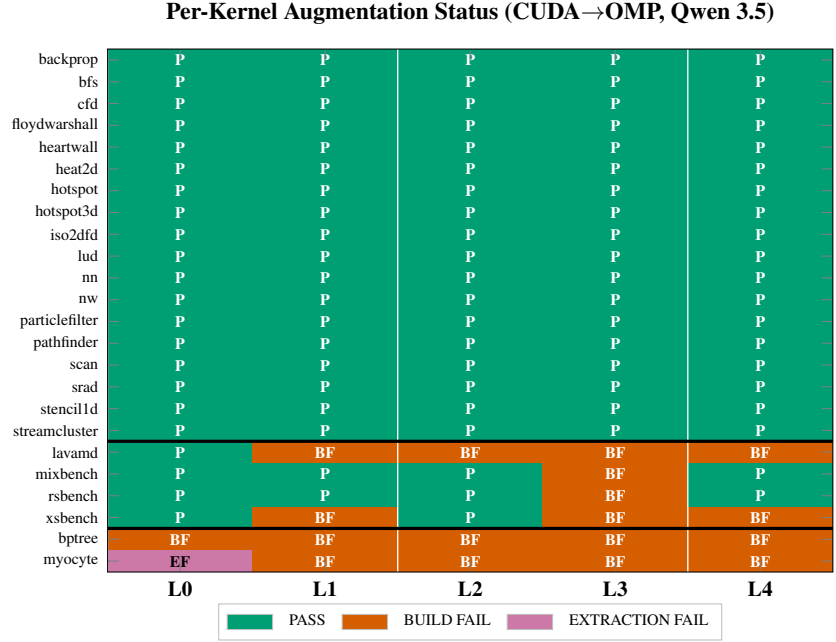}
  \caption{Per-kernel L0/L1--L4 outcomes for the CUDA-to-OpenMP subset under
  \qwenshort{}. Cell color indicates whether the corresponding analysis record
  receives declared-oracle \pass{} or a failure outcome at that level.}
  \label{fig:aug-heatmap}
\end{figure}

\paragraph{All-Model Augmentation Summary.}
Table~\ref{tab:augmentation-rates} and Figures~\ref{fig:augmentation}--\ref{fig:aug-heatmap}
extend the balanced-subset view to all three model campaigns. These summaries
remain descriptive because the L0-conditional filter changes the denominator
across models and levels.

\begin{table}[t]
  \centering
  \caption{Declared-oracle pass rates across augmentation levels. L0 covers all
  142 evaluation-eligible direction--kernel pairs ($n=426$ L0 records, three
  samples each). L1--L4 use one augmentation record per qualifying pair on
  model-specific L0-conditional subsets: pairs where at least one of three L0
  samples passes (\qwenshort{}: 50 pairs, \gptnew{}: 99 pairs, \codex{}: 97
  pairs).}
  \label{tab:augmentation-rates}
  \small
  \resizebox{\textwidth}{!}{%
  \begin{tabular}{@{}ccccccc@{}}
  \toprule
  Level & \makecell{\qwenshort{}\\(all dirs)} & \makecell{\qwenshort{}\\(C$\to$OMP)} & \makecell{\gptnew{}\\(all dirs)} & \makecell{\gptnew{}\\(C$\to$OMP)} & \makecell{\codex{}\\(all dirs)} & \makecell{\codex{}\\(C$\to$OMP)} \\
  \midrule
  L0 & 23.9\% ($n$=426) & 40.3\% ($n$=72) & 62.7\% ($n$=426) & 83.3\% ($n$=72) & 62.7\% ($n$=426) & 76.4\% ($n$=72) \\
  L1 & 74.0\% ($n$=50) & 75.0\% ($n$=12) & 88.9\% ($n$=99) & 90.9\% ($n$=22) & 86.6\% ($n$=97) & 85.0\% ($n$=20) \\
  L2 & 64.0\% ($n$=50) & 83.3\% ($n$=12) & 90.9\% ($n$=99) & 95.5\% ($n$=22) & 88.7\% ($n$=97) & 85.0\% ($n$=20) \\
  L3 & 62.0\% ($n$=50) & 75.0\% ($n$=12) & 86.9\% ($n$=99) & 81.8\% ($n$=22) & 86.6\% ($n$=97) & 85.0\% ($n$=20) \\
  L4 & 56.0\% ($n$=50) & 83.3\% ($n$=12) & 90.9\% ($n$=99) & 90.9\% ($n$=22) & 85.6\% ($n$=97) & 90.0\% ($n$=20) \\
  \bottomrule
  \end{tabular}%
  }
\end{table}

Table~\ref{tab:augmentation-rates} shows the broader pattern behind the
balanced 12-kernel subset: \gptnew{} and \codex{} stay above 85\% on their
L0-conditional subsets across L1--L4, while \qwenshort{} declines from L1 to
L4. These rates should still be read as surface-form robustness on the
qualifying subset rather than as a definitive memorization test.

\paragraph{Direction Asymmetry And Test Summary.}
Table~\ref{tab:stats-summary} consolidates the paired direction-asymmetry tests
and the descriptive augmentation summary used in the main-text interpretation.

\begin{table*}[t]
\centering
\caption{Statistical summary (\qwenshort{} canonical evaluation). All
confidence intervals are Wilson 95\% intervals. McNemar tests in the direction
asymmetry family use Bonferroni correction
($\alpha = 0.05/5 = 0.01$). Augmentation results are summarized descriptively
because the L0-conditional design changes the denominator across phases.}
\label{tab:stats-summary}
\resizebox{\textwidth}{!}{%
\begin{tabular}{@{}lllll@{}}
\toprule
Test & Statistic & $p$-value & Effect Size & Interpretation \\
\midrule
Qwen augmentation (descriptive) & 74.0\% $\to$ 56.0\% from L1 to L4 & -- & -- & Decline on the L0-conditional subset; not treated as confirmatory because selection depends on L0 success \\  
Direction: CUDA$\leftrightarrow$OMP (McNemar) & $n_{\text{paired}}=24$, exact & $p = 0.180$ & $|h| = 0.44$ & No significant asymmetry ($\alpha=0.01$) \\ 
Direction: CUDA$\leftrightarrow$OpenCL (McNemar) & $n_{\text{paired}}=19$, exact & $p = 1.000$ & $|h| = 0.46$ & No significant asymmetry ($\alpha=0.01$) \\ 
Direction: OMP$\leftrightarrow$OpenCL (McNemar) & $n_{\text{paired}}=17$, exact & $p = 0.289$ & $|h| = 0.53$ & No significant asymmetry ($\alpha=0.01$) \\ 
Direction: CUDA$\leftrightarrow$OMP-target (McNemar) & $n_{\text{paired}}=8$, exact & $p = 0.031$ & $|h| = 2.09$ & Large effect but not significant ($\alpha=0.01$; $n$ too small) \\ 
Direction: OMP$\leftrightarrow$OMP-target (McNemar) & $n_{\text{paired}}=3$, exact & $p = 1.000$ & $|h| = 1.23$ & Underpowered ($n=3$) \\ 
\bottomrule
\end{tabular}
}
\end{table*}


\clearpage 
\section{Translation Prompt Template}
\label{sec:appendix-g}

This appendix reproduces the translation prompt template used by the evaluation pipeline (\texttt{build\_translation\_prompt()} in \texttt{llm\_evaluate.py}). All prompts follow this template; the specific content--source code, build command, target API, and file lists--varies between tasks.

\subsection{System Message}

The system message follows the same template for all translation tasks, parameterized by the source and target API names:

\begin{lstlisting}[
  language={},
  caption={System message template. \texttt{\{src\_api\}} and \texttt{\{tgt\_api\}} are replaced with API display names (e.g., ``CUDA'', ``OpenMP''). \texttt{\{tgt\_lang\}} is the target language hint for the code fence (e.g., ``cuda'', ``cpp'', ``c'').},
  label={lst:system-msg}
]
You are a parallel programming expert
specializing in {src_api} to {tgt_api}
translation. Translate the provided source
code to {tgt_api}. Output ONLY the translated
code, no explanations. For each file, output
a markdown code fence with the filename on
the opening line:

```{tgt_lang} filename={filename}
<code>
```

Preserve the algorithm, I/O behavior, data
formats, and output format exactly. The
translated code must compile with the
provided build command.
\end{lstlisting}

\subsection{User Message Structure}

The user message is assembled from the following sections, in order:

\begin{enumerate}
  \item \textbf{Translation Task} -- source and target API display names. The kernel name and benchmark description are \emph{not} included (anonymization).
  \item \textbf{Target Files to Produce} -- list of genericized target filenames (e.g., \texttt{translated\_0.cpp}, \texttt{translated\_1.cl}). When target infrastructure context is provided, an explanatory note clarifies that only these files replace existing project files.
  \item \textbf{Build Command} -- the anonymized compilation command in a code fence, so the LLM can ensure API and flag compatibility.
  \item \textbf{Build Environment} -- system dependencies from the target spec (e.g., ``CUDA Toolkit $\geq$ 11.0'', ``GCC $\geq$ 9.0'').
  \item \textbf{Source Code} -- each source file presented as a numbered subsection (\texttt{Source File 1}, \texttt{Source File 2}, \ldots) with all C/C++ comments stripped. The section heading includes the source API name (e.g., ``Source Code (CUDA)'').
  \item \textbf{Support / Header Files} (optional) -- source-directory headers and code files, genericized as \texttt{Header File~$N$} and \texttt{Code File~$N$}, with an instruction to inline definitions rather than emit unresolvable \texttt{\#include} directives. Code files are included up to a cumulative 50{,}000-character limit.
  \item \textbf{Target Infrastructure Context} (optional) -- non-kernel target files (prompt payload entries not in \texttt{translation\_targets}, plus target support headers) provided as read-only reference so the LLM can match expected function signatures and data structures. Headed ``DO NOT MODIFY~-- for reference only'' and genericized as \texttt{Infrastructure File~$N$}. Omitted when no non-kernel target files or target support headers remain after filtering.
\end{enumerate}

\noindent Listing~\ref{lst:user-msg} shows the complete user message template with placeholders matching the system message convention
(Listing~\ref{lst:system-msg}).

\begin{lstlisting}[
  language={},
  caption={User message template. Sections marked \emph{(conditional)} appear only when the translation task includes the relevant files. \texttt{\{src\_api\}} and \texttt{\{tgt\_api\}} are API display names; \texttt{\{src\_lang\}} and \texttt{\{tgt\_lang\}} are language hints for code fences. All source code has C/C++ comments stripped before inclusion.},
  label={lst:user-msg}
]
## Translation Task
Source API: {src_api} -> Target API: {tgt_api}

## Target Files to Produce
- translated_0.{ext}
- translated_1.{ext}
  [... one entry per translation target]

_These files will replace the corresponding
files in the target project directory. All other
project files (Makefile, headers, utilities)
remain unchanged._

## Build Command (your code must work with this)
{anonymized_build_command}

## Build Environment
- {dependency_1}
- {dependency_2}

## Source Code ({src_api})
### Source File 1
```{src_lang}
{source_code_comments_stripped}
Source File 2

{source_code_comments_stripped}

Support / Header Files  (conditional)

These files exist in the source build directory
but may NOT exist in the target directory. If
your translated code needs definitions from them,
inline the definitions directly rather than
using #include.

Header File 1

{header_contents_comments_stripped}

Target Infrastructure Context  (conditional)

(DO NOT MODIFY -- for reference only)
These files exist in the target build directory
and will NOT be modified. Your translated code
must be compatible with them.

Infrastructure File 1

{infrastructure_code_comments_stripped}
\end{lstlisting}

\subsection{Anonymization Details}

Six anonymization measures are applied during prompt construction to reduce the risk that benchmark-specific identifiers trigger memorized translations. These measures complement the AST-driven source augmentation (Section~\ref{sec:augmentation-engine}): anonymization targets prompt \emph{metadata}, while augmentation modifies source code \emph{structure}.

\begin{enumerate}
  \item \textbf{Kernel identity omission.} The kernel name and benchmark description are excluded from the prompt entirely.
  \item \textbf{Comment stripping.} All C/C++ line (\texttt{//}) and block (\texttt{/* */}) comments are removed from every file shown to the LLM--source files, support files, and target infrastructure files alike. The stripper uses a state-machine parser that preserves string, character, and raw string literals.
  \item \textbf{Source and support filename genericization.} Source files are labeled \texttt{Source File~1}, \texttt{Source File~2}, etc. Support files are labeled \texttt{Header File~$N$} or \texttt{Code File~$N$} by extension. Target infrastructure files are labeled \texttt{Infrastructure File~$N$}.
  \item \textbf{Target filename genericization.} Target output filenames are replaced with \texttt{translated\_0.ext}, \texttt{translated\_1.ext}, etc., preserving original file extensions. An internal mapping restores original filenames when writing LLM output to disk, including any cross-file \texttt{\#include} references the LLM emits using the generic names.
  \item \textbf{Build command anonymization.} Kernel-specific identifiers are removed from the build command: \texttt{make~\textit{target}} is reduced to \texttt{make} (relying on the Makefile default target), and kernel names in other command strings are replaced with a generic placeholder.
  \item \textbf{API-name retention.} The source and target API names (e.g., ``CUDA'', ``OpenMP'') are intentionally \emph{not} anonymized, as they are essential to the translation task specification.
\end{enumerate}

\section{Evaluation Cost and Reproducibility}
\label{sec:appendix-j}

This appendix groups the practical information needed to reproduce the reported
evaluation campaigns: model/provider settings, the shared hardware/software
platform, and the aggregate token and wall-clock cost of the three campaigns.
Table~\ref{tab:model-config} records the provider-side model settings.
Table~\ref{tab:hardware} records the shared execution environment.
Table~\ref{tab:eval-cost} then reports aggregate campaign cost.

\begin{table}[htbp]
\centering
\caption{Model configurations. $^\dagger$Azure reasoning models do not accept
explicit temperature or top-$p$ parameters; sampling is controlled internally by
the provider.}
\label{tab:model-config}
\footnotesize
\begin{tabular}{@{}lllllc@{}}
\toprule
Model & Provider & Arch. & Parameters & Reasoning & Temp. \\
\midrule
\qwen{} & Together AI & MoE & 397B / 17B active & \texttt{enable\_thinking} & 0.7 \\
\gptnew{} & \gptprovider{} & proprietary & proprietary & \texttt{reasoning\_effort=medium} & N/A$^\dagger$ \\
\codex{} & \gptprovider{} & proprietary & proprietary & \texttt{reasoning\_effort=medium} & N/A$^\dagger$ \\
\bottomrule
\end{tabular}
\end{table}

\begin{table}[htbp]
\centering
\caption{Hardware and software configuration shared across all three model
campaigns.}
\label{tab:hardware}
\small
\begin{tabular}{@{}ll@{}}
\toprule
Component & Specification \\
\midrule
GPU & NVIDIA GeForce RTX 4070 (12\,GB, Ada Lovelace, sm\_89) \\
CPU & AMD Ryzen 9 7900X (12 cores / 24 threads, 4.7\,GHz) \\
RAM & 32\,GB DDR5 \\
OS & Ubuntu 24.04 LTS \\
CUDA toolkit & NVIDIA HPC SDK 24.3 (nvcc 12.3, V12.3.103) \\
C/C++ compiler & GCC 12.4.0 \\
OpenMP & GCC \texttt{-fopenmp}; nvc++ 24.3 for target offload \\
OpenCL runtime & NVIDIA via HPC SDK 24.3 \\
Python & 3.12.3 \\
\bottomrule
\end{tabular}
\end{table}

\begin{table}[h]
    \centering
    \caption{Evaluation campaign cost summary. All metrics are aggregated from per-result JSON metadata. Token counts reflect only the tokens recorded per evaluation call (see text).}
    \label{tab:eval-cost}
    \begin{tabular}{lrrr}
        \toprule
        \textbf{Metric} & \textbf{\qwenshort{}} & \textbf{\gptnew{}} & \textbf{\codex{}} \\
        \midrule
        Tasks evaluated & 708 & 822 & 814 \\
        Total tokens & 13.6M & 14.9M & 13.2M \\
        \quad Input & 8.9M & 9.3M & 9.2M \\
        \quad Output & 4.7M & 5.6M & 4.0M \\
        Avg.\ tokens per task & 19{,}149 & 18{,}138 & 16{,}204 \\
        Total LLM wall time (h) & 14.0 & 27.4 & 11.9 \\
        Campaign dates (2026) & Apr 21--24 & Apr 25--27 & Apr 30--May 1 \\
        \midrule
        \textbf{All models} & \multicolumn{3}{c}{2{,}344 tasks, 41.7M tokens, 53.4\,h total} \\
        \bottomrule
    \end{tabular}
\end{table}

The higher wall time for \gptnew{} reflects its use of internal
reasoning (\texttt{reasoning\_effort=medium}), which approximately doubles
per-call latency relative to the non-reasoning models at comparable output
lengths.

Per-result token counts reflect the usage statistics returned by each
provider's API, which include system-prompt tokens.
They may modestly undercount actual billed consumption due to
provider-internal formatting tokens or SDK-level retries on transient
failures. Researchers budgeting a replication should consult current
provider pricing and allow a margin for such overhead.
The \qwenshort{} task count includes 82~records involving \knownfail{}
specs that are excluded from pass-rate denominators
(Section~\ref{sec:appendix-a-extended}); the \gptnew{} and \codex{}
batches pre-excluded such specs, so all on-disk records are
evaluation-eligible.


\section{Evaluation Card}
\label{sec:appendix-k}

Following Model Cards~\cite{ModelCards2019} and Datasheets~\cite{Datasheets2021},
we provide an \emph{Evaluation Card} for \parbench{} that states what it
measures, what it does not measure, the claims it supports, the assumptions it
relies on, and how results should be reported.
Table~\ref{tab:eval-card-summary} summarizes the scope; the subsections below
justify each point.

\begin{table*}[htbp]
\centering
\caption{Evaluation Card summary. Each row states a key property of \parbench{}'s
measurement scope. Detailed discussion appears in the referenced subsection.}
\label{tab:eval-card-summary}
\small
\begin{tabularx}{\textwidth}{@{}p{2.8cm}Xr@{}}
\toprule
Dimension & Key Points & Ref. \\
\midrule
\multicolumn{3}{l}{\textit{What \textnormal{\parbench{}} measures}} \\
Declared-oracle pass/fail & Binary pass/fail via conjunctive build$\to$run$\to$verify pipeline & \S\ref{sec:ec-measures} \\
Failure taxonomy & Four failure classifications (\extractionfail{}, \buildfail{}, \runfail{}, \verifyfail{}) plus \pass{} & \S\ref{sec:ec-measures} \\
Direction asymmetry & Pass-rate variation across 10 translation directions (3 APIs + OMP target) & \S\ref{sec:ec-measures} \\
Surface-form robustness & Stability under six AST-level source transforms at four levels & \S\ref{sec:ec-measures} \\
\midrule
\multicolumn{3}{l}{\textit{What \textnormal{\parbench{}} does not measure}} \\
Performance & No speedup, throughput, occupancy, memory, or energy measurement & \S\ref{sec:ec-not-measures} \\
Numerical accuracy & 80 of 87 non-\knownfail{} specs use weak oracles (exit code + stdout pattern) & \S\ref{sec:ec-not-measures} \\
Code quality & No readability, maintainability, or style assessment of translations & \S\ref{sec:ec-not-measures} \\
Repo-level translation & Kernel-centric by design; host code and build infra remain fixed & \S\ref{sec:ec-not-measures} \\
Self-repair capability & Single-attempt evaluation; iterative feedback not exercised & \S\ref{sec:ec-not-measures} \\
\midrule
\multicolumn{3}{l}{\textit{Validity and assumptions}} \\
Valid claims & Pass@$k$, failure taxonomy, direction asymmetry, surface-form robustness on the L0-conditional subset & \S\ref{sec:ec-valid} \\
Invalid claims & ``Efficient code,'' ``production-ready,'' ``understands parallelism,'' ``not memorizing'' & \S\ref{sec:ec-invalid} \\
Key assumptions & Baseline-valid source specs, oracle sufficiency (weakest), single-platform generalization & \S\ref{sec:ec-assumptions} \\
\midrule
\multicolumn{3}{l}{\textit{Guidance}} \\
User guidance & Adding models/kernels, interpreting results, comparing models, common pitfalls & \S\ref{sec:ec-guidance} \\
Reporting protocol & Required and recommended items for publications using \parbench{} & \S\ref{sec:ec-reporting} \\
\bottomrule
\end{tabularx}
\end{table*}

\subsection{What \parbench{} Measures}
\label{sec:ec-measures}

\parbench{} measures the following properties of LLM-based parallel code
translation:
\begin{enumerate}
\item \textbf{Binary declared-oracle pass/fail of kernel-level translation.} 
    The harness evaluates whether LLM-translated code compiles, executes, and 
    produces output matching the spec's verification strategies. Verification 
    applies a conjunction of strategies: all declared checks (exit code, stdout 
    pattern, and optionally numeric comparison or file hash) must pass for a 
    record to receive \pass{} status. The result is \pass{} or one of four 
    failure statuses (\extractionfail{}, \buildfail{}, \runfail{}, 
    \verifyfail{}); timeouts are classified under the relevant pipeline stage 
    (\buildfail{} or \runfail{}).

  \item \textbf{Translation capability across three primary APIs and ten 
    directions.} CUDA, OpenMP, and OpenCL form six bidirectional standard 
    directions; OpenMP Target adds four case-study directions. The evaluation 
    matrix covers 142~unique source--target pairs 
    (Section~\ref{sec:eval-pipeline}).

  \item \textbf{Failure mode taxonomy.} Four failure classifications 
    (\extractionfail{}, \buildfail{}, \runfail{}, \verifyfail{}) enable 
    diagnostic analysis of \emph{where} in the pipeline translations fail. 
    Build failures indicate incomplete API-surface adaptation; run failures 
    indicate runtime errors or timeouts; verify failures indicate incorrect 
    output from otherwise executable code. This granularity distinguishes 
    \parbench{} from compile-only or text-similarity evaluations.

  \item \textbf{Surface-form robustness on the L0-conditional subset.} Six 
    AST-level transforms (SwapCondition, ArithmeticTransform, ChangeNames, 
    TypedefExpansion, PointerArithmeticToArrayIndex, ChangeFunctionNames) at 
    four intensity levels (L1--L4) test whether pass/fail status is stable 
    under AST-level code changes intended to preserve behavior 
    (Section~\ref{sec:augmentation-engine}). The augmentation campaign applies
    only to direction--kernel pairs that qualify under the L0-conditional
    filter.

  \item \textbf{Direction asymmetry.} Pass-rate variation across translation 
    directions quantifies structural difficulty differences between API pairs. 
    CUDA-to-OpenMP consistently passes at higher rates than OpenCL-to-CUDA 
    across all three evaluated models (Section~\ref{sec:direction-analysis}).
    \item \textbf{Cross-model discrimination.} The benchmark distinguishes models that differ in translation capability, subject to sampling-condition caveats (Section~\ref{sec:ec-assumptions}). In the current evaluation, task-level pairwise analysis on 142~shared L0 tasks yields Cohen's $h \approx 0.71$ between \qwenshort{} and both GPT models (task-level pass-any rates: \qwenshort{} 50/142, \gptnew{} 99/142, \codex{} 97/142), while \gptnew{} and \codex{} are statistically indistinguishable (Fisher's $p = 1.0$, McNemar $p = 0.72$).
    
    \item \textbf{Per-kernel difficulty heterogeneity.} For \qwenshort{}, canonical pass rates range from 72.2\% (floydwarshall, iso2dfd) to 0\% (10~kernels), enabling fine-grained analysis of which computational patterns--stencils, graph traversal, ODE integration, pointer-heavy data structures--are harder to translate (Appendix Table~\ref{tab:per-kernel-full} reports rates including augmentation levels).

\end{enumerate}

\subsection{What \parbench{} Does Not Measure}
\label{sec:ec-not-measures}

The following properties are explicitly outside \parbench{}'s measurement scope:

\begin{enumerate}
  \item \textbf{Performance and resource efficiency.} No speedup, throughput,
    occupancy, bandwidth, memory, or energy measurement is performed.
    Wall-clock time is captured but is unreliable for sub-millisecond baselines.
    No kernel-level profiling (\texttt{ncu}/\texttt{nsys} for CUDA,
    \texttt{omp\_get\_wtime} for OpenMP) is included. A harness-passing
    translation that runs 100$\times$ slower than the original receives
    \pass{}. TRACE~\cite{TRACE2026} addresses this gap.

  \item \textbf{Code quality, readability, or maintainability.} Translated
    code is stored in result files but no AST analysis, cyclomatic complexity,
    or style metrics are computed. A harness-passing but unreadable translation receives
    \pass{}.

  \item \textbf{Partial pass credit or oracle-strength gradations.} Verification
    is binary: a numerical result that deviates from the reference by 0.02\%
    beyond the configured tolerance receives \verifyfail{}, identical to a
    completely wrong output. There is no partial-credit classification.

  \item \textbf{Numerical accuracy for most specs.} Only 7 of 87 non-\knownfail{} specs (8\%) have medium or strong oracles (\texttt{numeric\_comparison} or \texttt{file\_hash}). The remaining 80 specs (92\%) rely on \texttt{exit\_code} and \texttt{stdout\_pattern}, which verify that the program runs and prints expected banners but do not independently validate numerical results. A translation that computes wrong numbers but prints the expected completion message would receive \pass{} on those specs. The oracle strength distribution across all 206~specs is: 2 strong (\texttt{file\_hash}), 5 medium (\texttt{numeric\_comparison}), 46 weak (\texttt{stdout\_pattern} + \texttt{exit\_code}), and 153 untagged (effectively weak). Upgrading weak oracles to numeric comparison requires per-kernel reference-output engineering; we prioritize this in future work.
  
  \item \textbf{Iterative self-repair or agentic translation.} Each sample is a single LLM call with no feedback loop. The evaluation pipeline implements iterative repair (multi-turn error feedback with linker analysis via \texttt{-{}-max-retries}), but all reported results use \texttt{max\_retries=1} (zero-shot). Evaluating multi-turn repair strategies--where build or verification failures are fed back to the model for corrected attempts--and agentic translation pipelines that combine planning, tool use, and iterative refinement remain future work.
  
  \item \textbf{Cross-platform portability.} All evaluations run on a single
    platform: NVIDIA RTX~4070 (sm\_89), AMD Ryzen~9~7900X, Ubuntu~24.04, HPC
    SDK~24.3 (Appendix Table~\ref{tab:hardware}). No testing on other GPU
    architectures (A100, H100), CPU-only environments, or other OS
    configurations.

  \item \textbf{Repository-level translation.} By design, \parbench{} isolates
    kernel translation from build-system reconstruction. Host code, Makefiles,
    and I/O routines remain fixed. The benchmark does not measure the LLM's
    ability to restructure project files, generate build systems, or coordinate
    multi-component applications (cf.\
    ParEval-Repo~\cite{ParEvalRepo2025}).

  \item \textbf{Training-data memorization beyond surface form (definitively).} Augmentation tests
    surface-form robustness, not algorithmic memorization
    (Section~\ref{sec:ec-invalid}).
\end{enumerate}

\subsection{Valid Claims}
\label{sec:ec-valid}

The following conclusions can be drawn from \parbench{} results when
accompanied by the stated conditions:

\begin{enumerate}
  \item \textbf{``Model~X achieves Y\% pass@$k$ on kernel-centric parallel
    code translation across $Z$ directions.''} Valid when: the exact pass@$k$
    metric is specified (pass@1 vs.\ pass@3), augmentation levels are stated
    (L0 only vs.\ L0--L4), directions are enumerated, and confidence intervals
    are included.

  \item \textbf{``Build-stage failure is the dominant failure mode for
    Model~X.''} Valid when: the failure taxonomy breakdown is reported with
    counts and percentages from the full record set.

  \item \textbf{``Direction A$\to$B is harder/easier than Direction C$\to$D
    for Model~X.''} Valid when: backed by per-direction pass rates with Wilson
    confidence intervals. Claims of statistical significance should use
    McNemar's test on paired kernels with appropriate multiple-comparison
    correction.

  \item \textbf{``Model~X maintains declared-oracle pass rates across
    L1--L4 on the L0-conditional subset.''} Valid when: the L0-conditional filter is disclosed, the
    qualifying subset size is stated, the direction restriction is noted, and
    the analysis is presented as descriptive rather than as a definitive
    memorization test.

  \item \textbf{``Model~X discriminates from Model~Y on \parbench{}.''} Valid
    when: (a)~all compared models are evaluated on the same task set, (b)~McNemar's
    paired test is reported with effect size and concordance table,
    (c)~sampling-configuration differences are explicitly disclosed if
    temperatures or reasoning modes differ, and (d)~the claim is scoped to
    ``\parbench{} tasks,'' not ``parallel translation in general.''

  \item \textbf{``Kernel~K is harder to translate than Kernel~J.''} Valid
    when: both kernels have sufficient sample sizes ($n \geq 18$ recommended)
    and confidence intervals are reported.
\end{enumerate}

\subsection{Invalid Claims}
\label{sec:ec-invalid}

The following conclusions \textbf{cannot} be drawn from \parbench{} results:

\begin{enumerate}
  \item \textbf{``Model~X produces efficient/fast parallel code.''} Performance
    is not measured. A \pass{} means the code compiles, runs, and satisfies the
    declared oracle--not that it runs at acceptable speed.
    
    \item \textbf{``Model~X's translations are production-ready.''} No code review, security audit, or race-condition detection is performed. Parallel code can have latent data races that produce correct output on some executions.
    
    \item \textbf{``Model~X understands parallel programming.''} \parbench{}
    measures behavioral outcomes, not internal representations. A model could
    pass by pattern matching without ``understanding'' synchronization
    semantics.

  \item \textbf{``\parbench{} pass rate generalizes to all HPC translation
    tasks.''} The corpus contains 87~non-\knownfail{} specs from five
    suites, with Rodinia contributing 53~(61\%). Domains such as distributed-memory MPI, GPU tensor operations, and FPGA HLS are outside its scope.

  \item \textbf{``Model~X is definitively better than Model~Y at parallel translation.''} Only valid under matched sampling conditions. If temperatures, reasoning modes, or provider-side controls differ (as they do between \qwenshort{} and the Azure models in this paper), the observed gap reflects an unknown mixture of model capability and sampling configuration (Section~\ref{sec:sampling-config}).

  \item \textbf{``Surface-form robustness proves the model is not
    memorizing.''} Surface-form augmentation (variable renaming, syntax sugar)
    tests robustness to cosmetic code changes. Algorithmic memorization--where
    the model recognizes the computation and produces a previously seen
    translation template--is not tested by these transforms.

  \item \textbf{``A \pass{} means the translation is numerically faithful.''}
    For the 80 of 87~specs with weak oracles, verification checks that the
    program ran and printed expected output banners. Numerical results are not
    independently validated. Only 7~specs have medium or strong oracles that
    verify numerical output.

    \item \textbf{``OpenCL$\to$CUDA is impossible for LLMs.''} \qwenshort{} achieves 0\% but both \gptnew{} and \codex{} achieve 19.3\% on the same direction. Zero-rate results for a single model reflect that model's capability, not an inherent impossibility.
\end{enumerate}

\subsection{Assumptions}
\label{sec:ec-assumptions}

\parbench{}'s evaluation results are meaningful under the following assumptions.
We note mitigations and residual risks for each.

\begin{enumerate}
  \item \textbf{Source implementations are baseline-valid on the reference platform.} \parbench{}
    treats the original benchmark implementations as the reference baseline. If a source
    implementation has a latent bug, faithful translations of that bug would
    pass the declared oracle. \textit{Mitigation:} all 87~non-\knownfail{} specs pass
    the baseline build/run/verify pipeline on the reference platform.

  \item \textbf{Declared oracles are sufficient to detect material
    translations.} This is the weakest assumption. For the 80~specs with weak
    oracles, a translation that produces wrong numerical results but expected
    program flow (runs to completion, prints expected banners) would receive a
    false \pass{}. \textit{Mitigation:} 7~specs have
    \texttt{numeric\_comparison} or \texttt{file\_hash} oracles; the dominant
    failure mode (\buildfail{}) is caught by all oracle types regardless of
    oracle strength. Upgrading weak oracles is identified as future work
    (Section~\ref{sec:discussion}).

  \item \textbf{The single evaluation platform generalizes to other NVIDIA GPU
    configurations.} All results are from one machine (RTX~4070, sm\_89).
    Different GPU architectures may have different compiler behavior, runtime
    characteristics, or memory limits.
    
    \item \textbf{Three samples per task capture meaningful stochastic variance under each provider's sampling regime.} For \qwenshort{} this uses temperature~0.7; for \gptnew{} and \codex{} sampling is provider-controlled (Section~\ref{sec:sampling-config}). The pass@1-to-pass@3 gap (11.3~pp for \qwenshort{}, 7.0~pp for \gptnew{}, 5.6~pp for \codex{}) suggests moderate within-task variance. More samples would provide tighter estimates at proportional cost.
      
      \item \textbf{Kernel-centric translation isolates parallel programming
    capability.} By fixing host code and build infrastructure, \parbench{}
    prevents build-system reconstruction from dominating results (cf.\
    ParEval-Repo's 0\% at $>$133~SLoC~\cite{ParEvalRepo2025}). However, this
    also means that a model's ability to handle multi-component integration--a
    real-world requirement--is explicitly not tested.

  \item \textbf{The 9 \knownfail{} exclusions are legitimately infrastructure
    failures.} Each exclusion is documented with a specific technical cause (CUDA~12 API deprecation, missing libraries, pre-existing build or runtime errors).

  \item \textbf{Benchmark suites are representative of real HPC workloads.}
    The five suites are well-established in HPC research (Rodinia: IISWC
    2009~\cite{Rodinia2009}, HeCBench: 2023~\cite{HeCBench2023}, XSBench:
    ANL~\cite{XSBench2014}). However, they over-represent structured stencil
    and reduction patterns and under-represent irregular computations,
    multi-physics coupling, and modern GPU programming idioms (tensor cores,
    cooperative groups).

    \item \textbf{Augmentation transforms are intended to preserve source behavior.} Each transform is backed by \texttt{libclang} AST analysis and validated by 15~unit tests. 80~of 87~specs pass all L1--L4 augmented baseline verification. The 7~failures are \texttt{omp\_target}-specific: condition operand swapping (e.g., \texttt{a < b} $\to$ \texttt{b > a}) causes the GPU-offloading compiler to generate device code that crashes or produces divergent output, even though the transform is behavior-preserving under standard~C compilation. Variable renaming is disabled for files containing OpenMP pragmas; the 5~HeCBench failures involve only condition swapping, while the 2~remaining failures (RSBench, XSBench at L4) co-occur with additional transforms on non-pragma helper files. These 7~augmented variants fail their own baseline verification, meaning the behavior-preserving assumption cannot be confirmed at those levels; they are flagged in analysis but still participate as source text in LLM evaluation (since the source is used as prompt input, not compiled).
    
    \item \textbf{Prompt anonymization does not change task difficulty.}
    Stripping comments, genericizing filenames, and removing kernel names reduce memorization cues but might also remove helpful context, potentially making \parbench{} results conservative relative to non-anonymized use.
\end{enumerate}

\subsection{User Guidance}
\label{sec:ec-guidance}

\textbf{Adding a new model.} Register the model in the evaluation pipeline's
model registry, specifying provider, API model identifier, and
thinking/temperature configuration. Run the canonical campaign with all suites;
the evaluation pipeline handles prompt construction, LLM calls, and file
extraction, and the harness handles build/run/verify.

\textbf{Adding new kernels.} Write a spec JSON following the schema documented
in Appendix~\ref{sec:appendix-a-schema}. The spec must define identity
(unique ID, API, suite), file partitioning (prompt payload, translation targets,
support files), build commands, run configuration, and verification strategies.
Validate the baseline by running the harness verify command. Append to the
manifest (append-only).

\textbf{Interpreting pass rates.} Always report the denominator (number of
evaluation-eligible records after \knownfail{} exclusion), the augmentation level (L0 vs.\
L0--L4), and Wilson 95\% confidence intervals. Per-direction breakdowns are
essential--aggregate rates conceal strong direction asymmetry.

\textbf{Comparing models.} Use paired tests on task outcomes (same tasks, same direction--kernel pairs): Fisher's exact test for pairwise comparisons, McNemar's test for concordance analysis. Report concordance tables (both-pass, both-fail, discordant). Always disclose temperature, reasoning mode, and any provider-side sampling controls. If sampling conditions are unmatched, state this explicitly and scope the comparison accordingly.

\textbf{Augmentation analysis.} Use the L0-conditional filter: include a pair in the augmentation campaign only if $\geq$1 of its L0 samples passes. Report the filter rate (fraction of pairs qualifying). Results apply only to the qualifying subset. Single samples per augmentation level (L1--L4) trade statistical precision for budget efficiency.

\textbf{Common pitfalls.}
\begin{itemize}
  \item Do not compare aggregate pass rates across models without disclosing
    sampling conditions.
  \item Do not claim ``the model understands parallel programming'' from pass
    rates alone.
  \item Do not interpret weak-oracle \pass{} as evidence of numerical
    fidelity.
  \item Do not modify the \knownfail{} exclusion policy without documenting
    the change and its effect on denominators.
  \item Do not change spec run arguments without reading the source code's
    \texttt{argc} check--a documented cause of past silent failures.
\end{itemize}

\subsection{Recommended Reporting Protocol}
\label{sec:ec-reporting}

When publishing results obtained with \parbench{}, we recommend reporting at
minimum the items in Table~\ref{tab:reporting-required}. For cross-model
comparisons and augmentation analyses, the additional items in
Table~\ref{tab:reporting-recommended} strengthen interpretability.

\begin{table}[htbp]
\centering
\caption{Required reporting items for \parbench{} results.}
\label{tab:reporting-required}
\small
\begin{tabularx}{\columnwidth}{@{}p{2.3cm}X@{}}
\toprule
Item & What to report \\
\midrule
Model identity & Provider, model ID, version or date \\
Sampling config & Temperature/top-$p$ if caller-controlled, otherwise ``provider-controlled''; reasoning mode; $n$ (samples) \\
pass@$k$ & pass@1 and pass@3 with confidence intervals \\
Failure taxonomy & Counts and percentages for each failure type \\
Direction breakdown & Per-direction pass rates with Wilson CIs \\
Denominator & Total evaluation-eligible records after \knownfail{} exclusion \\
Augmentation scope & Levels evaluated, filter criterion, subset size \\
Platform & GPU, CPU, OS, compiler versions \\
\bottomrule
\end{tabularx}
\end{table}

\begin{table}[htbp]
\centering
\caption{Recommended additional reporting items.}
\label{tab:reporting-recommended}
\small
\begin{tabularx}{\columnwidth}{@{}p{2.3cm}X@{}}
\toprule
Item & What to report \\
\midrule
\multicolumn{2}{l}{\textit{Cross-model comparisons}} \\
Paired test & McNemar on paired tasks with concordance table \\
Effect size & Cohen's $h$ for overall and per-direction gaps \\
Sampling caveat & Explicit statement if conditions are unmatched \\
Task variance & Fraction of hard-fail (0/$n$), noisy, always-pass ($n$/$n$) \\
\midrule
\multicolumn{2}{l}{\textit{Augmentation analysis}} \\
Per-level rates & Descriptive rates at L0--L4 on balanced subset \\
Filter disclosure & ``$X$ of 142 pairs qualify ($Y$\%)'' \\
Direction scope & Which direction(s) the analysis covers \\
Power analysis & Detectable effect size at 80\% power \\
\bottomrule
\end{tabularx}
\end{table}


\section{Artifact Availability}
\label{sec:artifact-availability}

The \parbench{} artifact is publicly available at \url{https://github.com/Scientific-Computing-Lab/ParBench}. It contains the curated \parbench{} specifications, the build-run-verify harness, the evaluation and augmentation pipelines, and per-record result JSONs for all three evaluated models. The artifact contains 206 generated JSON specifications, including the 96 curated specifications used in the paper and the 87 eval-eligible specifications that pass the baseline build-run-verify pipeline. Each kernel specification embeds its provenance: the upstream suite, repository URL, pinned commit hash, license, build recipe, run configuration, and verification oracle (Section~\ref{sec:suite-selection}). The \parbench{} framework is released under the MIT license, and a versioned archival snapshot will be deposited on Zenodo to provide a persistent DOI. Upstream suites retain their original licenses: Rodinia (BSD-like), XSBench (MIT), RSBench (MIT), HeCBench (BSD-3-Clause), and mixbench (GPL-2.0). Structured metadata is provided through the JSON specification schema and per-spec provenance fields. Because \parbench{} is an executable evaluation benchmark rather than a model-training dataset, Croissant metadata is not applicable here.


\section{Additional Tables and Figures}
\label{sec:appendix-f}

This appendix retains the dense reference figures and full per-kernel tables
that are easier to consult separately from the narrative appendices above.
Per-kernel translation heatmaps appear in Figure~\ref{fig:kernel-heatmap-unified}
(main body). The figures below focus on direction-level pass@$k$ behavior,
cross-suite comparisons, and full per-kernel rate tables.



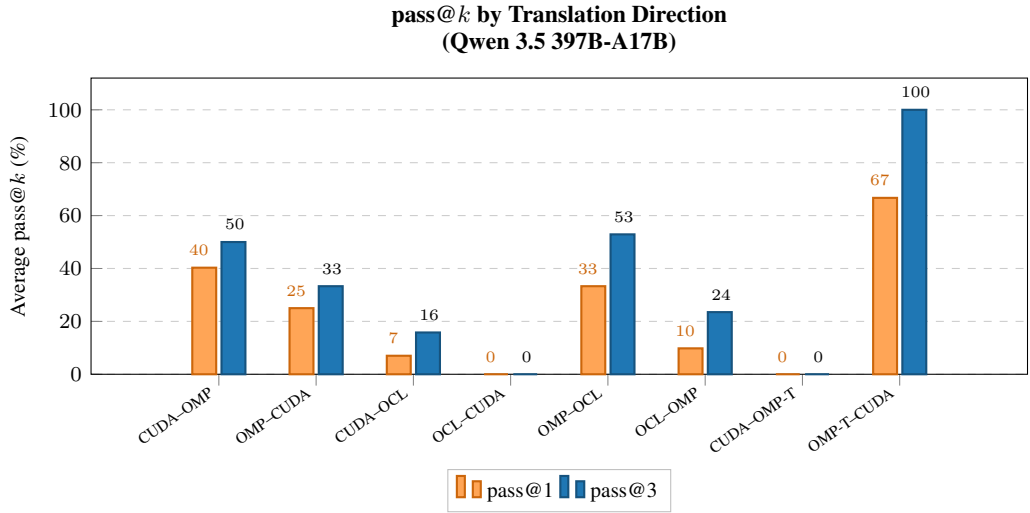
\begin{figure}[!htbp]
\centering

\begin{tikzpicture}
\begin{axis}[
  parbench compact,
  ybar,
  bar width=9pt,
  ymin=0, ymax=112,
  xmin=-0.65, xmax=7.65,
  xtick={0,1,2,3,4,5,6,7},
  xticklabels={
    {CUDA--OMP},
    {OMP--CUDA},
    {CUDA--OCL},
    {OCL--CUDA},
    {OMP--OCL},
    {OCL--OMP},
    {CUDA--OMP-T},
    {OMP-T--CUDA}
  },
  xticklabel style={font=\tiny, rotate=35, anchor=east},
  ytick={0,20,40,60,80,100},
  ylabel={Average pass@$k$ (\%)},
  grid style={dashed, draw=gray!40},
  title={pass@$k$ by Translation Direction\\(Qwen~3.5 397B-A17B)},
  legend style={
    at={(0.5,-0.32)}, anchor=north,
    legend columns=2,
  },
  nodes near coords={\pgfmathprintnumber[fixed,precision=0]{\pgfplotspointmeta}},
]

\addplot[
  fill=pbOrange!70,
  draw=pbOrange!80!black,
  line width=0.8pt,
  every node near coord/.style={
    font=\tiny\bfseries,
    anchor=south,
    xshift=-7.5pt,
    yshift=1pt,
    color=pbOrange!80!black
  },
] table[x=dir, y=p1] {figures/figures_tek_version/f5_pass_at_k_by_direction_qwen.dat};
\addlegendentry{pass@1}

\addplot[
  fill=pbRodinia,
  draw=pbRodinia!70!black,
  line width=0.8pt,
  every node near coord/.style={
    font=\tiny\bfseries,
    anchor=south,
    xshift=6pt,
    yshift=1pt,
    color=black
  },
] table[x=dir, y=p3] {figures/figures_tek_version/f5_pass_at_k_by_direction_qwen.dat};
\addlegendentry{pass@3}

\end{axis}
\end{tikzpicture}
\caption{Pass@$k$ by translation direction (\qwenshort{}, L0, three samples per task).}
\label{fig:pass-at-k-qwen}
\end{figure}

\begin{figure}[!htbp]
\centering

\begin{tikzpicture}
\begin{axis}[
  parbench compact,
  ybar,
  bar width=9pt,
  ymin=0, ymax=112,
  xmin=-0.65, xmax=7.65,
  xtick={0,1,2,3,4,5,6,7},
  xticklabels={
    {CUDA--OMP},
    {OMP--CUDA},
    {CUDA--OCL},
    {OCL--CUDA},
    {OMP--OCL},
    {OCL--OMP},
    {CUDA--OMP-T},
    {OMP-T--CUDA}
  },
  xticklabel style={font=\tiny, rotate=35, anchor=east},
  ytick={0,20,40,60,80,100},
  ylabel={Average pass@$k$ (\%)},
  grid style={dashed, draw=gray!40},
  title={pass@$k$ by Translation Direction\\(GPT-5.4)},
  legend style={
    at={(0.5,-0.32)}, anchor=north,
    legend columns=2,
  },
  nodes near coords={\pgfmathprintnumber[fixed,precision=0]{\pgfplotspointmeta}},
]

\addplot[
  fill=pbOrange!70,
  draw=pbOrange!80!black,
  line width=0.8pt,
  nodes near coords,
  every node near coord/.style={font=\tiny\bfseries, anchor=south, xshift=-7.5pt, yshift=1pt, color=pbOrange!80!black},
] table[x=dir, y=p1] {figures/figures_tek_version/f5_pass_at_k_by_direction_gpt_5-4.dat};
\addlegendentry{pass@1}

\addplot[
  fill=pbRodinia,
  draw=pbRodinia!70!black,
  line width=0.8pt,
  nodes near coords,
  every node near coord/.style={font=\tiny\bfseries, anchor=south, xshift=6pt, yshift=1pt, color=black},
] table[x=dir, y=p3] {figures/figures_tek_version/f5_pass_at_k_by_direction_gpt_5-4.dat};
\addlegendentry{pass@3}

\end{axis}
\end{tikzpicture}
\caption{Pass@$k$ by translation direction (\gptnew{}, L0, three samples per task).}
\label{fig:pass-at-k-gpt54}
\end{figure}
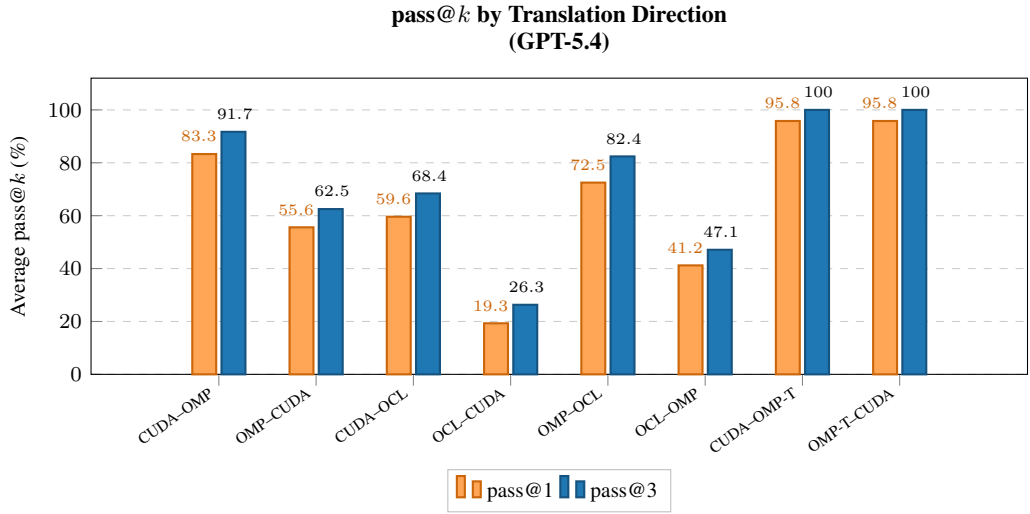

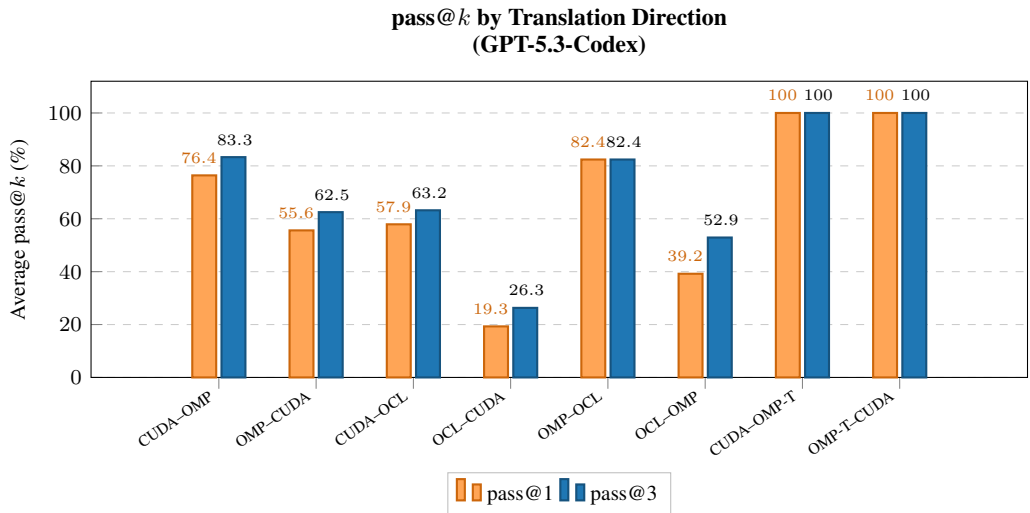
\begin{figure}[!htbp]
\centering

\begin{tikzpicture}
\begin{axis}[
  parbench compact,
  ybar,
  bar width=9pt,
  ymin=0, ymax=112,
  xmin=-0.65, xmax=7.65,
  xtick={0,1,2,3,4,5,6,7},
  xticklabels={
    {CUDA--OMP},
    {OMP--CUDA},
    {CUDA--OCL},
    {OCL--CUDA},
    {OMP--OCL},
    {OCL--OMP},
    {CUDA--OMP-T},
    {OMP-T--CUDA}
  },
  xticklabel style={font=\tiny, rotate=35, anchor=east},
  ytick={0,20,40,60,80,100},
  ylabel={Average pass@$k$ (\%)},
  grid style={dashed, draw=gray!40},
  title={pass@$k$ by Translation Direction\\(GPT-5.3-Codex)},
  legend style={
    at={(0.5,-0.32)}, anchor=north,
    legend columns=2,
  },
  nodes near coords={\pgfmathprintnumber[fixed,precision=0]{\pgfplotspointmeta}},
]

\addplot[
  fill=pbOrange!70,
  draw=pbOrange!80!black,
  line width=0.8pt,
  nodes near coords,
  every node near coord/.style={font=\tiny\bfseries, anchor=south, xshift=-7.5pt, yshift=1pt, color=pbOrange!80!black},
] table[x=dir, y=p1] {figures/figures_tek_version/f5_pass_at_k_by_direction_gpt_5-3_codex.dat};
\addlegendentry{pass@1}

\addplot[
  fill=pbRodinia,
  draw=pbRodinia!70!black,
  line width=0.8pt,
  nodes near coords,
  every node near coord/.style={font=\tiny\bfseries, anchor=south, xshift=6pt, yshift=1pt, color=black},
] table[x=dir, y=p3] {figures/figures_tek_version/f5_pass_at_k_by_direction_gpt_5-3_codex.dat};
\addlegendentry{pass@3}

\end{axis}
\end{tikzpicture}
\caption{Pass@$k$ by translation direction (\codex{}, L0, three samples per task).}
\label{fig:pass-at-k-codex}
\end{figure}

\begin{figure}[!htbp]
\centering
\begin{tikzpicture}
\begin{axis}[
  parbench compact,
  ybar,
  bar width=22pt,
  ymin=0, ymax=1.08,
  xmin=-0.6, xmax=4.6,
  xtick={0,1,2,3,4},
  xticklabels={Rodinia, XSBench, RSBench, mixbench, HeCBench},
  xticklabel style={font=\footnotesize},
  ytick={0,0.2,0.4,0.6,0.8,1.0},
  ylabel={Pass Rate},
  title={L0 Pass Rate by Suite (Qwen~3.5 397B-A17B)},
  grid style={dashed, draw=gray!40},
]
\addplot[
  fill=pbRodinia,
  draw=pbRodinia!70!black,
  line width=0.8pt,
  error bars/.cd,
    y dir=both,
    y explicit,
    error bar style={line width=0.8pt, black},
    error mark options={rotate=90, mark size=3pt, line width=0.8pt, black},
] table[x=x, y=y, y error minus=errm, y error plus=errp] {figures/figures_tek_version/f6_cross_suite_comparison_qwen.dat};
  \node[font=\tiny\bfseries, anchor=south] at (axis cs:0,0.2479) {15/110};
  \node[font=\tiny\bfseries, anchor=south] at (axis cs:1,0.4253) {0/6};
  \node[font=\tiny\bfseries, anchor=south] at (axis cs:2,0.4253) {0/6};
  \node[font=\tiny\bfseries, anchor=south] at (axis cs:3,0.4253) {0/6};
  \node[font=\tiny\bfseries, anchor=south] at (axis cs:4,0.6118) {12/30};
\end{axis}
\end{tikzpicture}
\caption{Cross-suite pass rate comparison (L0, \qwenshort{}). Per-suite aggregate pass rates with Wilson 95\% CIs across all five benchmark suites.}
\label{fig:cross-suite-qwen}
\end{figure}
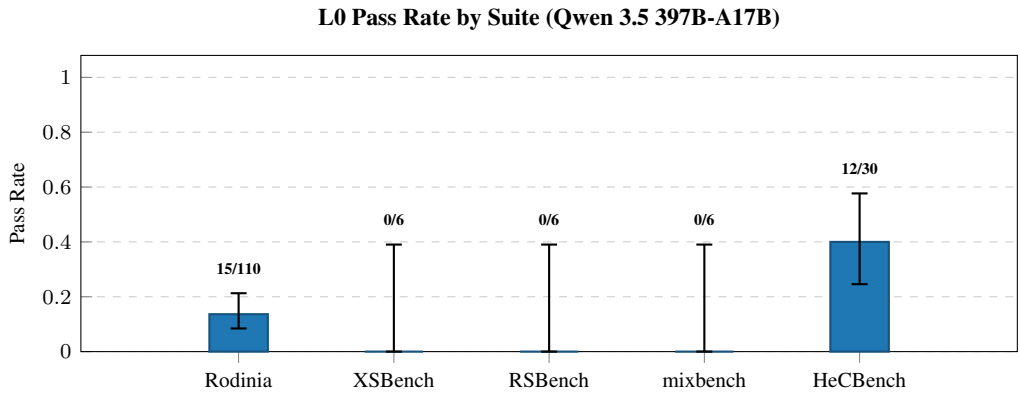

\begin{figure}[!htbp]
\centering
\begin{tikzpicture}
\begin{axis}[
  parbench compact,
  ybar,
  bar width=22pt,
  ymin=0, ymax=1.08,
  xmin=-0.6, xmax=4.6,
  xtick={0,1,2,3,4},
  xticklabels={Rodinia, XSBench, RSBench, mixbench, HeCBench},
  xticklabel style={font=\footnotesize},
  ytick={0,0.2,0.4,0.6,0.8,1.0},
  ylabel={Pass Rate},
  title={L0 Pass Rate by Suite (GPT-5.4)},
  grid style={dashed, draw=gray!40},
]
\addplot[
  fill=pbRodinia,
  draw=pbRodinia!70!black,
  line width=0.8pt,
  error bars/.cd,
    y dir=both,
    y explicit,
    error bar style={line width=0.8pt, black},
    error mark options={rotate=90, mark size=3pt, line width=0.8pt, black},
] table[x=x, y=y, y error minus=errm, y error plus=errp] {figures/figures_tek_version/f6_cross_suite_comparison_gpt_5-4.dat};
  \node[font=\tiny\bfseries, anchor=south] at (axis cs:0,0.6455) {47/92};
  \node[font=\tiny\bfseries, anchor=south] at (axis cs:1,0.9382) {4/6};
  \node[font=\tiny\bfseries, anchor=south] at (axis cs:2,0.9382) {4/6};
  \node[font=\tiny\bfseries, anchor=south] at (axis cs:3,0.9382) {4/6};
  \node[font=\tiny\bfseries, anchor=south] at (axis cs:4,1.0136) {24/26};
\end{axis}
\end{tikzpicture}
\caption{Cross-suite pass rate comparison (L0, \gptnew{}). Per-suite aggregate pass rates with Wilson 95\% CIs across all five benchmark suites.}
\label{fig:cross-suite-gpt54}
\end{figure}
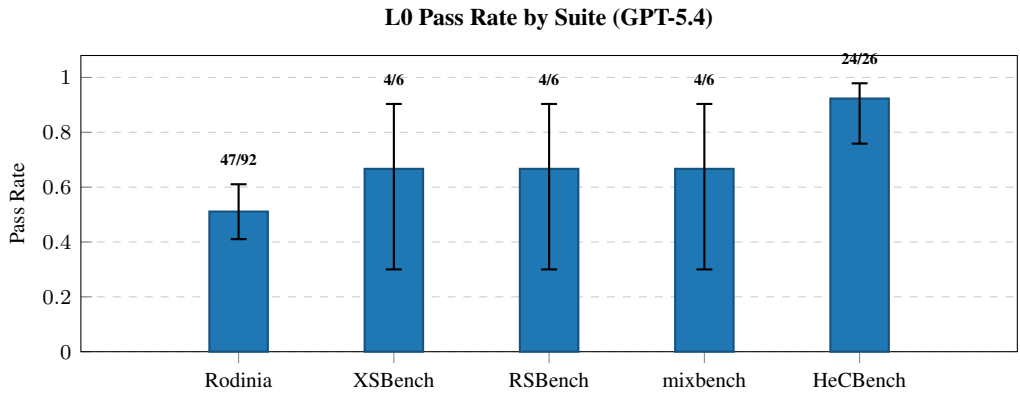

\begin{figure}[!htbp]
\centering
\begin{tikzpicture}
\begin{axis}[
  parbench compact,
  ybar,
  bar width=22pt,
  ymin=0, ymax=1.08,
  xmin=-0.6, xmax=4.6,
  xtick={0,1,2,3,4},
  xticklabels={Rodinia, XSBench, RSBench, mixbench, HeCBench},
  xticklabel style={font=\footnotesize},
  ytick={0,0.2,0.4,0.6,0.8,1.0},
  ylabel={Pass Rate},
  title={L0 Pass Rate by Suite (GPT-5.3-Codex)},
  grid style={dashed, draw=gray!40},
]
\addplot[
  fill=pbRodinia,
  draw=pbRodinia!70!black,
  line width=0.8pt,
  error bars/.cd,
    y dir=both,
    y explicit,
    error bar style={line width=0.8pt, black},
    error mark options={rotate=90, mark size=3pt, line width=0.8pt, black},
] table[x=x, y=y, y error minus=errm, y error plus=errp] {figures/figures_tek_version/f6_cross_suite_comparison_gpt_5-3_codex.dat};
  \node[font=\tiny\bfseries, anchor=south] at (axis cs:0,0.6455) {47/92};
  \node[font=\tiny\bfseries, anchor=south] at (axis cs:1,0.7350) {2/6};
  \node[font=\tiny\bfseries, anchor=south] at (axis cs:2,0.9382) {4/6};
  \node[font=\tiny\bfseries, anchor=south] at (axis cs:3,0.9382) {4/6};
  \node[font=\tiny\bfseries, anchor=south] at (axis cs:4,1.0350) {26/26};
\end{axis}
\end{tikzpicture}
\caption{Cross-suite pass rate comparison (L0, \codex{}). Per-suite aggregate pass rates with Wilson 95\% CIs across all five benchmark suites.}
\label{fig:cross-suite-codex}
\end{figure}
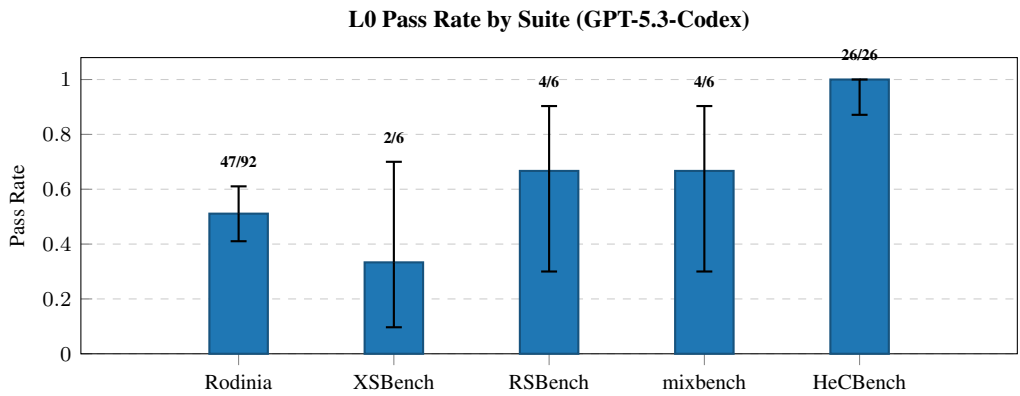


\begin{table*}[!htbp]
\centering
\caption{Full per-kernel pass rates across all 31 evaluated kernels (\qwenshort{}, 626 evaluation-eligible records including L0 and augmentation levels, 31~kernels across 5~suites). Results involving \knownfail{} specs excluded.}
\label{tab:per-kernel-full}
\footnotesize
\begin{tabular}{@{}llrrrrl@{}}
\toprule
Suite & Kernel & Total & PASS & Fail & Rate & 95\% Wilson CI \\
\midrule
HeCBench & convolution1d & 10 & 2 & 8 & 20.0\% & [5.7\%, 51.0\%] \\
HeCBench & floydwarshall & 38 & 29 & 9 & 76.3\% & [60.8\%, 87.0\%] \\
HeCBench & heat2d & 38 & 29 & 9 & 76.3\% & [60.8\%, 87.0\%] \\
HeCBench & iso2dfd & 38 & 32 & 6 & 84.2\% & [69.6\%, 92.6\%] \\
HeCBench & jacobi & 10 & 3 & 7 & 30.0\% & [10.8\%, 60.3\%] \\
HeCBench & md & 10 & 4 & 6 & 40.0\% & [16.8\%, 68.7\%] \\
HeCBench & nqueen & 10 & 6 & 4 & 60.0\% & [31.3\%, 83.2\%] \\
HeCBench & page-rank & 10 & 6 & 4 & 60.0\% & [31.3\%, 83.2\%] \\
HeCBench & scan & 6 & 0 & 6 & 0.0\% & [0.0\%, 39.0\%] \\
HeCBench & stencil1d & 14 & 9 & 5 & 64.3\% & [38.8\%, 83.7\%] \\
\midrule
Rodinia & backprop & 6 & 0 & 6 & 0.0\% & [0.0\%, 39.0\%] \\
Rodinia & bfs & 34 & 15 & 19 & 44.1\% & [28.9\%, 60.5\%] \\
Rodinia & bptree & 18 & 0 & 18 & 0.0\% & [0.0\%, 17.6\%] \\
Rodinia & cfd & 22 & 6 & 16 & 27.3\% & [13.2\%, 48.2\%] \\
Rodinia & dwt2d & 6 & 0 & 6 & 0.0\% & [0.0\%, 39.0\%] \\
Rodinia & gaussian & 6 & 0 & 6 & 0.0\% & [0.0\%, 39.0\%] \\
Rodinia & heartwall & 18 & 0 & 18 & 0.0\% & [0.0\%, 17.6\%] \\
Rodinia & hotspot & 30 & 14 & 16 & 46.7\% & [30.2\%, 63.9\%] \\
Rodinia & hotspot3d & 34 & 20 & 14 & 58.8\% & [42.2\%, 73.6\%] \\
Rodinia & lavamd & 18 & 0 & 18 & 0.0\% & [0.0\%, 17.6\%] \\
Rodinia & lud & 30 & 14 & 16 & 46.7\% & [30.2\%, 63.9\%] \\
Rodinia & myocyte & 18 & 0 & 18 & 0.0\% & [0.0\%, 17.6\%] \\
Rodinia & nn & 6 & 0 & 6 & 0.0\% & [0.0\%, 39.0\%] \\
Rodinia & nw & 30 & 11 & 19 & 36.7\% & [21.9\%, 54.5\%] \\
Rodinia & particlefilter & 26 & 9 & 17 & 34.6\% & [19.4\%, 53.8\%] \\
Rodinia & pathfinder & 30 & 8 & 22 & 26.7\% & [14.2\%, 44.4\%] \\
Rodinia & srad & 26 & 9 & 17 & 34.6\% & [19.4\%, 53.8\%] \\
Rodinia & streamcluster & 22 & 1 & 21 & 4.5\% & [0.8\%, 21.8\%] \\
\midrule
mixbench & mixbench & 22 & 2 & 20 & 9.1\% & [2.5\%, 27.8\%] \\
RSBench & rsbench & 18 & 0 & 18 & 0.0\% & [0.0\%, 17.6\%] \\
XSBench & xsbench & 22 & 1 & 21 & 4.5\% & [0.8\%, 21.8\%] \\
\bottomrule
\end{tabular}
\end{table*}

\begin{table*}[!htbp]
\centering
\caption{Full per-kernel pass rates across all 31 evaluated kernels (\gptnew{}, 822 evaluation-eligible records including L0 and augmentation levels, 31~kernels across 5~suites). \knownfail{} specs pre-excluded from evaluation batch.}
\label{tab:per-kernel-full-gpt54}
\footnotesize
\begin{tabular}{@{}llrrrrl@{}}
\toprule
Suite & Kernel & Total & PASS & Fail & Rate & 95\% Wilson CI \\
\midrule
HeCBench & convolution1d & 14 & 14 & 0 & 100.0\% & [78.5\%, 100.0\%] \\
HeCBench & floydwarshall & 42 & 41 & 1 & 97.6\% & [87.7\%, 99.6\%] \\
HeCBench & heat2d & 42 & 40 & 2 & 95.2\% & [84.2\%, 98.7\%] \\
HeCBench & iso2dfd & 42 & 42 & 0 & 100.0\% & [91.6\%, 100.0\%] \\
HeCBench & jacobi & 14 & 14 & 0 & 100.0\% & [78.5\%, 100.0\%] \\
HeCBench & md & 14 & 13 & 1 & 92.9\% & [68.5\%, 98.7\%] \\
HeCBench & nqueen & 14 & 14 & 0 & 100.0\% & [78.5\%, 100.0\%] \\
HeCBench & page-rank & 14 & 12 & 2 & 85.7\% & [60.1\%, 96.0\%] \\
HeCBench & scan & 14 & 14 & 0 & 100.0\% & [78.5\%, 100.0\%] \\
HeCBench & stencil1d & 14 & 14 & 0 & 100.0\% & [78.5\%, 100.0\%] \\
\midrule
Rodinia & backprop & 10 & 7 & 3 & 70.0\% & [39.7\%, 89.2\%] \\
Rodinia & bfs & 38 & 35 & 3 & 92.1\% & [79.2\%, 97.3\%] \\
Rodinia & bptree & 26 & 13 & 13 & 50.0\% & [32.1\%, 67.9\%] \\
Rodinia & cfd & 34 & 28 & 6 & 82.4\% & [66.5\%, 91.7\%] \\
Rodinia & dwt2d & 10 & 7 & 3 & 70.0\% & [39.7\%, 89.2\%] \\
Rodinia & gaussian & 6 & 0 & 6 & 0.0\% & [0.0\%, 39.0\%] \\
Rodinia & heartwall & 22 & 6 & 16 & 27.3\% & [13.2\%, 48.2\%] \\
Rodinia & hotspot & 34 & 28 & 6 & 82.4\% & [66.5\%, 91.7\%] \\
Rodinia & hotspot3d & 38 & 26 & 12 & 68.4\% & [52.5\%, 80.9\%] \\
Rodinia & lavamd & 26 & 8 & 18 & 30.8\% & [16.5\%, 50.0\%] \\
Rodinia & lud & 34 & 27 & 7 & 79.4\% & [63.2\%, 89.7\%] \\
Rodinia & myocyte & 22 & 5 & 17 & 22.7\% & [10.1\%, 43.4\%] \\
Rodinia & nn & 14 & 14 & 0 & 100.0\% & [78.5\%, 100.0\%] \\
Rodinia & nw & 34 & 23 & 11 & 67.6\% & [50.8\%, 80.9\%] \\
Rodinia & particlefilter & 42 & 33 & 9 & 78.6\% & [64.1\%, 88.3\%] \\
Rodinia & pathfinder & 34 & 27 & 7 & 79.4\% & [63.2\%, 89.7\%] \\
Rodinia & srad & 38 & 28 & 10 & 73.7\% & [58.0\%, 85.0\%] \\
Rodinia & streamcluster & 22 & 7 & 15 & 31.8\% & [16.4\%, 52.7\%] \\
\midrule
mixbench & mixbench & 38 & 27 & 11 & 71.1\% & [55.2\%, 83.0\%] \\
RSBench & rsbench & 38 & 29 & 9 & 76.3\% & [60.8\%, 87.0\%] \\
XSBench & xsbench & 38 & 25 & 13 & 65.8\% & [49.9\%, 78.8\%] \\
\bottomrule
\end{tabular}
\end{table*}

\begin{table*}[!htbp]
\centering
\caption{Full per-kernel pass rates across all 31 evaluated kernels (\codex{}, 814 evaluation-eligible records including L0 and augmentation levels, 31~kernels across 5~suites). \knownfail{} specs pre-excluded from evaluation batch.}
\label{tab:per-kernel-full-codex}
\footnotesize
\begin{tabular}{@{}llrrrrl@{}}
\toprule
Suite & Kernel & Total & PASS & Fail & Rate & 95\% Wilson CI \\
\midrule
HeCBench & convolution1d & 14 & 14 & 0 & 100.0\% & [78.5\%, 100.0\%] \\
HeCBench & floydwarshall & 42 & 40 & 2 & 95.2\% & [84.2\%, 98.7\%] \\
HeCBench & heat2d & 42 & 39 & 3 & 92.9\% & [81.0\%, 97.5\%] \\
HeCBench & iso2dfd & 42 & 42 & 0 & 100.0\% & [91.6\%, 100.0\%] \\
HeCBench & jacobi & 14 & 14 & 0 & 100.0\% & [78.5\%, 100.0\%] \\
HeCBench & md & 14 & 14 & 0 & 100.0\% & [78.5\%, 100.0\%] \\
HeCBench & nqueen & 14 & 13 & 1 & 92.9\% & [68.5\%, 98.7\%] \\
HeCBench & page-rank & 14 & 13 & 1 & 92.9\% & [68.5\%, 98.7\%] \\
HeCBench & scan & 14 & 14 & 0 & 100.0\% & [78.5\%, 100.0\%] \\
HeCBench & stencil1d & 14 & 14 & 0 & 100.0\% & [78.5\%, 100.0\%] \\
\midrule
Rodinia & backprop & 6 & 0 & 6 & 0.0\% & [0.0\%, 39.0\%] \\
Rodinia & bfs & 38 & 35 & 3 & 92.1\% & [79.2\%, 97.3\%] \\
Rodinia & bptree & 30 & 13 & 17 & 43.3\% & [27.4\%, 60.8\%] \\
Rodinia & cfd & 34 & 27 & 7 & 79.4\% & [63.2\%, 89.7\%] \\
Rodinia & dwt2d & 10 & 6 & 4 & 60.0\% & [31.3\%, 83.2\%] \\
Rodinia & gaussian & 6 & 0 & 6 & 0.0\% & [0.0\%, 39.0\%] \\
Rodinia & heartwall & 22 & 3 & 19 & 13.6\% & [4.7\%, 33.3\%] \\
Rodinia & hotspot & 34 & 28 & 6 & 82.4\% & [66.5\%, 91.7\%] \\
Rodinia & hotspot3d & 38 & 33 & 5 & 86.8\% & [72.7\%, 94.2\%] \\
Rodinia & lavamd & 22 & 6 & 16 & 27.3\% & [13.2\%, 48.2\%] \\
Rodinia & lud & 34 & 28 & 6 & 82.4\% & [66.5\%, 91.7\%] \\
Rodinia & myocyte & 22 & 7 & 15 & 31.8\% & [16.4\%, 52.7\%] \\
Rodinia & nn & 14 & 14 & 0 & 100.0\% & [78.5\%, 100.0\%] \\
Rodinia & nw & 38 & 26 & 12 & 68.4\% & [52.5\%, 80.9\%] \\
Rodinia & particlefilter & 38 & 29 & 9 & 76.3\% & [60.8\%, 87.0\%] \\
Rodinia & pathfinder & 34 & 28 & 6 & 82.4\% & [66.5\%, 91.7\%] \\
Rodinia & srad & 34 & 26 & 8 & 76.5\% & [60.0\%, 87.6\%] \\
Rodinia & streamcluster & 22 & 7 & 15 & 31.8\% & [16.4\%, 52.7\%] \\
\midrule
mixbench & mixbench & 42 & 32 & 10 & 76.2\% & [61.5\%, 86.5\%] \\
RSBench & rsbench & 38 & 20 & 18 & 52.6\% & [37.3\%, 67.5\%] \\
XSBench & xsbench & 34 & 19 & 15 & 55.9\% & [39.5\%, 71.1\%] \\
\bottomrule
\end{tabular}
\end{table*}

\end{document}